\documentclass[10pt,twocolumn,letterpaper]{article}
 \usepackage{cvpr}              
\definecolor{cvprblue}{rgb}{0.21,0.49,0.74}
\usepackage{multirow}
\usepackage{makecell}
\usepackage{bm}
\usepackage[pagebackref,breaklinks,colorlinks,allcolors=cvprblue]{hyperref}

\def\paperID{} 
\def\confName{CVPR}
\def\confYear{2026}

\newcommand{\mypar}[1]{\vspace{0.0em}\noindent\textbf{#1}\textbf{.}}
\newcommand{\ourbench}{{\sc {AVA-Bench}}\xspace}

\title{AVA-Bench: \underline{A}tomic \underline{V}isual \underline{A}bility Benchmark for Vision Foundation Models}

\author{
{Zheda Mai}$^{1*}$ \hspace{0.6em}
{Arpita Chowdhury}$^{1*}$ \hspace{0.6em}
{Zihe Wang}$^{1}$ \hspace{0.6em}
{Sooyoung Jeon}$^{1}$\\
{Lemeng Wang}$^{1}$ \hspace{0.6em}
{Jiacheng Hou}$^{1}$ \hspace{0.6em}
{Jihyung Kil}$^{2\dagger}$ \hspace{0.6em}
{Wei-Lun Chao}$^{3}$\\
$^{1}$The Ohio State University  \qquad $^{2}$Adobe Research \qquad $^{3}$Boston University \\
{\tt\small \url{https://zheda-mai.github.io/AVA-Bench}}  \quad  {\tt\small \{mai.145, chowdhury.150\}@osu.edu}
}

\begin{document}
\twocolumn[{
    \renewcommand\twocolumn[1][]{#1}
    \maketitle
\vspace{-1.8em}
    \centering
    \includegraphics[width=\linewidth]{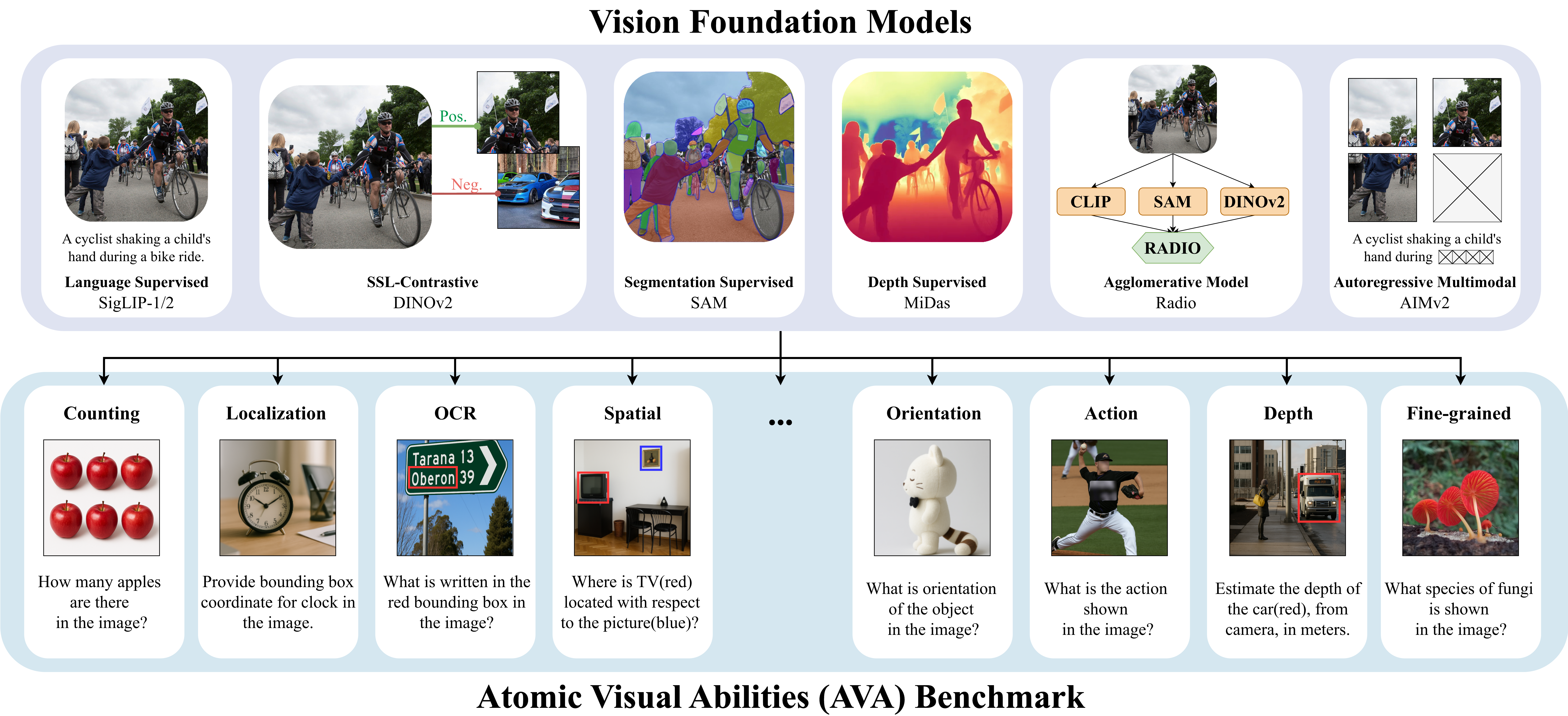}
    \vspace{-1.5em}
    \captionof{figure}{
      \small Vision foundation models (VFMs) trained with different data and objectives are evaluated on the proposed \ourbench\ to assess their strengths and limitations across atomic visual abilities (AVAs).
    }
    \label{fig:ava_eval}
    \vspace{1.5em}
}]
{
\renewcommand{\thefootnote}{\fnsymbol{footnote}}
\footnotetext[1]{Equal contribution.}
\footnotetext[2]{\hangindent=1.8em Acted in an advisory capacity and did not process, store, or direct the use of data or models in this project.}
}


\begin{abstract}





The rise of vision foundation models (VFMs) calls for systematic evaluation. A common approach pairs VFMs with large language models (LLMs) as general-purpose heads, followed by evaluation on broad Visual Question Answering (VQA) benchmarks. However, this protocol has two key blind spots: (i) Instruction tuning data may not align with VQA test distributions, meaning a wrong prediction can stem from such data mismatch rather than VFMs' visual shortcomings; (ii) VQA benchmarks often require multiple visual abilities in a single question, making it difficult to determine whether errors arise from the lack of all required abilities or just one key ability. 
To address these gaps, we introduce \ourbench, the first benchmark that explicitly disentangles 14 Atomic Visual Abilities (AVAs)---foundational skills like localization, depth estimation, and spatial understanding that collectively support complex visual reasoning tasks. By decoupling AVAs and matching training and test distributions within each, \ourbench pinpoints exactly where a VFM excels or falters.
Applying \ourbench to leading VFMs thus reveals distinctive “ability fingerprints,” turning VFM selection from educated guesswork into principled engineering. Notably, we find that a 0.5B LLM yields similar VFM rankings as a 7B LLM while cutting GPU hours by {8$\times$}, enabling more efficient evaluation. By offering a comprehensive and transparent benchmark, we hope \ourbench lays the foundation for the next generation of VFMs.
\end{abstract}

\section{Introduction}

 \begin{figure*}
    \centering
    \includegraphics[width=1\linewidth]{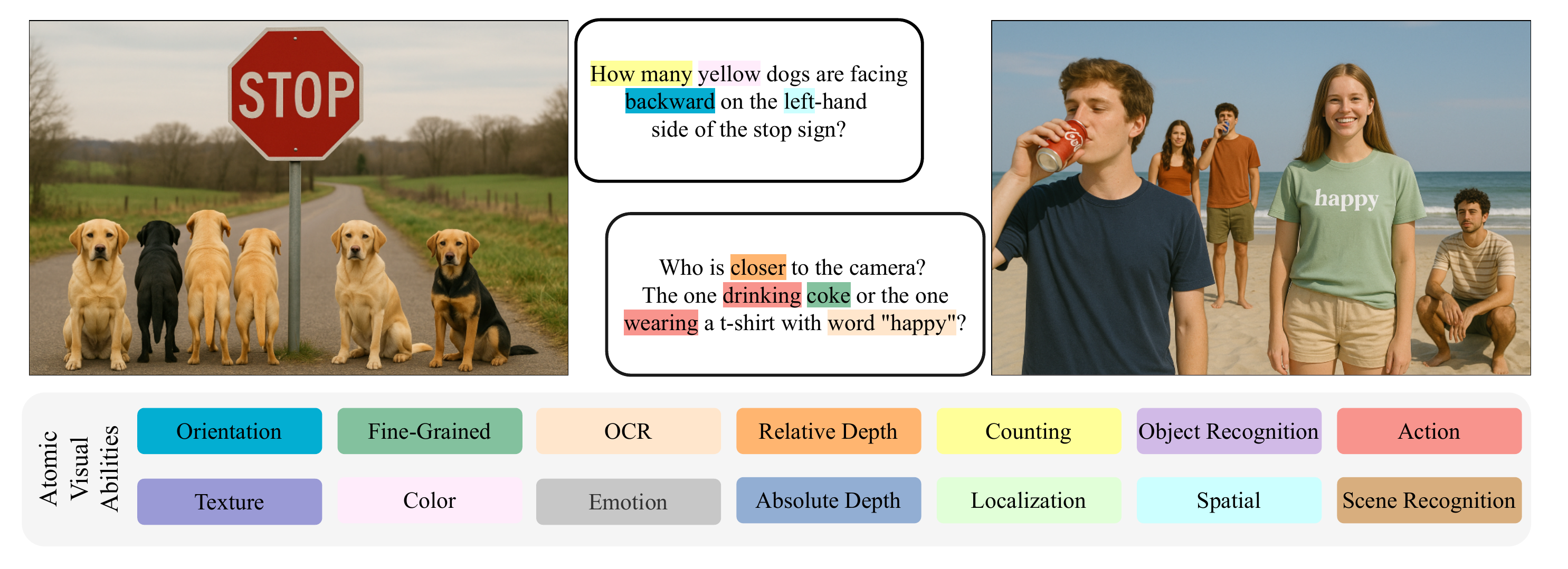}
    \vspace{-2.2em}
    \caption{\small Visual Question Answering (VQA) often requires multiple atomic visual abilities to answer a question. When a model makes an incorrect prediction, it's hard to determine whether it stems from a failure to capture all required AVAs or just a single critical one. 
    }
    \vskip-2pt
    \label{fig: ava_demo}
\end{figure*}


Vision Foundation Models (VFMs), pre-trained on large and diverse datasets, have become central to AI by providing transferable features for a wide range of downstream tasks~\citep{khan2022transformers,awais2025foundation,chowdhury2025prompt,bommasani2021opportunities}. The variety of pre-training objectives and supervision signals has led to a proliferation of specialized VFMs---such as DINOv2~\citep{oquab2023dinov2}, CLIP~\citep{radford2021learning}, and SAM~\citep{kirillov2023segment}---each excelling in distinct visual capabilities while often exhibiting interesting emergent properties~\citep{goldblum2023battle,caron2021emerging,naseer2021intriguing}. 
Thus, establishing a systematic and effective evaluation protocol for VFMs becomes increasingly crucial.

\begin{figure*}
    \centering
    \includegraphics[width=1\linewidth]{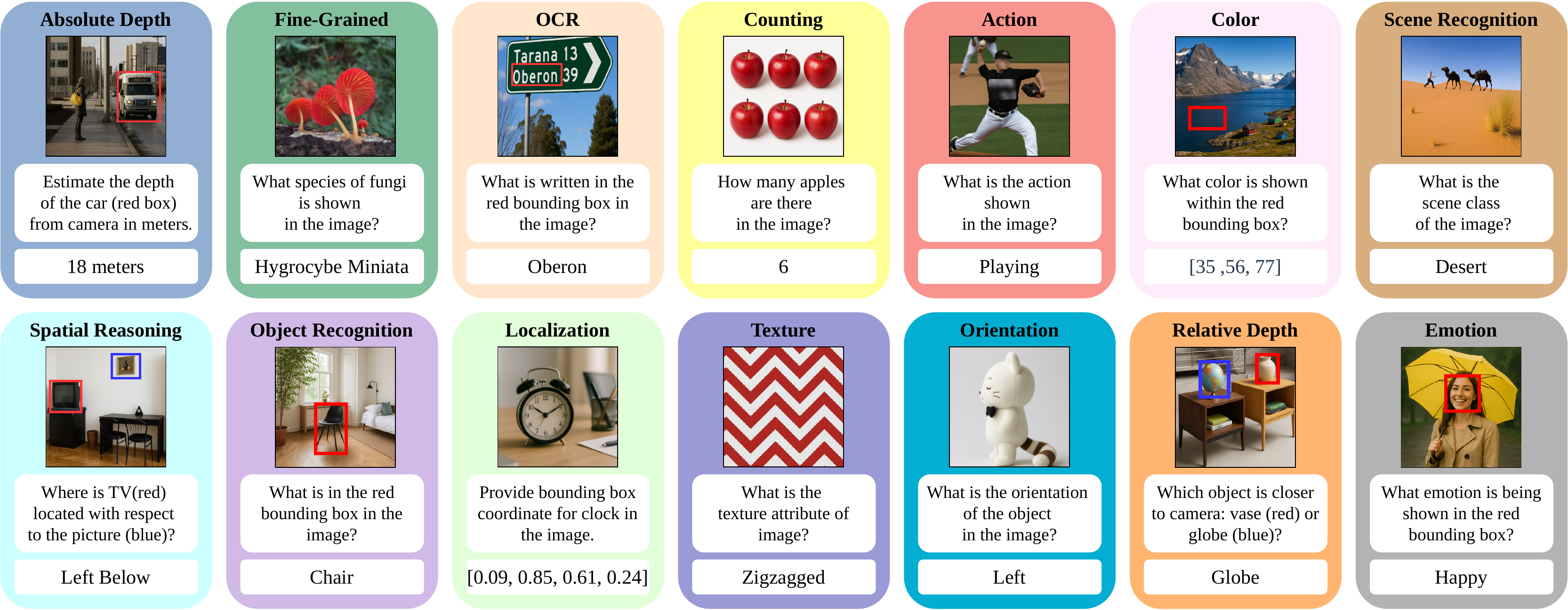}
    \vspace{-1.3em}
    \caption{\small \ourbench consists of 14 Atomic Visual Abilities that can be combined to address more complex visual reasoning tasks. }
    \label{fig: overall_ava}
    \vskip-2pt
\end{figure*}

Existing evaluation protocols can generally be categorized into two groups. The first focuses on task-specific capabilities, typically attaching tailored heads to VFMs, followed by fine-tuning and evaluation on dedicated datasets such as ImageNet for classification~\citep{han2022survey,mai2024fine} and COCO for detection or segmentation~\citep{thisanke2023semantic, balachandran2024eureka}. To better capture the diverse and complex perception challenges of the real world, recent studies advocate a more generic approach that leverages large language models (LLMs) as general-purpose heads, evaluating VFMs on broad Visual Question Answering (VQA) benchmarks~\citep{liu2023visual,zhu2023minigpt,chowdhery2023palm}. 

While increasingly adopted, this generic protocol may suffer from two potential blind spots: (i) discrepancies between instruction tuning data and VQA test data lead to performance drops due to data mismatch rather than genuine visual limitations in VFMs, and (ii) existing VQA benchmarks typically require multiple visual abilities simultaneously, making it difficult to determine whether a failure arises from the absence of multiple abilities or merely a single critical one.

To this end, we introduce \textbf{\ourbench}, the first VFM evaluation benchmark explicitly designed to disentangle  {\textbf{\underline{A}}tomic \textbf{\underline{V}}isual \textbf{\underline{A}}bilities (\textbf{AVAs})}---the fundamental visual capabilities that combine to solve complex visual reasoning tasks.
For instance, answering typical VQA questions shown in \autoref{fig: ava_demo} necessitates integrating several AVAs.

Specifically, \ourbench evaluates VFMs across \textbf{14} carefully identified AVAs (see \autoref{fig: overall_ava}), including \textbf{localization}, \textbf{counting}, \textbf{spatial reasoning}, \textbf{orientation}, \textbf{absolute and relative depth estimation}, and recognition of \textbf{textures}, \textbf{colors}, \textbf{objects}, \textbf{actions},  \textbf{emotions}, \textbf{optical characters (OCR)}, and \textbf{scenes}. 
Each AVA comes with distribution-matched train and test splits and is probed \emph{in isolation}, eliminating the two aforementioned ambiguities. 
This enables \ourbench to pinpoint exactly where a VFM excels or falters, providing a clear picture of its strengths and weaknesses.

We systematically benchmark leading VFMs trained under diverse objectives and data (see \autoref{fig:ava_eval}), covering language-supervised (\eg, SigLIP-1/2~\citep{tschannen2025siglip,zhai2023sigmoid}, CLIP~\citep{radford2021learning}, InternVL-2.5~\citep{chen2024expanding}), multimodal autoregressive (AIMv2~\citep{fini2024multimodal}), segmentation-supervised (SAM~\citep{kirillov2023segment}), depth-supervised (MiDas~\citep{ranftl2020towards}), contrastive self-supervised (DINOv2~\citep{oquab2023dinov2}), and agglomerative models (RADIO~\citep{ranzinger2024radio}). We follow the standard protocol of adding an LLM on top, but fine-tune it separately for each AVA.

Our extensive analyses reveal the following main findings: \textbf{(1)} SigLIP-1/2 and AIMv2 emerge as the most versatile VFMs, achieving the highest average rank across all AVAs---\emph{highlighting the critical role of language supervision in enhancing general visual capability}; \textbf{(2)} For vision-centric AVAs such as localization, absolute depth estimation, and orientation, the SSL-based DINOv2 performs comparably or better than language-supervised counterparts; \textbf{(3)} Conversely, language-centric tasks such as OCR strongly favor language-supervised VFMs; \textbf{(4)} Last but not least, we observe that VFMs universally excel at low- to mid-level AVAs (\eg, texture, relative depth estimation, object recognition), regardless of their training objectives---suggesting that VQA failures typically stem from \emph{deficiencies in specific critical AVAs} rather than broad visual incompetence. These insights turn VFM selection from educated guesswork into principled engineering, enabling practitioners to choose (or ensemble) VFMs based on the specific AVA strengths required by their downstream tasks.

Alongside our main study, we identify a more resource-efficient evaluation strategy. Existing LLM-based VFM evaluations typically rely on heavyweight models such as Vicuna-1.5(7B/13B)~\citep{tong2024cambrian,huang2024survey}, aiming for high absolute accuracy but incurring significant computational overhead. 
However, when the goal is to compare VFMs, we advocate \emph{prioritizing relative performance over absolute accuracy.} 
We demonstrate that a lightweight \underline{0.5B} LLM preserves reliable VFM rankings while reducing evaluation costs by \textbf{8$\times$}, making large-scale analysis substantially more practical. We contextualize our findings within related work~\citep{goldblum2023battle,tong2024cambrian} to offer a holistic understanding of VFMs and highlight future research directions (\autoref{sec: discussion}).

\noindent\textbf{Contributions.} Our key contributions are three-fold:
\begin{itemize}[leftmargin=*,
                itemsep=0pt,
                parsep=0pt,
                topsep=0pt,
                partopsep=0pt]
    \item  We identify critical blind spots in existing evaluation protocols and introduce \ourbench, a systematic, diagnostic, and comprehensive VFM evaluation benchmark covering 14 atomic visual abilities (AVAs), clearly highlighting VFMs' fundamental strengths and weaknesses.
    \item We conduct a detailed evaluation and insightful analysis of diverse leading VFMs, deriving actionable guidance for VFM selection in downstream applications such as customized MLLMs.
    \item We release a resource-efficient evaluation protocol with an open-source codebase to facilitate the development of the next generation of accountable and versatile VFMs.
\end{itemize}

\noindent\textbf{Remark.} We want to emphasize that \ourbench is designed to systematically evaluate VFMs (\eg, DINOv2) instead of  MLLMs (\eg, LLaVA). MLLM in our study solely acts as a general-purpose \textit{{prediction head}} for the underlying VFM, enabling a unified evaluation interface across tasks.

\section{Vision Foundation Models and Evaluation}



\subsection{Vision Foundation Models (VFMs)}

\begin{figure*}[t] 
    \centering 
    \includegraphics[width=\linewidth]{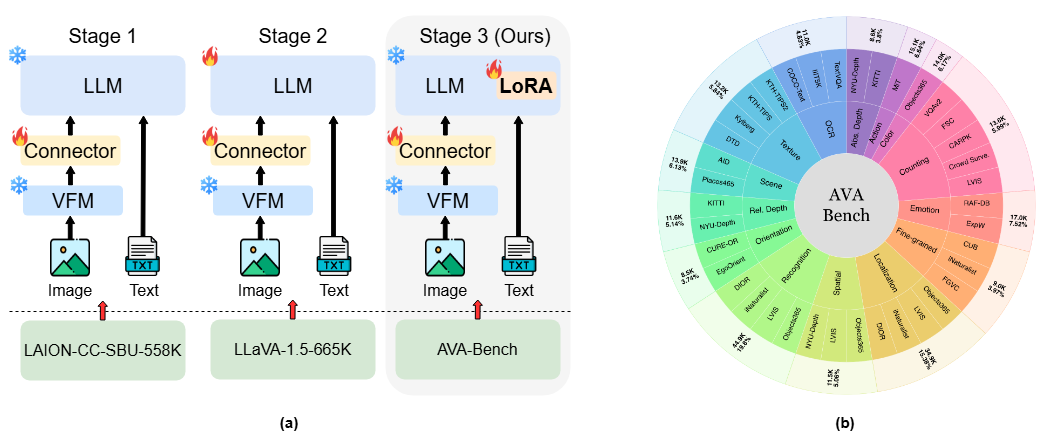} 
    \vspace{-2.5em} 
    \caption{\small \textbf{(a)} Evaluation pipeline for \ourbench: The standard LLaVA-style two-stage training prepares the connector and LLM for VFM evaluation. For each AVA, only the connector and LoRA are trained. \textbf{(b)} Overall statistics of \ourbench.} 
    \label{fig: arch_and_ava} 
\end{figure*}

A wide range of architectures and learning strategies have been explored to develop vision foundation models (VFMs). In general, they fall into two categories based on the nature of their training data. On one hand, VFMs are pre-trained purely on visual data. For instance, the Vision Transformer (ViT) ~\citep{dosovitskiy2020image} is trained on labeled images using the supervised signal to capture visual representations. DINOv2 ~\citep{oquab2023dinov2}, in contrast, adopts a self-supervised learning approach, enabling the model to learn those features without labeling. Some VFMs are designed for specific vision tasks---for example, SAM~\citep{kirillov2023segment} specializes in open-vocabulary segmentation, while MiDaS~\citep{ranftl2020towards} focuses on monocular depth estimation. RADIO ~\citep{ranzinger2024radio} instead introduces a multi-teacher distillation framework that unifies the strengths of different VFMs (\eg, CLIP, DINOv2, and SAM) into a single efficient student model.

Another family of VFMs leverages image-text pairs. CLIP ~\citep{radford2021learning} uses contrastive loss to align image and text captions, while SigLIP  ~\citep{tschannen2025siglip,zhai2023sigmoid} replaces the contrastive loss with a sigmoid one for efficient training. Unlike the conventional language-guided VFMs that process images and text separately, AIMv2 ~\citep{fini2024multimodal} integrates visual and textual understanding into a single auto-regressive framework, which is simple yet effective. Due to their different architectures and learning objectives, VFMs may have varying strengths and limitations in visual understanding.


\subsection{LLM-based Evaluation for VFMs}
\label{sec: prev_eval}


Unlike the traditional paradigm, where different tasks (\eg classification and segmentation) necessitate task-specific models~\citep{han2022survey, awais2025foundation}, the widespread adoption of Large Language Models (LLMs) as versatile interfaces has significantly shifted this paradigm~\citep{zhang2024mm, wu2023multimodal}. Reflecting this shift, recent studies advocate leveraging LLMs as general-purpose heads and evaluating VFMs on broad Visual Question Answering (VQA) benchmarks~\citep{tong2024cambrian}. Specifically, following the LLaVA approach, these studies utilize a two-stage framework: (i) pre-training a connector between frozen LLM and VFMs with image-text pairs for feature alignment (ii) fine-tuning both the connector and the LLM using instruction-tuning data, while keeping VFMs frozen during both stages (see \autoref{fig: arch_and_ava} (a)). This LLM-based evaluation protocol has rapidly gained popularity as it closely mirrors the contemporary multimodal LLM setting and effectively captures diverse real-world perception challenges.
\begin{figure*}
    \includegraphics[width=\linewidth]{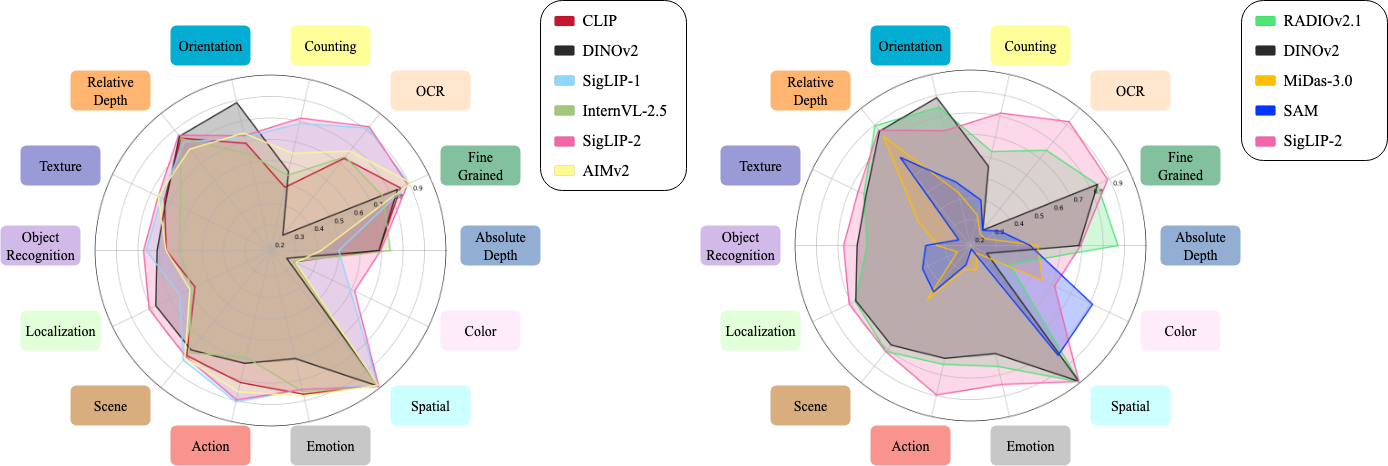}
    \vspace{-1.2em}
    \caption{\small Performance comparison of VFMs across all AVAs. (Left) Language-Supervised VFMs with DINOv2 as a reference. (Right) Other VFMs with the SigLIP-2 as a reference. }
    \vskip-5pt
    \label{fig: spider}
\end{figure*}

However, this evaluation protocol suffers from two potential blind spots. First, discrepancies may exist between instruction-tuning data and the test VQA datasets. Thus, a miss-prediction might arise from data mismatch rather than genuine visual deficiencies in VFMs. Second, typical VQA benchmarks typically require multiple visual abilities simultaneously to produce correct answers. This multi-ability requirement obscures the exact cause behind a model’s incorrect predictions. For instance, if a VFM is proficient in almost all visual abilities except orientation recognition (forward or backward), it would fail on the question in \autoref{fig: ava_demo} (left). In such cases, current evaluations only provide relative comparisons between VFMs without elucidating the specific missing capabilities. Thus, \ourbench aims to complement the LLM-based evaluation by pinpointing exactly where a VFM excels or falters, yielding a holistic understanding of their strengths and shortcomings.

\section{\ourbench}


\subsection{{\textbf{\underline{A}}tomic \textbf{\underline{V}}isual \textbf{\underline{A}}bilities (\textbf{AVAs})}}
\label{ss:atomic_visual_abil}


To address the blind spots discussed earlier, we introduce \ourbench, the first systematic evaluation suite explicitly designed to disentangle 14 fundamental perceptual skills--\textbf{\underline{A}}tomic \textbf{\underline{V}}isual \textbf{\underline{A}}bilities (\textbf{AVAs})--for VFMs.
AVAs are fundamental perceptual capabilities that can be combined to address more complex visual reasoning tasks. Rather than treating VQA as a monolithic task, we break it down into 14 fundamental perceptual abilities--such as counting, depth estimation, localization and spatial reasoning--each of which can be composed to answer more complex questions. For example, answering the question, \textit{“How many yellow dogs are facing backward on the left-hand side of the stop sign?”}, requires multiple AVAs: counting, color recognition, localization, and spatial reasoning (see \autoref{fig: ava_demo}). By explicitly defining the set of AVAs required to interpret a question, our benchmark quantitatively characterizes where a VFM excels or falters across each core ability.

The AVA selection is grounded in a thorough literature analysis, including compositional text-to-images benchmarks~\citep{huang2023t2i,kil2024mllm,wu2024conceptmix} and VQA questions~\citep{goyal2017making, ainslie2023gqa} (details in \autoref{-ss: ava-bench_details}), focusing strictly on pure perceptual tasks and excluding non-perceptual reasoning skills (\eg, mathematical reasoning).  Examples of each AVA can be found in~\autoref{fig: overall_ava} and detailed definitions can be found in \autoref{-ss: ava-bench_details}.

\subsection{Dataset Curation}
\label{ss:dataset_curation}
Constructing \ourbench required carefully isolating image–question pairs that specifically test individual AVAs. Existing MLLM and VQA datasets often blend multiple perceptual abilities, complicating direct assessments~\citep{tong2024eyes,zhang2024mme,yu2023mm}. To achieve clear isolation, we assembled a comprehensive suite comprising image–question pairs from \textbf{26} diverse datasets, explicitly aligning each pair with a single targeted AVA. These datasets span a broad range of domains—including general scenes, wildlife (e.g., birds, fungi, plants), vehicles, indoor/outdoor settings, and remote-sensing imagery.

 Given an AVA, image–question pairs in \ourbench were carefully designed or adapted to focus solely on that AVA. For instance, the question \textit{“What's the depth of the car from the camera?”} involves both localization and depth estimation. To isolate depth estimation, we explicitly provide the car’s bounding box. This approach enables fine-grained diagnostic insights into VFMs' AVA-specific strengths and weaknesses. Examples are illustrated in \autoref{fig: overall_ava}, with details below and in the \autoref{-ss: ava-bench_details}.

\subsubsection{Localization}

We curate \textbf{34.8K} localization-focused image–question pairs sourced from: \textbf{Objects365}~\citep{shao2019objects365} and \textbf{LVIS}~\citep{gupta2019lvis} (open-domain), \textbf{iNat}~\citep{van2021benchmarking} (birds and animals), and \textbf{DIOR}~\citep{li2020object} (remote-sensing). To ensure clarity, we include images with a single instance of the target object and exclude objects with small bounding boxes.    

\subsubsection{Counting}
We collect \textbf{13.6K} pairs from: \textbf{VQAv2}~\citep{goyal2017making} (open-domain), \textbf{FSC}~\citep{ranjan2021learning} (open-domain counting), \textbf{CARPK}~\citep{hsieh2017drone} (cars), \textbf{Crowd Surveillance Dataset}~\citep{li2022video} (people), and \textbf{LVIS}~\citep{gupta2019lvis} (open-domain). FSC, CARPK, and Crowd datasets explicitly cater to counting tasks, ensuring clear and distinct instance counts. LVIS instance masks are utilized to generate counting questions, complemented by counting-focused pairs from VQAv2.

\subsubsection{Spatial Reasoning}
We curate \textbf{11.5K} pairs from  \textbf{NYU-Depth V2}~\citep{silberman2012indoor} for indoor scenes, and \textbf{LVIS}~\citep{gupta2019lvis} and \textbf{Objects365}~\citep{shao2019objects365} for open-domain scenarios. For each image, we select two non-overlapping objects—one marked with a blue bounding box (the reference object) and another with a red bounding box (the target). The model is then asked to determine the relative spatial position of the target object with respect to the reference one, choosing from: \emph{top-left}, \emph{top-right}, \emph{bottom-left}, and \emph{bottom-right}.

\begin{figure*}
\centering
    \includegraphics[width=0.9\linewidth]{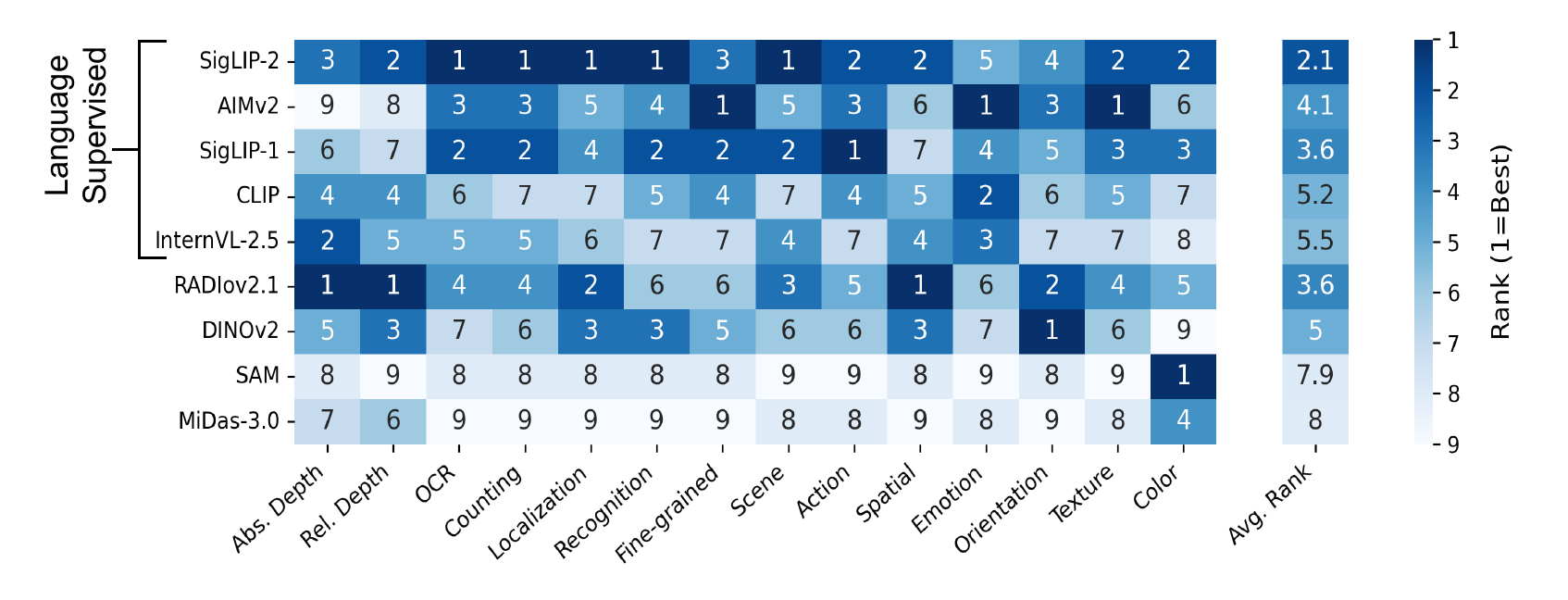}
    \vspace{-1em}
    \caption{\small The ranks of VFMs across all AVAs. Detailed results in \autoref{-ss: results}. }
    
    \label{fig: rank}

\end{figure*}

\subsubsection{Visual Attribute}
\mypar{Orientation} We compile \textbf{8.5K} pairs from two specialized datasets, \textbf{CURE-OR}~\citep{Temel2018_ICMLA} and \textbf{EgoOrientBench}~\citep{jung2024right}, providing uncluttered images of objects from nine distinct orientations: \emph{front}, \emph{back}, \emph{left}, \emph{right}, \emph{top}, \emph{front left}, \emph{front right}, \emph{back left}, and \emph{back right}. 

\mypar{Color} We curate \textbf{14K} pair sourcing from from open-domain datasets, such as \textbf{Objects365}~\citep{shao2019objects365} and \textbf{LVIS}~\citep{gupta2019lvis}. To isolate the color recognition from object semantics or background clutter,  we provide bounding boxes that tightly localize regions with minimal color variance, ensuring that the model focuses on RGB prediction.

\mypar{Texture} We curate \textbf{13.2K} pairs from: \textbf{DTD}~\citep{cimpoi14describing}, \textbf{Kylberg}~\citep{kylberg2011kylberg}, \textbf{KTH-TIPS} and \textbf{KTH-TIPS2}~\citep{article} with diverse texture types, such as \emph{striped}, \emph{aluminum foil}, and \emph{zigzagged}. 

\mypar{Emotion} We gather \textbf{17.0K} pairs from\textbf{RAF-DB}~\citep{li2019reliable} and \textbf{ExpW}~\citep{lian2020expression}, covering 7 annotated human emotional states: \emph{surprise}, \emph{neutral}, \emph{disgust}, \emph{fear}, \emph{happy}, \emph{sad}, and \emph{angry}.

\subsubsection{Depth Estimation}
We curate depth-focused image-question pairs from \textbf{NYU-Depth V2}~\citep{silberman2012indoor} for indoor scenes and \textbf{KITTI}~\citep{geiger2013vision} for outdoor scenes.  

\mypar{Absolute Depth} We assemble \textbf{9K} pairs and for each sample, we place a bounding box on a target object and ask the model to estimate its distance from the camera.  We ensure that each object class spans at least five distinct depth bins and that the samples within each bin are balanced.

\mypar{Relative Depth}  We collect \textbf{11.5K} pairs. Each image is annotated with two non-overlapping bounding boxes for two distinct objects. The model is asked to determine which object is closer to the camera. To prevent annotation bias, we ensure that each object class is evenly distributed between being the nearer or farther object to avoid cases where certain objects are always closer.

\subsubsection{Recognition}
\mypar{Action} We construct \textbf{15K} pairs from \textbf{Moments in Time}~\citep{monfort2019moments}, a video dataset with diverse human actions. We extract the middle frame from each video for 302 actions

\mypar{Fine-grained}
 We curate \textbf{9K} pairs from: \textbf{CUB-200}~\citep{wah2011caltech} (birds), \textbf{iNat-21}~\citep{van2021benchmarking} (fungi, plants, animals), and \textbf{Aircraft}~\citep{maji2013fine} (objects).  We randomly select 50 species from {iNat}, 100 species from {CUB} and all aircraft classes, resulting in a total of \textbf{300} classes.

\mypar{Object}
We curate \textbf{44.9K} pairs from 4 datasets with diverse domains across 70 unique objects: \textbf{Objects365}~\citep{shao2019objects365} and \textbf{LVIS}~\citep{gupta2019lvis} (open-domain), \textbf{iNaturalist-2021}~\citep{van2021benchmarking} (birds and animals), and \textbf{DIOR}~\citep{li2020object} (remote sensing). For each image, we provide a bounding box to eliminate the need for localization. 

\mypar{Scene}
We curate \textbf{13.9K} pairs from two datasets: \textbf{Places434}~\citep{zhou2017places} (open-domain) and \textbf{AID}~\citep{xia2017aid} (remote sensing). The model is required to select the correct scene class from 30 randomly sampled options. The full set includes \textbf{464} classes spanning a wide range of scenes.

\begin{figure*}
    \centering
    \includegraphics[width=1\linewidth]{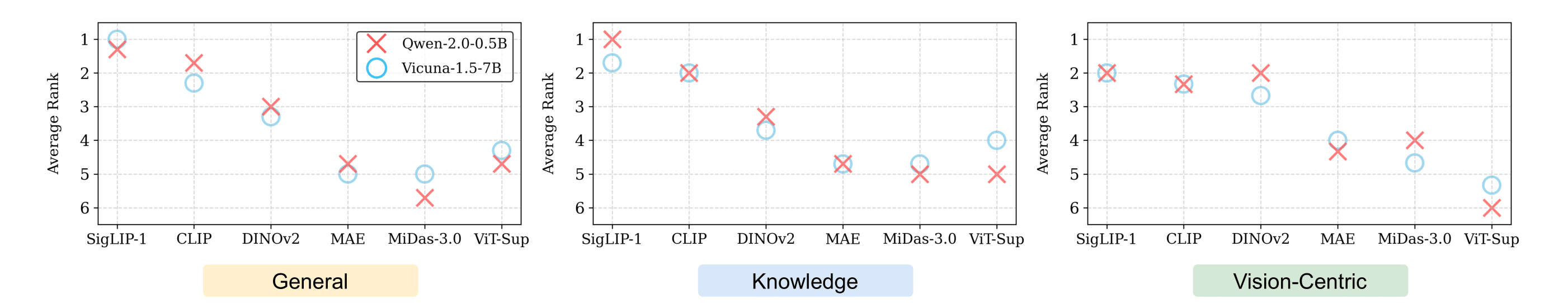}
    \vspace{-1.5em}
    \caption{\small A much smaller \underline{0.5B} LLM achieves similar relative VFM rankings comparable to a 7B LLM, while dramatically reducing evaluation costs by approximately \textbf{8$\times$}. }
    \label{fig: llm}
    \vspace{-0.8em}
\end{figure*}

\subsection{Quality Control and Dataset Statistics}
\label{ss:dataset_quality_ctrl}
We emphasize rigorous quality control to ensure fair, balanced, and unbiased assessments. Every AVA follows an 80/20 split in which the exact object classes and answer bins that appear in training are mirrored in testing. This ensures that performance differences truly reflect the VFM's perceptual capabilities rather than train-test distribution mismatches, effectively addressing the concerns highlighted in \autoref{sec: prev_eval}. Moreover, we try to avoid potential bias in the benchmark.  For example, in counting, we explicitly balance the number of samples per counting bin and per object type in both training and testing sets. This approach avoids biases wherein certain counts (e.g., “7 apples”) might dominate the training data, artificially inflating accuracy for specific numerical predictions. In another example, for localization, we set a clear threshold for minimum bounding-box area to ensure object visibility.  More details about how we ensure the quality of \ourbench can be found in \autoref{-ss: ava-bench_details}.  

To improve the generalizability within each AVA, we aggregate data from diverse datasets. This ensures that models encounter the same core visual ability across varied scenes, object types, and data distributions, making the assessment more robust and less dataset-specific. Image-question pairs are intentionally crafted or selected to be simple, clear, and explicitly focused on testing only \textbf{one} AVA at a time. In summary, \ourbench comprises \textbf{218K} meticulously curated image-question pairs that robustly isolate individual AVAs, carefully control dataset balance, visibility, and systematically prevent annotation biases. Statistics are provided in \autoref{fig: arch_and_ava} (b) , with more details in \autoref{-ss: ava-bench_details}.

\section{Evaluation Pipeline of \ourbench}
\label{sec: eval_pipeline}
To evaluate VFMs using \ourbench, we adopt the established LLM-based VFM evaluation protocol, employing the standard LLaVA-style two-stage training procedure to prepare the connector and LLM for VFM evaluation (details in \autoref{sec: prev_eval} and \autoref{fig: arch_and_ava} (a)). For each AVA in \ourbench, we fine-tune the connector and LLM while keeping the VFM frozen. Given the modest size of the training sets per AVA (typically around 6K–10K), we employ Parameter-Efficient Fine-Tuning (PEFT)~\citep{mai2025lessons, houlsby2019parameter, tu2023visual}, specifically Low-Rank Adaptation (LoRA)~\citep{hu2022lora}, to mitigate potential overfitting. Subsequently, the fine-tuned model is evaluated on the corresponding AVA-specific test sets.

\subsection{Is a Heavyweight LLM Evaluator Necessary?}

As discussed in \autoref{sec: prev_eval}, traditional LLM-based VFM evaluations, following the LLaVA protocol, predominantly rely on heavyweight models such as Vicuna-1.5 (7B/13B)\citep{liu2023visual}, aiming for high absolute accuracy but incurring considerable computational costs.  Nevertheless, a heavyweight LLM may not be mandatory for reliable comparative evaluations. When the goal is to compare VFMs, we advocate \emph{prioritizing relative performance over absolute metrics.} As depicted in \autoref{fig: llm}, a significantly smaller \underline{0.5B} LLM (Qwen2) achieves similar relative VFM rankings comparable to a 7B Vicuna-1.5, while dramatically reducing evaluation costs by approximately \textbf{8$\times$} (additional details in the \autoref{-ss: eval_efficiency}), making large-scale analysis substantially more practical. Thus, we utilize the lightweight 0.5B LLM for all subsequent experiments.

\section{Experiments}


\mypar{Metrics for AVAs}
 For absolute depth and counting AVAs, we utilized a normalized mean absolute error (MAE) relative to the ground-truth. This normalization ensures that errors involving greater distances or counts, which are inherently more challenging, are proportionally penalized less severely. Localization performance is evaluated using Generalized Intersection-over-Union (GIoU~\citep{rezatofighi2019generalized}), color via CIEDE2000~\citep{luo2001development}, and OCR through Average Normalized Levenshtein Similarity (ANLS)~\citep{biten2019scene}. All other AVAs employ standard accuracy metrics. Please refer to \autoref{-ss: experiment_details} for more metric details.

\subsection{Observations and Analyses}

 \begin{figure*}
  \centering
  \includegraphics[width=\linewidth]{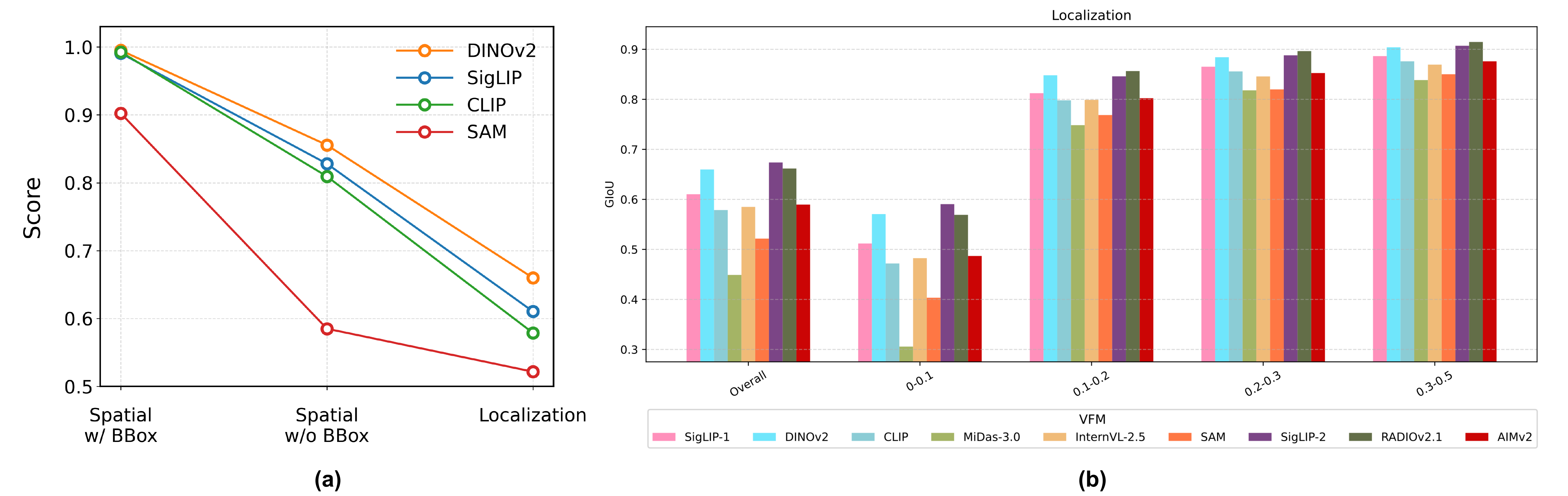}
      \vspace{-2em}
  \caption{\small \textbf{(a)} Impact of bounding boxes on spatial reasoning performance. With bounding boxes, all VFMs perform perfectly in spatial AVA; without them, models with weaker localization (MiDaS, SAM) perform worse. \textbf{(b)} Localization results for different splits based on ground-truth's bounding box sizes. 0.1 means the bounding box size is 10\% of the image size. Higher GIoU is better. }
  \label{fig:spatial_vs_localization}
  \vskip-5pt
\end{figure*}

\textbf{Bounding boxes isolate and assess specific AVAs.} Bounding boxes serve as a crucial tool for disentangling AVAs, allowing us to isolate and evaluate specific AVAs rather than compounded performance on complex tasks. For example, VFMs exhibit uniformly strong and similar performance in spatial reasoning when provided with ground-truth bounding boxes, indicating a shared strong spatial ability. However, removing the bounding boxes transforms this into a composite task requiring localization and spatial reasoning.  As shown in \autoref{fig:spatial_vs_localization}(a), such removal leads to substantial divergence in model performance, with rankings closely mirroring their localization ability. This contrast confirms that performance degradation on composite tasks is often attributable to deficiencies in specific underdeveloped AVAs rather than a fundamental failure in all visual capacities.

\mypar{Subgroup analyses reveal exceptions hidden by aggregate metrics} Aggregate metrics can sometimes obscure nuanced insights. To gain a deeper understanding, we conduct detailed analyses by partitioning test samples based on specific criteria (e.g., object size in localization tasks) and examining whether these subgroup trends align with overall performance. We split localization testing samples based on normalized bounding box sizes (relative to image size), where 0.1 indicates an object occupies 10\% of the image area. As shown in \autoref{fig:spatial_vs_localization} (b), VFMs surprisingly exhibit minimal performance differences when localizing large objects (0.3–0.5). Conversely, performance disparities amplify as object size decreases, revealing significant weaknesses in MiDas and SAM for smaller objects. More subgroup analyses for other AVAs are in the \autoref{-ss: results}.


\mypar{Niche mastery of a modest model} Interestingly, almost all VFMs, including those with generally lower performance, excel in at least one specific AVA (\autoref{fig: rank}). For example, SAM achieves exceptional results in color recognition, and DINOv2 excels notably in orientation.

\mypar{Consistently good VFM performance across lower and mid-level AVAs} All VFMs, regardless of their training strategies, demonstrate good performance in low- to mid-level AVAs such as texture recognition, relative depth estimation, and object recognition ( \autoref{fig: spider}). This uniformity implies that failures in complex visual reasoning predominantly arise from deficiencies in specific, critical AVAs rather than general shortcomings in visual understanding.

\mypar{Language-aligned pretraining boosts performance on language-centric AVAs} AVAs that involve understanding visual information intertwined with text, such as OCR, substantially benefit from language-aligned pretraining. Non-language-aligned VFMs significantly underperform in these tasks, as illustrated in \autoref{fig: spider}, highlighting the importance of language alignment in VFM training.

\mypar{The role of language supervision in VFM} Language-supervised VFMs, specifically SigLIP-1/2 and AIMv2, demonstrate broad competency across AVAs (\autoref{fig: rank} \& \autoref{fig: spider}). Their consistent high ranking highlights that language supervision is key to developing general-purpose visual abilities.

\section{Discussion}
\label{sec: discussion}




\textbf{Struggles of Non-Language-Aligned VFMs.} \autoref{fig: rank} and \autoref{fig: spider} indicate that non-language-aligned VFMs, despite exhibiting strengths in certain vision-centric or low-to-mid-level AVAs (e.g., DINOv2 in orientation and SAM in color), typically underperform in most AVAs. To investigate the underlying cause of these shortcomings, we conducted a preliminary study on visual feature representations before and after the LLM connector. Using linear probing on max-pooled visual features from DINOv2, we observed a substantial accuracy degradation from 66.3\% (pre-connector) to 25.67\% (post-connector) for the fine-grained recognition AVA, aligning with findings in previous work~\citep{finer}. This sharp performance drop suggests that critical visual information is often compromised during modality alignment processes. Recent studies have shown that fine-tuning the last few layers of VFMs can enhance performance~\citep{chen2024sharegpt4v}; however, this approach risks eroding the generalizability that initially makes these VFMs valuable. Alternatively, agglomerative models~\citep{ranzinger2024radio, heinrich2024radio} such as RADIO-2.1 demonstrate relatively robust performance on \ourbench, suggesting potential in combining specialized VFMs. Nevertheless, effectively aligning non-language-aligned VFMs to language modalities without sacrificing their inherent visual strengths remains a challenge, highlighting an important avenue for future research.


\noindent\textbf{Platonic Representation Hypothesis Holds?} Recent research suggests that large-scale training may lead VFMs toward converging onto similar representations--Platonic Representation Hypothesis~\citep{platonic}. Our findings partially support this hypothesis: for certain AVAs (\eg, object recognition, texture, and relative depth estimation), VFMs show universally strong performance, indicative of similar visual representations irrespective of training objectives. Conversely, significant performance disparities among VFMs in other AVAs indicate limitations to the generality of this hypothesis. Thus, our study suggests that this hypothesis might only hold in certain cases and warrants further empirical scrutiny in more diverse and challenging contexts.


\textbf{What Are Things Going From Here?} While MLLM have demonstrated remarkable versatility, they are not universally effective in all scenarios, especially in specialized domains~\citep{cheng2024domain, liang2024comprehensive}. Thus, there is a growing necessity for developing specialized MLLMs~\citep{kumar2024diagnostics, li2025openvision}. Currently, selecting appropriate VFMs for such customized MLLMs remains largely heuristic~\citep{tong2024cambrian, sun2025multi, li2025openvision}. Our work provides actionable insights that transform this selection process from heuristic guesswork into principled engineering. By clearly identifying AVA-specific strengths and weaknesses, practitioners can now systematically choose VFMs to precisely address the particular visual demands of targeted downstream tasks. Moreover, \ourbench represents a critical step towards developing next-generation VFMs by providing a systematic, diagnostic, and comprehensive evaluation framework. This benchmark enables VFM developers to accurately pinpoint specific deficiencies and implement targeted improvements, fostering the creation of more robust, versatile, and well-rounded VFMs in the future.


\newpage
\clearpage
\section*{Acknowledgments}
This research is supported by grants from the National Science Foundation (IIS-2107077, ICICLE: OAC-2112606). We are grateful for the support of the Ohio Supercomputer Center for providing computational resources.
{
    \small
    \bibliographystyle{ieeenat_fullname}
    \bibliography{main}

@String(CVPR= {IEEE Conf. Comput. Vis. Pattern Recog.})

@String(BMVC= {Brit. Mach. Vis. Conf.})

@String(ICASSP=	{ICASSP})

@String(ICLR = {Int. Conf. Learn. Represent.})

@String(AAAI = {AAAI})

@String(CVPR  = {CVPR})

@String(BMVC  =	{BMVC})

@String(ICLR  = {ICLR})

@misc{kth-tips_dataset,
title = { KTH-TIPS Dataset },
type = { Open Source Dataset },
author = { texture classification and segmentation },
howpublished = { \url{ https://universe.roboflow.com/texture-classification-and-segmentation/kth-tips } },
url = { https://universe.roboflow.com/texture-classification-and-segmentation/kth-tips },
journal = { Roboflow Universe },
publisher = { Roboflow },
year = { 2024 },
month = { feb },
note = { visited on 2025-05-16 },
}

@article{article,
author = {Mallikarjuna, P and Targhi, Alireza and Fritz, Mario and Hayman, Eric and Caputo, Barbara and Eklundh, J.-O},
year = {2006},
month = {07},
pages = {},
title = {THE KTH-TIPS2 database}
}

@article{zheng2023judging,
  title={Judging llm-as-a-judge with mt-bench and chatbot arena},
  author={Zheng, Lianmin and Chiang, Wei-Lin and Sheng, Ying and Zhuang, Siyuan and Wu, Zhanghao and Zhuang, Yonghao and Lin, Zi and Li, Zhuohan and Li, Dacheng and Xing, Eric and others},
  journal={Advances in Neural Information Processing Systems},
  volume={36},
  pages={46595--46623},
  year={2023}
}

@article{grattafiori2024llama,
  title={The llama 3 herd of models},
  author={Grattafiori, Aaron and Dubey, Abhimanyu and Jauhri, Abhinav and Pandey, Abhinav and Kadian, Abhishek and Al-Dahle, Ahmad and Letman, Aiesha and Mathur, Akhil and Schelten, Alan and Vaughan, Alex and others},
  journal={arXiv preprint arXiv:2407.21783},
  year={2024}
}

@article{espinosa2024there,
  title={There is no SAMantics! Exploring SAM as a Backbone for Visual Understanding Tasks},
  author={Espinosa, Miguel and Yang, Chenhongyi and Ericsson, Linus and McDonagh, Steven and Crowley, Elliot J},
  journal={arXiv preprint arXiv:2411.15288},
  year={2024}
}

@article{wu2025compact,
  title={COMPACT: COMPositional Atomic-to-Complex Visual Capability Tuning},
  author={Wu, Xindi and Hwang, Hee Seung and Kirichenko, Polina and Russakovsky, Olga},
  journal={arXiv preprint arXiv:2504.21850},
  year={2025}
}

@inproceedings{chae24decomposing,
  title={Decomposing Complex Visual Comprehension into Atomic Visual Skills for Vision Language Models},
  author={Chae, Hyunsik and Yoon, Seungwoo and Chun, Chloe Yewon and Go, Gyehun and Cho, Yongin and Lee, Gyeongmin and Ryu, Ernest K},
  booktitle={The 4th Workshop on Mathematical Reasoning and AI at NeurIPS'24}
}

@article{huang2023t2i,
  title={T2i-compbench: A comprehensive benchmark for open-world compositional text-to-image generation},
  author={Huang, Kaiyi and Sun, Kaiyue and Xie, Enze and Li, Zhenguo and Liu, Xihui},
  journal={Advances in Neural Information Processing Systems},
  volume={36},
  pages={78723--78747},
  year={2023}
}

@article{wu2024conceptmix,
  title={Conceptmix: A compositional image generation benchmark with controllable difficulty},
  author={Wu, Xindi and Yu, Dingli and Huang, Yangsibo and Russakovsky, Olga and Arora, Sanjeev},
  journal={arXiv preprint arXiv:2408.14339},
  year={2024}
}

@article{ainslie2023gqa,
  title={Gqa: Training generalized multi-query transformer models from multi-head checkpoints},
  author={Ainslie, Joshua and Lee-Thorp, James and De Jong, Michiel and Zemlyanskiy, Yury and Lebr{\'o}n, Federico and Sanghai, Sumit},
  journal={arXiv preprint arXiv:2305.13245},
  year={2023}
}

@online{RealWorldQA,
  author = {xAI},
  title = {grok},
  year = 2024,
  url = {https://x.ai/news/grok-1.5v},
  urldate = {2024}
}

@article{wah2011caltech,
  title={The caltech-ucsd birds-200-2011 dataset},
  author={Wah, Catherine and Branson, Steve and Welinder, Peter and Perona, Pietro and Belongie, Serge},
  year={2011},
  publisher={California Institute of Technology}
}

@article{li2019reliable,
  title={Reliable Crowdsourcing and Deep Locality-Preserving Learning for Unconstrained Facial Expression Recognition},
  author={Li, Shan and Deng, Weihong},
  journal={IEEE Transactions on Image Processing},
  volume={28},
  number={1},
  pages={356--370},
  year={2019},
  publisher={IEEE}
}

@article{monfort2019moments,
  title={Moments in time dataset: one million videos for event understanding},
  author={Monfort, Mathew and Andonian, Alex and Zhou, Bolei and Ramakrishnan, Kandan and Bargal, Sarah Adel and Yan, Tom and Brown, Lisa and Fan, Quanfu and Gutfreund, Dan and Vondrick, Carl and others},
  journal={IEEE transactions on pattern analysis and machine intelligence},
  volume={42},
  number={2},
  pages={502--508},
  year={2019},
  publisher={IEEE}
}

@article{geiger2013vision,
  title={Vision meets robotics: The kitti dataset},
  author={Geiger, Andreas and Lenz, Philip and Stiller, Christoph and Urtasun, Raquel},
  journal={The international journal of robotics research},
  volume={32},
  number={11},
  pages={1231--1237},
  year={2013},
  publisher={Sage Publications Sage UK: London, England}
}

@article{lian2020expression,
  title={Expression analysis based on face regions in real-world conditions},
  author={Lian, Zheng and Li, Ya and Tao, Jian-Hua and Huang, Jian and Niu, Ming-Yue},
  journal={International Journal of Automation and Computing},
  volume={17},
  pages={96--107},
  year={2020},
  publisher={Springer}
}

@article{liang2024comprehensive,
  title={A Comprehensive Survey and Guide to Multimodal Large Language Models in Vision-Language Tasks},
  author={Liang, Chia Xin and Tian, Pu and Yin, Caitlyn Heqi and Yua, Yao and An-Hou, Wei and Ming, Li and Wang, Tianyang and Bi, Ziqian and Liu, Ming},
  journal={arXiv preprint arXiv:2411.06284},
  year={2024}
}

@article{cheng2024domain,
  title={On Domain-Specific Post-Training for Multimodal Large Language Models},
  author={Cheng, Daixuan and Huang, Shaohan and Zhu, Ziyu and Zhang, Xintong and Zhao, Wayne Xin and Luan, Zhongzhi and Dai, Bo and Zhang, Zhenliang},
  journal={arXiv preprint arXiv:2411.19930},
  year={2024}
}

@inproceedings{kumar2024diagnostics,
  title={Diagnostics-LLaVA: A Visual Language Model for Domain-Specific Diagnostics of Equipment},
  author={Kumar, Aman and Alam, Mahbubul and Farahat, Ahmed and Somineni, Maheshjabu and Gupta, Chetan},
  booktitle={Annual Conference of the PHM Society},
  volume={16},
  number={1},
  year={2024}
}

@inproceedings{Temel2018_ICMLA,
author      = {D. Temel and J. Lee and G. AlRegib},
booktitle   = {2018 17th IEEE International Conference on Machine Learning and Applications (ICMLA)},
title       = {CURE-OR: Challenging unreal and real environments for object recognition},
year        = {2018},}

@inproceedings{silberman2012indoor,
  title={Indoor segmentation and support inference from rgbd images},
  author={Silberman, Nathan and Hoiem, Derek and Kohli, Pushmeet and Fergus, Rob},
  booktitle={Computer Vision--ECCV 2012: 12th European Conference on Computer Vision, Florence, Italy, October 7-13, 2012, Proceedings, Part V 12},
  pages={746--760},
  year={2012},
  organization={Springer}
}

@article{mai2024fine,
  title={Fine-tuning is fine, if calibrated},
  author={Mai, Zheda and Chowdhury, Arpita and Zhang, Ping and Tu, Cheng-Hao and Chen, Hong-You and Pahuja, Vardaan and Berger-Wolf, Tanya and Gao, Song and Stewart, Charles and Su, Yu and others},
  journal={Advances in Neural Information Processing Systems},
  volume={37},
  pages={136084--136119},
  year={2024}
}

@inproceedings{ranjan2021learning,
  title={Learning to count everything},
  author={Ranjan, Viresh and Sharma, Udbhav and Nguyen, Thu and Hoai, Minh},
  booktitle={Proceedings of the IEEE/CVF Conference on Computer Vision and Pattern Recognition},
  pages={3394--3403},
  year={2021}
}

@article{chowdhury2025prompt,
  title={Prompt-CAM: A Simpler Interpretable Transformer for Fine-Grained Analysis},
  author={Chowdhury, Arpita and Paul, Dipanjyoti and Mai, Zheda and Gu, Jianyang and Zhang, Ziheng and Mehrab, Kazi Sajeed and Campolongo, Elizabeth G and Rubenstein, Daniel and Stewart, Charles V and Karpatne, Anuj and others},
  journal={arXiv preprint arXiv:2501.09333},
  year={2025}
}

@book{kylberg2011kylberg,
  title={Kylberg texture dataset v. 1.0},
  author={Kylberg, Gustaf},
  year={2011},
  publisher={Centre for Image Analysis, Swedish University of Agricultural Sciences and~…}
}

@InProceedings{cimpoi14describing,
	      Author    = {M. Cimpoi and S. Maji and I. Kokkinos and S. Mohamed and and A. Vedaldi},
	      Title     = {Describing Textures in the Wild},
	      Booktitle = {Proceedings of the {IEEE} Conf. on Computer Vision and Pattern Recognition ({CVPR})},
	      Year      = {2014}}

@inproceedings{hsieh2017drone,
  title={Drone-based object counting by spatially regularized regional proposal network},
  author={Hsieh, Meng-Ru and Lin, Yen-Liang and Hsu, Winston H},
  booktitle={Proceedings of the IEEE international conference on computer vision},
  pages={4145--4153},
  year={2017}
}

@article{xia2017aid,
  title={AID: A benchmark data set for performance evaluation of aerial scene classification},
  author={Xia, Gui-Song and Hu, Jingwen and Hu, Fan and Shi, Baoguang and Bai, Xiang and Zhong, Yanfei and Zhang, Liangpei and Lu, Xiaoqiang},
  journal={IEEE Transactions on Geoscience and Remote Sensing},
  volume={55},
  number={7},
  pages={3965--3981},
  year={2017},
  publisher={IEEE}
}

@article{lomonaco2022cvpr,
  title={CVPR 2020 continual learning in computer vision competition: Approaches, results, current challenges and future directions},
  author={Lomonaco, Vincenzo and Pellegrini, Lorenzo and Rodriguez, Pau and Caccia, Massimo and She, Qi and Chen, Yu and Jodelet, Quentin and Wang, Ruiping and Mai, Zheda and Vazquez, David and others},
  journal={Artificial Intelligence},
  volume={303},
  pages={103635},
  year={2022},
  publisher={Elsevier}
}

@article{mai2020batch,
  title={Batch-level experience replay with review for continual learning},
  author={Mai, Zheda and Kim, Hyunwoo and Jeong, Jihwan and Sanner, Scott},
  journal={arXiv preprint arXiv:2007.05683},
  year={2020}
}

@inproceedings{shim2021online,
  title={Online class-incremental continual learning with adversarial shapley value},
  author={Shim, Dongsub and Mai, Zheda and Jeong, Jihwan and Sanner, Scott and Kim, Hyunwoo and Jang, Jongseong},
  booktitle={Proceedings of the AAAI Conference on Artificial Intelligence},
  volume={35},
  number={11},
  pages={9630--9638},
  year={2021}
}

@inproceedings{mai2021supervised,
  title={Supervised contrastive replay: Revisiting the nearest class mean classifier in online class-incremental continual learning},
  author={Mai, Zheda and Li, Ruiwen and Kim, Hyunwoo and Sanner, Scott},
  booktitle={Proceedings of the IEEE/CVF conference on computer vision and pattern recognition},
  pages={3589--3599},
  year={2021}
}

@article{mai2022online,
  title={Online continual learning in image classification: An empirical survey},
  author={Mai, Zheda and Li, Ruiwen and Jeong, Jihwan and Quispe, David and Kim, Hyunwoo and Sanner, Scott},
  journal={Neurocomputing},
  volume={469},
  pages={28--51},
  year={2022},
  publisher={Elsevier}
}

@inproceedings{xie2025efficiently,
  title={Efficiently Mitigating Video Content Misalignment on Large Vision Model with Time-Series Data Alignment},
  author={Xie, Hanchen and Ma, Rose and Zhu, Jiageng and Mai, Zheda and Abd-Almageed, Wael and Abraham, Zubin},
  booktitle={Proceedings of the Computer Vision and Pattern Recognition Conference},
  pages={3301--3307},
  year={2025}
}

@article{chen2023segment,
  title={Segment anything model (sam) enhanced pseudo labels for weakly supervised semantic segmentation},
  author={Chen, Tianle and Mai, Zheda and Li, Ruiwen and Chao, Wei-lun},
  journal={arXiv preprint arXiv:2305.05803},
  year={2023}
}

@inproceedings{houlsby2019parameter,
  title={Parameter-efficient transfer learning for NLP},
  author={Houlsby, Neil and Giurgiu, Andrei and Jastrzebski, Stanislaw and Morrone, Bruna and De Laroussilhe, Quentin and Gesmundo, Andrea and Attariyan, Mona and Gelly, Sylvain},
  booktitle={International conference on machine learning},
  pages={2790--2799},
  year={2019},
  organization={PMLR}
}

@inproceedings{tu2023visual,
  title={Visual query tuning: Towards effective usage of intermediate representations for parameter and memory efficient transfer learning},
  author={Tu, Cheng-Hao and Mai, Zheda and Chao, Wei-Lun},
  booktitle={Proceedings of the IEEE/CVF Conference on Computer Vision and Pattern Recognition},
  pages={7725--7735},
  year={2023}
}

@article{zhou2017places,
  title={Places: A 10 million Image Database for Scene Recognition},
  author={Zhou, Bolei and Lapedriza, Agata and Khosla, Aditya and Oliva, Aude and Torralba, Antonio},
  journal={IEEE Transactions on Pattern Analysis and Machine Intelligence},
  year={2017},
  publisher={IEEE}
}

@inproceedings{rezatofighi2019generalized,
  title={Generalized intersection over union: A metric and a loss for bounding box regression},
  author={Rezatofighi, Hamid and Tsoi, Nathan and Gwak, JunYoung and Sadeghian, Amir and Reid, Ian and Savarese, Silvio},
  booktitle={Proceedings of the IEEE/CVF conference on computer vision and pattern recognition},
  pages={658--666},
  year={2019}
}

@inproceedings{biten2019scene,
  title={Scene text visual question answering},
  author={Biten, Ali Furkan and Tito, Ruben and Mafla, Andres and Gomez, Lluis and Rusinol, Mar{\c{c}}al and Valveny, Ernest and Jawahar, CV and Karatzas, Dimosthenis},
  booktitle={Proceedings of the IEEE/CVF international conference on computer vision},
  pages={4291--4301},
  year={2019}
}

@article{kil2024mllm,
  title={Mllm-compbench: A comparative reasoning benchmark for multimodal llms},
  author={Kil, Jihyung and Mai, Zheda and Lee, Justin and Chowdhury, Arpita and Wang, Zihe and Cheng, Kerrie and Wang, Lemeng and Liu, Ye and Chao, Wei-Lun Harry},
  journal={Advances in Neural Information Processing Systems},
  volume={37},
  pages={28798--28827},
  year={2024}
}

@article{luo2001development,
  title={The development of the CIE 2000 colour-difference formula: CIEDE2000},
  author={Luo, M Ronnier and Cui, Guihua and Rigg, Bryan},
  journal={Color Research \& Application: Endorsed by Inter-Society Color Council, The Colour Group (Great Britain), Canadian Society for Color, Color Science Association of Japan, Dutch Society for the Study of Color, The Swedish Colour Centre Foundation, Colour Society of Australia, Centre Fran{\c{c}}ais de la Couleur},
  volume={26},
  number={5},
  pages={340--350},
  year={2001},
  publisher={Wiley Online Library}
}

@inproceedings{goyal2017making,
  title={Making the v in vqa matter: Elevating the role of image understanding in visual question answering},
  author={Goyal, Yash and Khot, Tejas and Summers-Stay, Douglas and Batra, Dhruv and Parikh, Devi},
  booktitle={Proceedings of the IEEE conference on computer vision and pattern recognition},
  pages={6904--6913},
  year={2017}
}

@article{awais2025foundation,
  title={Foundation Models Defining a New Era in Vision: a Survey and Outlook},
  author={Awais, Muhammad and Naseer, Muzammal and Khan, Salman and Anwer, Rao Muhammad and Cholakkal, Hisham and Shah, Mubarak and Yang, Ming-Hsuan and Khan, Fahad Shahbaz},
  journal={IEEE Transactions on Pattern Analysis and Machine Intelligence},
  year={2025},
  publisher={IEEE}
}

@article{liu2023visual,
  title={Visual instruction tuning},
  author={Liu, Haotian and Li, Chunyuan and Wu, Qingyang and Lee, Yong Jae},
  journal={Advances in neural information processing systems},
  volume={36},
  pages={34892--34916},
  year={2023}
}

@inproceedings{shao2019objects365,
  title={Objects365: A large-scale, high-quality dataset for object detection},
  author={Shao, Shuai and Li, Zeming and Zhang, Tianyuan and Peng, Chao and Yu, Gang and Zhang, Xiangyu and Li, Jing and Sun, Jian},
  booktitle={Proceedings of the IEEE/CVF international conference on computer vision},
  pages={8430--8439},
  year={2019}
}

@inproceedings{gupta2019lvis,
  title={Lvis: A dataset for large vocabulary instance segmentation},
  author={Gupta, Agrim and Dollar, Piotr and Girshick, Ross},
  booktitle={Proceedings of the IEEE/CVF conference on computer vision and pattern recognition},
  pages={5356--5364},
  year={2019}
}

@inproceedings{van2021benchmarking,
  title={Benchmarking representation learning for natural world image collections},
  author={Van Horn, Grant and Cole, Elijah and Beery, Sara and Wilber, Kimberly and Belongie, Serge and Mac Aodha, Oisin},
  booktitle={Proceedings of the IEEE/CVF conference on computer vision and pattern recognition},
  pages={12884--12893},
  year={2021}
}

@article{li2020object,
  title={Object detection in optical remote sensing images: A survey and a new benchmark},
  author={Li, Ke and Wan, Gang and Cheng, Gong and Meng, Liqiu and Han, Junwei},
  journal={ISPRS journal of photogrammetry and remote sensing},
  volume={159},
  pages={296--307},
  year={2020},
  publisher={Elsevier}
}

@inproceedings{mai2025lessons,
  title={Lessons and Insights from a Unifying Study of Parameter-Efficient Fine-Tuning (PEFT) in Visual Recognition},
  author={Mai, Zheda and Zhang, Ping and Tu, Cheng-Hao and Chen, Hong-You and Nguyen, Quang-Huy and Zhang, Li and Chao, Wei-Lun},
  booktitle={Proceedings of the Computer Vision and Pattern Recognition Conference},
  pages={14845--14857},
  year={2025}
}

@article{hu2022lora,
  title={Lora: Low-rank adaptation of large language models.},
  author={Hu, Edward J and Shen, Yelong and Wallis, Phillip and Allen-Zhu, Zeyuan and Li, Yuanzhi and Wang, Shean and Wang, Lu and Chen, Weizhu and others},
  journal={ICLR},
  volume={1},
  number={2},
  pages={3},
  year={2022}
}

@article{li2025openvision,
  title={OpenVision: A Fully-Open, Cost-Effective Family of Advanced Vision Encoders for Multimodal Learning},
  author={Li, Xianhang and Liu, Yanqing and Tu, Haoqin and Zhu, Hongru and Xie, Cihang},
  journal={arXiv preprint arXiv:2505.04601},
  year={2025}
}

@inproceedings{sun2025multi,
  title={Multi-Modal Large Language Models are Effective Vision Learners},
  author={Sun, Li and Ahuja, Chaitanya and Chen, Peng and D'Zmura, Matt and Batmanghelich, Kayhan and Bontrager, Philip},
  booktitle={2025 IEEE/CVF Winter Conference on Applications of Computer Vision (WACV)},
  pages={8617--8626},
  year={2025},
  organization={IEEE}
}

@article{dosovitskiy2020image,
  title={An image is worth 16x16 words: Transformers for image recognition at scale},
  author={Dosovitskiy, Alexey and Beyer, Lucas and Kolesnikov, Alexander and Weissenborn, Dirk and Zhai, Xiaohua and Unterthiner, Thomas and Dehghani, Mostafa and Minderer, Matthias and Heigold, Georg and Gelly, Sylvain and others},
  journal={arXiv preprint arXiv:2010.11929},
  year={2020}
}

@article{tong2024cambrian,
  title={Cambrian-1: A fully open, vision-centric exploration of multimodal llms},
  author={Tong, Peter and Brown, Ellis and Wu, Penghao and Woo, Sanghyun and IYER, Adithya Jairam Vedagiri and Akula, Sai Charitha and Yang, Shusheng and Yang, Jihan and Middepogu, Manoj and Wang, Ziteng and others},
  journal={Advances in Neural Information Processing Systems},
  volume={37},
  pages={87310--87356},
  year={2024}
}

@inproceedings{tong2024eyes,
  title={Eyes wide shut? exploring the visual shortcomings of multimodal llms},
  author={Tong, Shengbang and Liu, Zhuang and Zhai, Yuexiang and Ma, Yi and LeCun, Yann and Xie, Saining},
  booktitle={Proceedings of the IEEE/CVF Conference on Computer Vision and Pattern Recognition},
  pages={9568--9578},
  year={2024}
}

@article{zhang2024mme,
  title={MME-RealWorld: Could Your Multimodal LLM Challenge High-Resolution Real-World Scenarios that are Difficult for Humans?},
  author={Zhang, Yi-Fan and Zhang, Huanyu and Tian, Haochen and Fu, Chaoyou and Zhang, Shuangqing and Wu, Junfei and Li, Feng and Wang, Kun and Wen, Qingsong and Zhang, Zhang and others},
  journal={arXiv preprint arXiv:2408.13257},
  year={2024}
}

@article{yu2023mm,
  title={Mm-vet: Evaluating large multimodal models for integrated capabilities},
  author={Yu, Weihao and Yang, Zhengyuan and Li, Linjie and Wang, Jianfeng and Lin, Kevin and Liu, Zicheng and Wang, Xinchao and Wang, Lijuan},
  journal={arXiv preprint arXiv:2308.02490},
  year={2023}
}

@article{huang2024survey,
  title={A survey on evaluation of multimodal large language models},
  author={Huang, Jiaxing and Zhang, Jingyi},
  journal={arXiv preprint arXiv:2408.15769},
  year={2024}
}

@article{fini2024multimodal,
  title={Multimodal autoregressive pre-training of large vision encoders},
  author={Fini, Enrico and Shukor, Mustafa and Li, Xiujun and Dufter, Philipp and Klein, Michal and Haldimann, David and Aitharaju, Sai and da Costa, Victor Guilherme Turrisi and B{\'e}thune, Louis and Gan, Zhe and others},
  journal={arXiv preprint arXiv:2411.14402},
  year={2024}
}

@article{ranftl2020towards,
  title={Towards robust monocular depth estimation: Mixing datasets for zero-shot cross-dataset transfer},
  author={Ranftl, Ren{\'e} and Lasinger, Katrin and Hafner, David and Schindler, Konrad and Koltun, Vladlen},
  journal={IEEE transactions on pattern analysis and machine intelligence},
  volume={44},
  number={3},
  pages={1623--1637},
  year={2020},
  publisher={IEEE}
}

@article{heinrich2024radio,
  title={RADIO Amplified: Improved Baselines for Agglomerative Vision Foundation Models},
  author={Heinrich, Greg and Ranzinger, Mike and Lu, Yao and Kautz, Jan and Tao, Andrew and Catanzaro, Bryan and Molchanov, Pavlo and others},
  journal={arXiv preprint arXiv:2412.07679},
  year={2024}
}

@article{li2022video,
  title={Video Crowd Localization With Multifocus Gaussian Neighborhood Attention and a Large-Scale Benchmark},
  author={Li, Haopeng and Liu, Lingbo and Yang, Kunlin and Liu, Shinan and Gao, Junyu and Zhao, Bin and Zhang, Rui and Hou, Jun},
  journal={IEEE Transactions on Image Processing},
  volume={31},
  pages={6032--6047},
  year={2022},
  publisher={IEEE}
}

@article{maji2013fine,
  title={Fine-grained visual classification of aircraft},
  author={Maji, Subhransu and Rahtu, Esa and Kannala, Juho and Blaschko, Matthew and Vedaldi, Andrea},
  journal={arXiv preprint arXiv:1306.5151},
  year={2013}
}

@article{chen2024expanding,
  title={Expanding performance boundaries of open-source multimodal models with model, data, and test-time scaling},
  author={Chen, Zhe and Wang, Weiyun and Cao, Yue and Liu, Yangzhou and Gao, Zhangwei and Cui, Erfei and Zhu, Jinguo and Ye, Shenglong and Tian, Hao and Liu, Zhaoyang and others},
  journal={arXiv preprint arXiv:2412.05271},
  year={2024}
}

@inproceedings{ranzinger2024radio,
  title={Am-radio: Agglomerative vision foundation model reduce all domains into one},
  author={Ranzinger, Mike and Heinrich, Greg and Kautz, Jan and Molchanov, Pavlo},
  booktitle={Proceedings of the IEEE/CVF Conference on Computer Vision and Pattern Recognition},
  pages={12490--12500},
  year={2024}
}

@article{tschannen2025siglip,
  title={Siglip 2: Multilingual vision-language encoders with improved semantic understanding, localization, and dense features},
  author={Tschannen, Michael and Gritsenko, Alexey and Wang, Xiao and Naeem, Muhammad Ferjad and Alabdulmohsin, Ibrahim and Parthasarathy, Nikhil and Evans, Talfan and Beyer, Lucas and Xia, Ye and Mustafa, Basil and others},
  journal={arXiv preprint arXiv:2502.14786},
  year={2025}
}

@inproceedings{zhai2023sigmoid,
  title={Sigmoid loss for language image pre-training},
  author={Zhai, Xiaohua and Mustafa, Basil and Kolesnikov, Alexander and Beyer, Lucas},
  booktitle={Proceedings of the IEEE/CVF international conference on computer vision},
  pages={11975--11986},
  year={2023}
}

@article{zhu2023minigpt,
  title={Minigpt-4: Enhancing vision-language understanding with advanced large language models},
  author={Zhu, Deyao and Chen, Jun and Shen, Xiaoqian and Li, Xiang and Elhoseiny, Mohamed},
  journal={arXiv preprint arXiv:2304.10592},
  year={2023}
}

@article{chowdhery2023palm,
  title={Palm: Scaling language modeling with pathways},
  author={Chowdhery, Aakanksha and Narang, Sharan and Devlin, Jacob and Bosma, Maarten and Mishra, Gaurav and Roberts, Adam and Barham, Paul and Chung, Hyung Won and Sutton, Charles and Gehrmann, Sebastian and others},
  journal={Journal of Machine Learning Research},
  volume={24},
  number={240},
  pages={1--113},
  year={2023}
}

@article{han2022survey,
  title={A survey on vision transformer},
  author={Han, Kai and Wang, Yunhe and Chen, Hanting and Chen, Xinghao and Guo, Jianyuan and Liu, Zhenhua and Tang, Yehui and Xiao, An and Xu, Chunjing and Xu, Yixing and others},
  journal={IEEE transactions on pattern analysis and machine intelligence},
  volume={45},
  number={1},
  pages={87--110},
  year={2022},
  publisher={IEEE}
}

@article{thisanke2023semantic,
  title={Semantic segmentation using Vision Transformers: A survey},
  author={Thisanke, Hans and Deshan, Chamli and Chamith, Kavindu and Seneviratne, Sachith and Vidanaarachchi, Rajith and Herath, Damayanthi},
  journal={Engineering Applications of Artificial Intelligence},
  volume={126},
  pages={106669},
  year={2023},
  publisher={Elsevier}
}

@article{li2024llmcount,
  title={LLMCount: Enhancing Stationary mmWave Detection with Multimodal-LLM},
  author={Li, Boyan and Ding, Shengyi and Ma, Deen and Wu, Yixuan and Liao, Hongjie and Hu, Kaiyuan},
  journal={arXiv preprint arXiv:2409.16209},
  year={2024}
}

@inproceedings{yao2025countllm,
  title={Countllm: Towards generalizable repetitive action counting via large language model},
  author={Yao, Ziyu and Cheng, Xuxin and Huang, Zhiqi and Li, Lei},
  booktitle={Proceedings of the Computer Vision and Pattern Recognition Conference},
  pages={19143--19153},
  year={2025}
}

@article{zhang2025call,
  title={A Call for New Recipes to Enhance Spatial Reasoning in MLLMs},
  author={Zhang, Huanyu and Li, Chengzu and Wu, Wenshan and Mao, Shaoguang and Vuli{\'c}, Ivan and Zhang, Zhang and Wang, Liang and Tan, Tieniu and Wei, Furu and others},
  journal={arXiv preprint arXiv:2504.15037},
  year={2025}
}

@article{wu2025spatial,
  title={Spatial-MLLM: Boosting MLLM Capabilities in Visual-based Spatial Intelligence},
  author={Wu, Diankun and Liu, Fangfu and Hung, Yi-Hsin and Duan, Yueqi},
  journal={arXiv preprint arXiv:2505.23747},
  year={2025}
}

@inproceedings{caron2021emerging,
  title={Emerging properties in self-supervised vision transformers},
  author={Caron, Mathilde and Touvron, Hugo and Misra, Ishan and J{\'e}gou, Herv{\'e} and Mairal, Julien and Bojanowski, Piotr and Joulin, Armand},
  booktitle={Proceedings of the IEEE/CVF international conference on computer vision},
  pages={9650--9660},
  year={2021}
}

@article{naseer2021intriguing,
  title={Intriguing properties of vision transformers},
  author={Naseer, Muhammad Muzammal and Ranasinghe, Kanchana and Khan, Salman H and Hayat, Munawar and Shahbaz Khan, Fahad and Yang, Ming-Hsuan},
  journal={Advances in Neural Information Processing Systems},
  volume={34},
  pages={23296--23308},
  year={2021}
}

@article{goldblum2023battle,
  title={Battle of the backbones: A large-scale comparison of pretrained models across computer vision tasks},
  author={Goldblum, Micah and Souri, Hossein and Ni, Renkun and Shu, Manli and Prabhu, Viraj and Somepalli, Gowthami and Chattopadhyay, Prithvijit and Ibrahim, Mark and Bardes, Adrien and Hoffman, Judy and others},
  journal={Advances in Neural Information Processing Systems},
  volume={36},
  pages={29343--29371},
  year={2023}
}

@article{bommasani2021opportunities,
  title={On the opportunities and risks of foundation models},
  author={Bommasani, Rishi and Hudson, Drew A and Adeli, Ehsan and Altman, Russ and Arora, Simran and von Arx, Sydney and Bernstein, Michael S and Bohg, Jeannette and Bosselut, Antoine and Brunskill, Emma and others},
  journal={arXiv preprint arXiv:2108.07258},
  year={2021}
}

@article{khan2022transformers,
  title={Transformers in vision: A survey},
  author={Khan, Salman and Naseer, Muzammal and Hayat, Munawar and Zamir, Syed Waqas and Khan, Fahad Shahbaz and Shah, Mubarak},
  journal={ACM computing surveys (CSUR)},
  volume={54},
  number={10s},
  pages={1--41},
  year={2022},
  publisher={ACM New York, NY}
}

@inproceedings{wu2023multimodal,
  title={Multimodal large language models: A survey},
  author={Wu, Jiayang and Gan, Wensheng and Chen, Zefeng and Wan, Shicheng and Yu, Philip S},
  booktitle={2023 IEEE International Conference on Big Data (BigData)},
  pages={2247--2256},
  year={2023},
  organization={IEEE}
}

@article{zhang2024mm,
  title={Mm-llms: Recent advances in multimodal large language models},
  author={Zhang, Duzhen and Yu, Yahan and Dong, Jiahua and Li, Chenxing and Su, Dan and Chu, Chenhui and Yu, Dong},
  journal={arXiv preprint arXiv:2401.13601},
  year={2024}
}

@article{gu2025bioclip,
  title={BioCLIP 2: Emergent Properties from Scaling Hierarchical Contrastive Learning},
  author={Gu, Jianyang and Stevens, Samuel and Campolongo, Elizabeth G and Thompson, Matthew J and Zhang, Net and Wu, Jiaman and Kopanev, Andrei and Mai, Zheda and White, Alexander E and Balhoff, James and others},
  journal={arXiv preprint arXiv:2505.23883},
  year={2025}
}

@article{huang2025ocr,
  title={OCR-Reasoning Benchmark: Unveiling the True Capabilities of MLLMs in Complex Text-Rich Image Reasoning},
  author={Huang, Mingxin and Shi, Yongxin and Peng, Dezhi and Lai, Songxuan and Xie, Zecheng and Jin, Lianwen},
  journal={arXiv preprint arXiv:2505.17163},
  year={2025}
}

@article{wu2024visionllm,
  title={Visionllm v2: An end-to-end generalist multimodal large language model for hundreds of vision-language tasks},
  author={Wu, Jiannan and Zhong, Muyan and Xing, Sen and Lai, Zeqiang and Liu, Zhaoyang and Chen, Zhe and Wang, Wenhai and Zhu, Xizhou and Lu, Lewei and Lu, Tong and others},
  journal={Advances in Neural Information Processing Systems},
  volume={37},
  pages={69925--69975},
  year={2024}
}

@article{sapkota2025object,
  title={Object detection with multimodal large vision-language models: An in-depth review},
  author={Sapkota, Ranjan and Karkee, Manoj},
  journal={Available at SSRN 5233953},
  year={2025}
}

@inproceedings{fu2024blink,
  title={Blink: Multimodal large language models can see but not perceive},
  author={Fu, Xingyu and Hu, Yushi and Li, Bangzheng and Feng, Yu and Wang, Haoyu and Lin, Xudong and Roth, Dan and Smith, Noah A and Ma, Wei-Chiu and Krishna, Ranjay},
  booktitle={European Conference on Computer Vision},
  pages={148--166},
  year={2024},
  organization={Springer}
}

@article{szot2024grounding,
  title={Grounding multimodal large language models in actions},
  author={Szot, Andrew and Mazoure, Bogdan and Agrawal, Harsh and Hjelm, R Devon and Kira, Zsolt and Toshev, Alexander},
  journal={Advances in Neural Information Processing Systems},
  volume={37},
  pages={20198--20224},
  year={2024}
}

@article{wang2025harnessing,
  title={Harnessing Multi-modal Large Language Models for Measuring and Interpreting Color Differences},
  author={Wang, Zhihua and Long, Yu and Jiang, Qiuping and Huang, Chao and Cao, Xiaochun},
  journal={IEEE Transactions on Image Processing},
  year={2025},
  publisher={IEEE}
}

@article{tu2023holistic,
  title={Holistic transfer: Towards non-disruptive fine-tuning with partial target data},
  author={Tu, Cheng-Hao and Chen, Hong-You and Mai, Zheda and Zhong, Jike and Pahuja, Vardaan and Berger-Wolf, Tanya and Gao, Song and Stewart, Charles and Su, Yu and Chao, Wei-Lun Harry},
  journal={Advances in Neural Information Processing Systems},
  volume={36},
  pages={29149--29173},
  year={2023}
}

@article{eppel2025shape,
  title={Shape and texture recognition in large vision-language models},
  author={Eppel, Sagi and Bismut, Mor and Faktor, Alona},
  journal={arXiv preprint arXiv:2503.23062},
  year={2025}
}

@article{xia2024large,
  title={Large Language Models Can Understanding Depth from Monocular Images},
  author={Xia, Zhongyi and Wu, Tianzhao},
  journal={arXiv preprint arXiv:2409.01133},
  year={2024}
}

@inproceedings{mi2024hierarchical,
  title={Hierarchical Interpretable Vision Reasoning Driven Through a Multi-Modal Large Language Model for Depth Estimation},
  author={Mi, Wenfeng and Chen, He and Liu, Weipeng},
  booktitle={2024 China Automation Congress (CAC)},
  pages={829--834},
  year={2024},
  organization={IEEE}
}

@inproceedings{zhang2025vision,
  title={Vision-Language Embodiment for Monocular Depth Estimation},
  author={Zhang, Jinchang and Lu, Guoyu},
  booktitle={Proceedings of the Computer Vision and Pattern Recognition Conference},
  pages={29479--29489},
  year={2025}
}

@article{jung2024right,
  title={Is' Right'Right? Enhancing Object Orientation Understanding in Multimodal Language Models through Egocentric Instruction Tuning},
  author={Jung, Ji Hyeok and Kim, Eun Tae and Kim, Seo Yeon and Lee, Joo Ho and Kim, Bumsoo and Chang, Buru},
  journal={arXiv preprint arXiv:2411.16761},
  year={2024}
}

@inproceedings{yin2025multimodal,
  title={Do multimodal language models really understand direction? a benchmark for compass direction reasoning},
  author={Yin, Hang and Lin, Zhifeng and Liu, Xin and Sun, Bin and Li, Kan},
  booktitle={ICASSP 2025-2025 IEEE International Conference on Acoustics, Speech and Signal Processing (ICASSP)},
  pages={1--5},
  year={2025},
  organization={IEEE}
}

@article{gavrikov2024vision,
  title={Are Vision Language Models Texture or Shape Biased and Can We Steer Them?},
  author={Gavrikov, Paul and Lukasik, Jovita and Jung, Steffen and Geirhos, Robert and Lamm, Bianca and Mirza, Muhammad Jehanzeb and Keuper, Margret and Keuper, Janis},
  journal={arXiv preprint arXiv:2403.09193},
  year={2024}
}

@article{dai2025humanvlm,
  title={Humanvlm: Foundation for human-scene vision-language model},
  author={Dai, Dawei and Xu, Long and Li, Yutang and Zhang, Yuanhui and Xia, Shuyin},
  journal={Information Fusion},
  pages={103271},
  year={2025},
  publisher={Elsevier}
}

@article{fan2024mllm,
  title={MLLM-SUL: Multimodal Large Language Model for Semantic Scene Understanding and Localization in Traffic Scenarios},
  author={Fan, Jiaqi and Wu, Jianhua and Gao, Jincheng and Yu, Jianhao and Wang, Yafei and Chu, Hongqing and Gao, Bingzhao},
  journal={arXiv preprint arXiv:2412.19406},
  year={2024}
}

@article{chiu2024megacoin,
  title={MegaCOIN: Enhancing Medium-Grained Color Perception for Vision-Language Models},
  author={Chiu, Ming-Chang and Wen, Shicheng and Chen, Pin-Yu and Ma, Xuezhe},
  journal={arXiv preprint arXiv:2412.03927},
  year={2024}
}

@article{zhang2025finer,
  title={Finer-CAM: Spotting the Difference Reveals Finer Details for Visual Explanation},
  author={Zhang, Ziheng and Gu, Jianyang and Chowdhury, Arpita and Mai, Zheda and Carlyn, David and Berger-Wolf, Tanya and Su, Yu and Chao, Wei-Lun},
  journal={arXiv preprint arXiv:2501.11309},
  year={2025}
}

@article{li2024eald,
  title={Eald-mllm: Emotion analysis in long-sequential and de-identity videos with multi-modal large language model},
  author={Li, Deng and Liu, Xin and Xing, Bohao and Xia, Baiqiang and Zong, Yuan and Wen, Bihan and K{\"a}lvi{\"a}inen, Heikki},
  journal={arXiv preprint arXiv:2405.00574},
  year={2024}
}

@article{yang2024emollm,
  title={Emollm: Multimodal emotional understanding meets large language models},
  author={Yang, Qu and Ye, Mang and Du, Bo},
  journal={arXiv preprint arXiv:2406.16442},
  year={2024}
}

@article{zhang2025revisiting,
  title={Revisiting semi-supervised learning in the era of foundation models},
  author={Zhang, Ping and Mai, Zheda and Nguyen, Quang-Huy and Chao, Wei-Lun},
  journal={arXiv preprint arXiv:2503.09707},
  year={2025}
}

@article{liu2024ocrbench,
  title={OCRBench: on the hidden mystery of OCR in large multimodal models},
  author={Liu, Yuliang and Li, Zhang and Huang, Mingxin and Yang, Biao and Yu, Wenwen and Li, Chunyuan and Yin, Xu-Cheng and Liu, Cheng-Lin and Jin, Lianwen and Bai, Xiang},
  journal={Science China Information Sciences},
  volume={67},
  number={12},
  pages={220102},
  year={2024},
  publisher={Springer}
}

@inproceedings{kirillov2023segment,
  title={Segment anything},
  author={Kirillov, Alexander and Mintun, Eric and Ravi, Nikhila and Mao, Hanzi and Rolland, Chloe and Gustafson, Laura and Xiao, Tete and Whitehead, Spencer and Berg, Alexander C and Lo, Wan-Yen and others},
  booktitle={Proceedings of the IEEE/CVF international conference on computer vision},
  pages={4015--4026},
  year={2023}
}

@article{balachandran2024eureka,
  title={Eureka: Evaluating and understanding large foundation models},
  author={Balachandran, Vidhisha and Chen, Jingya and Joshi, Neel and Nushi, Besmira and Palangi, Hamid and Salinas, Eduardo and Vineet, Vibhav and Woffinden-Luey, James and Yousefi, Safoora},
  journal={arXiv preprint arXiv:2409.10566},
  year={2024}
}

@article{paul2023simple,
  title={A Simple Interpretable Transformer for Fine-Grained Image Classification and Analysis.},
  author={Paul, Dipanjyoti and Chowdhury, Arpita and Xiong, Xinqi and Chang, Feng-Ju and Carlyn, David and Stevens, Samuel and Provost, Kaiya and Karpatne, Anuj and Carstens, Bryan and Rubenstein, Daniel I and others},
  year={2023}
}

@inproceedings{radford2021learning,
  title={Learning transferable visual models from natural language supervision},
  author={Radford, Alec and Kim, Jong Wook and Hallacy, Chris and Ramesh, Aditya and Goh, Gabriel and Agarwal, Sandhini and Sastry, Girish and Askell, Amanda and Mishkin, Pamela and Clark, Jack and others},
  booktitle={International conference on machine learning},
  pages={8748--8763},
  year={2021},
  organization={PmLR}
}

@inproceedings{platonic,
  title={Position: The platonic representation hypothesis},
  author={Huh, Minyoung and Cheung, Brian and Wang, Tongzhou and Isola, Phillip},
  booktitle={Forty-first International Conference on Machine Learning},
  year={2024}
}

@inproceedings{finer,
  title={Finer: Investigating and Enhancing Fine-Grained Visual Concept Recognition in Large Vision Language Models},
  author={Kim, Jeonghwan and Ji, Heng},
  booktitle={Proceedings of the 2024 Conference on Empirical Methods in Natural Language Processing},
  pages={6187--6207},
  year={2024}
}

@inproceedings{chen2024sharegpt4v,
  title={Sharegpt4v: Improving large multi-modal models with better captions},
  author={Chen, Lin and Li, Jinsong and Dong, Xiaoyi and Zhang, Pan and He, Conghui and Wang, Jiaqi and Zhao, Feng and Lin, Dahua},
  booktitle={European Conference on Computer Vision},
  pages={370--387},
  year={2024},
  organization={Springer}
}

@article{oquab2023dinov2,
  title={Dinov2: Learning robust visual features without supervision},
  author={Oquab, Maxime and Darcet, Timoth{\'e}e and Moutakanni, Th{\'e}o and Vo, Huy and Szafraniec, Marc and Khalidov, Vasil and Fernandez, Pierre and Haziza, Daniel and Massa, Francisco and El-Nouby, Alaaeldin and others},
  journal={arXiv preprint arXiv:2304.07193},
  year={2023}
}

@inproceedings{balanced_vqa_v2,
author = {Yash Goyal and Tejas Khot and Douglas Summers{-}Stay and Dhruv Batra and Devi Parikh},
title = {Making the {V} in {VQA} Matter: Elevating the Role of Image Understanding in {V}isual {Q}uestion {A}nswering},
booktitle = {Conference on Computer Vision and Pattern Recognition (CVPR)},
year = {2017},
}

@inproceedings{MishraBMVC12,
  author    = "Mishra, A. and Alahari, K. and Jawahar, C.~V.",
  title     = "Scene Text Recognition using Higher Order Language Priors",
  booktitle = "BMVC",
  year      = "2012",
}

@misc{veit2016cocotextdatasetbenchmarktext,
      title={COCO-Text: Dataset and Benchmark for Text Detection and Recognition in Natural Images}, 
      author={Andreas Veit and Tomas Matera and Lukas Neumann and Jiri Matas and Serge Belongie},
      year={2016},
      eprint={1601.07140},
      archivePrefix={arXiv},
      primaryClass={cs.CV},
      url={https://arxiv.org/abs/1601.07140}, 
}
}

\clearpage
\setcounter{page}{1}
\maketitlesupplementary
\appendix
\section*{Appendix}

\noindent\textbf{Disclosure of LLM Usage.} Portions of this manuscript were polished for clarity and readability using an LLM. The LLM was not used to generate research ideas, design experiments, analyze data, or draw conclusions. All scientific content, methods, and results are the authors’ original work.

We provide details omitted in the main paper. 

\begin{itemize}
    \item  \autoref{-ss: ava-bench_details} : Example and curation details of each AVA of \ourbench
    \item \autoref{-ss: experiment_details} : Details of hyperparameter and metrics used in experiments
    \item \autoref{-ss: results}: Additional results and detailed analysis
    \item \autoref{-ss: vfm_details}: Detailed overview of VFMs
    \item \autoref{-ss: related}: Related work
    \item \autoref{-ss: eval_efficiency}: Evaluation of efficiency 
    \item \autoref{-ss: license} Dataset copyright/license
\end{itemize}

\section{\ourbench Details}
\label{-ss: ava-bench_details}

\begin{table*}[h!]
\small
\centering
\resizebox{\textwidth}{!}{%
  \begin{tabular}{l|p{5cm}|p{6cm}}
    \hline
    \noalign{\vskip 5pt}
    \textbf{Atomic Visual Ability} & \textbf{Definition} & \textbf{Example Question} \\ 
    \noalign{\vskip 5pt}\hline
    Counting       & Determining the number of instances of an object                   & How many apples are in the image? \\ \hline
    Localization   & Identifying the location of an object in the image                  & Provide bounding box coordinate for bicycle. \\ \hline
    Fine-Grained   & Differentiating between similar sub-categories of objects           & What species of fungi is in the image? \\ \hline
    OCR            & Reading and interpreting text visible in the image                  & What is written in the red bounding box in the image? \\ \hline
    Absolute Depth & Estimating how far an object is from the camera                     & From the camera's perspective, estimate how far the closest point of the car (red box) is from the camera in real-world distance, in meters. \\ \hline
    Relative Depth & Comparing distances of two objects from the camera    & Which object is closer to the camera, the car (red box) or the cyclist (blue box) to the camera? \\ 
    \hline
    Orientation & 	Determining the facing direction or angle of an object & What is the orientation of the toy bus in the image?\\
    \hline
    Spatial & Inferring layout and spatial relations & Considering the relative positions of two objects in the image, where is the bicycle (red box) located with respect to the towel (blue box)?\\
    \hline
    Object Recognition & Identifying objects present in the image given bounding box & What is in the red bounding box in the image?\\
    \hline
    Scene Recognition & Identifying the broader environment or type of setting & What is the scene class of the image?\\
    \hline
    Action Recognition & Determining what action is being performed & Which action or activity is shown in the image?\\
    \hline
    Texture & Describing surface appearance or material of objects & What is the texture attribute of image? \\
    \hline
    Color & Identifying colors of objects given bounding box & What color is shown within the bounding box?\\
    \hline
    Emotion & Recognizing emotional expressions in humans given bounding box & What emotion is being shown in the image?\\
    
    \hline
  \end{tabular}
}

\vskip10pt
\caption{%
  \textbf{Atomic Visual Abilities (AVAs).} We identify 14 AVAs, serving as the foundational capabilities that can be combined to tackle complex visual reasoning tasks. For each AVA, we provide the definition and an example question in \texttt{ourbench}.
}
\label{-tab:ava_bench_def}
\end{table*}

\subsection{{\textbf{\underline{A}}tomic \textbf{\underline{V}}isual \textbf{\underline{A}}bilities (\textbf{AVAs})}}
\label{-ss: ava}
As mentioned in Section 3.1, AVAs are elemental visual capabilities that can be combined to address more complex visual reasoning tasks. The definitions and representative questions for each AVA can be found in \autoref{-tab:ava_bench_def}. Additional qualitative illustrations are provided in Figure \ref{fig: ava_demo} and Figure \ref{fig: arch_and_ava} (b).

The 14 AVAs selected for \ourbench are grounded in a thorough literature analysis. 
\begin{enumerate}
    \item \textbf{Compositional text-to-image (T2I) benchmarks.} Studies on controllable generation motivate core visual primitives—number, colour, texture, object identity, spatial relations, and more—used to construct compositional prompts \citep{huang2023t2i, wu2024conceptmix}. These primitives form an initial pool of candidate abilities.
    \item \textbf{VQA question analysis.} We employ GPT-4 to summarize the visual skills demanded by VQA questions in various commonly-used datasets (VQAv2~\citep{goyal2017making}, RealWorldQA~\citep{RealWorldQA}, GQA~\citep{ainslie2023gqa}, etc.), thereby enriching the pool with abilities emphasized by real-world questions.
\end{enumerate}

\noindent Intersecting these two sources yields a concise yet crucial set of AVAs. Moreover, we focus strictly on pure perceptual tasks and exclude non-perceptual reasoning skills (e.g., historical context, mathematical reasoning). We provide more related work discussion in \autoref{-ss: related}.

\subsection{Dataset Curation}
\begin{figure*}
    \centering
    \includegraphics[width=1\linewidth]{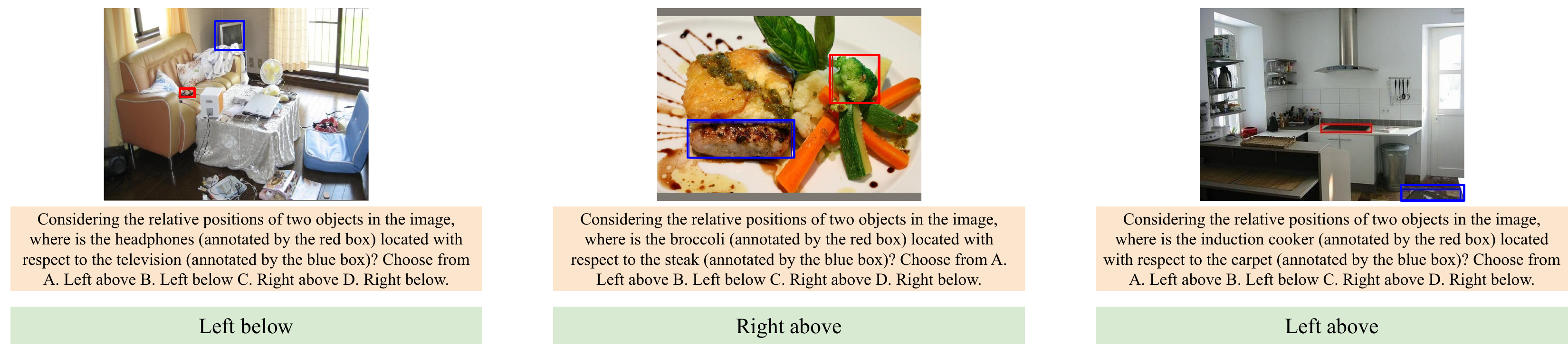}
    \caption{\small Examples of Spatial Reasoning AVA Samples. }
    \label{fig:append_spatial_AVA}
\end{figure*}

\mypar{Spatial Reasoning~\citep{wu2025spatial, zhang2025call}}
We curate \textbf{11.5K} image pairs from \textbf{NYU-Depth V2}~\citep{silberman2012indoor} (indoor scenes) and \textbf{LVIS}~\citep{gupta2019lvis} and \textbf{Objects365}~\citep{shao2019objects365} (open-domain scenes), all containing instance segmentation annotations. In each image, two distinct, non-overlapping objects are selected—one highlighted with a blue bounding box (reference object) and another with a red bounding box (target object). The model must identify the relative spatial position of the red box with respect to the blue box, choosing from four multiple-choice options: \emph{Left above}, \emph{Left below}, \emph{Right above}, and \emph{Right below} (\autoref{fig:append_spatial_AVA}). The preprocessing steps for dataset creation are summarized below:

\begin{itemize}
    \item To prevent ambiguity in interpretation, we restrict each image to contain only one instance of the target and reference objects.

    \item Object pairs whose bounding boxes overlapped either horizontally or vertically were excluded, ensuring unambiguous assignment to the four spatial categories.
    
    \item Extremely small bounding boxes complicating localization were filtered out. Specifically, object instances covering at least 2\% of the image area for NYU-Depth V2, and at least 0.2\% for LVIS and Objects365, were retained.
    
    \item For every question, we have a target and a reference object with a spatial position (the relative position of the target based on the reference object).  Each object class was ensured to appear in multiple spatial positions, with 40 samples per spatial position category. For each target object class in each spatial position, we ensure diversity of reference by selecting samples from 8 distinct reference classes and drawing 5 rows per reference, thereby preventing overfitting to a small set of co-occurring anchors.
    
    \item Each object class was ensured to appear in multiple spatial positions, with 40 samples per spatial position category. This prevents models from memorizing fixed layouts.
    
    \item For each target object class, reference object class, and spatial position category, an 80\% train and 20\% test split was ensured, for uniform distribution and fair evaluation.
    \item The question for each pair: \textit{``Considering the relative positions of two objects in the image, where is the microphone (annotated by the red box) located with respect to the speaker (annotated by the blue box)? Choose from A. Left above, B. Left below, C. Right above, D. Right below.''}
\end{itemize}

\begin{figure*}
    \centering
    \includegraphics[width=1\linewidth]{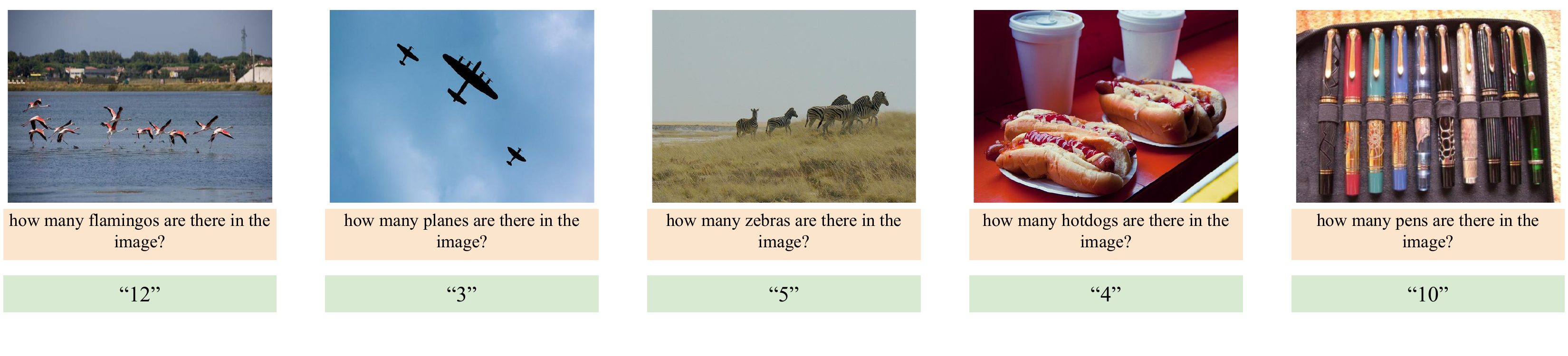}
    \caption{\small Examples of Counting AVA Samples. }
    \label{fig:append_counting_AVA}
\end{figure*}

\mypar{Counting~\citep{yao2025countllm, li2024llmcount}} 
We curate a total of \textbf{13.6K} images from five datasets— \textbf{VQAv2}~\citep{goyal2017making},  \textbf{FSC-147}~\citep{ranjan2021learning},  \textbf{CARPK}~\citep{hsieh2017drone},  \textbf{LVIS}~\citep{gupta2019lvis} , and  \textbf{CrowdHuman}~\citep{li2022video} —to evaluate object counting abilities across diverse domains, including open-domain scenes, natural objects, structured environments, and densely crowded contexts. Each image is paired with a question prompting the model to count the number of instances of a specified object category, and the model must return an integer-valued answer(\autoref{fig:append_counting_AVA}). The preprocessing steps for dataset creation are summarized below:

\begin{figure*}
    \centering
    \includegraphics[width=1\linewidth]{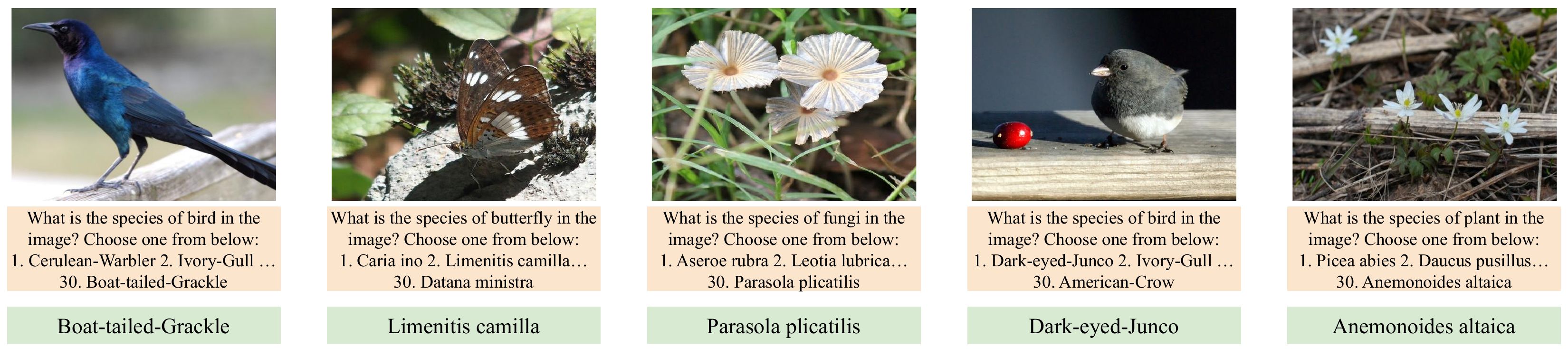}
    \caption{\small Examples of Fine-grained recognition AVA Samples. }
    \label{fig:append_fine_grained_AVA}
\end{figure*}

\begin{itemize}
    \item To ensure valid supervision, we filter all samples to retain only those with non-zero object counts and with object\_count~$\leq$~40.
    
    \item For each object category (object id), we require at least 4--5 distinct object count values to be represented, preventing overfitting to static object layouts.
    
    \item For each object count, we sample a fixed range of images per object\_id—between 6 and 30 depending on the dataset—to balance frequency and diversity.
    
    \item Dataset-specific sampling rules are applied:
    \begin{itemize}
        \item VQAv2 and LVIS: 15--30 images per object count; $\geq$5 count values per object id.
        \item FSC-147: 6--12 images per object count; $\geq$4 count values per object id.
        \item CARPK: 10--20 images per object count; $\geq$5 count values per object id.
        \item CrowdHuman: object count capped at 40; 25--50 images per count level.
    \end{itemize}
    
    \item An 80\% train / 20\% test split is maintained for each object id and count level to ensure balanced distribution during evaluation.
    
    \item The question for each image is: \textit{``How many [object] are there in the image?''}, where [object] refers to the annotated target category.
\end{itemize}

\begin{table*}[h!]
\centering
\resizebox{\textwidth}{!}{%
\begin{tabular}{lllll}
\hline
\noalign{\vskip 5pt}
Atomic Visual Abilities (AVA) & Dataset                    & Domain              & \# Train Samples              & \# Test Samples               \\
\noalign{\vskip 5pt}
\hline
Localization                  & Objects365~\citep{shao2019objects365}                 & Open                & 27.9K                         & 6.9K                          \\
                              & LVIS~\citep{gupta2019lvis}                       & Open                &                               &                               \\
                              & iNaturalist-2021~\citep{van2021benchmarking}           & Bird, Animal        &                               &                               \\
                              & DIOR~\citep{li2020object}                       & Remote-Sensing      &                               &                               \\
\hline
Counting                      & VQAv2~\citep{goyal2017making}                      & Open                & 10.8K                          & 2.8K                          \\
                              & FSC~\citep{ranjan2021learning}                        & Open                &                               &                               \\
                              & CARPK~\citep{hsieh2017drone}                      & Car                 &                               &                               \\
                              & Crowd Surveillance~\citep{li2022video} & People              &                               &                               \\
                              & LVIS~\citep{gupta2019lvis}                        & Open                &                               &                               \\
\hline
Fine-grained                  & CUB-200-2011~\citep{wah2011caltech}               & Bird                & 7.2K                          & 1.8K                          \\
                              & iNaturalist-2021~\citep{van2021benchmarking}           & Fungi,Plant, Animal &                               &                               \\
                              & FGVC-Aircraft~\citep{maji2013fine}              & Object              &                               &                               \\
\hline
Absolute Depth                & NYU-Depth V2~\citep{silberman2012indoor}               & Indoor Scene        & 6.8K                          & 1.8K                          \\
                              & KITTI~\citep{geiger2013vision}                      & Outdoor Scene       &                               &                               \\
\hline
Relative Depth                & NYU-Depth V2~\citep{silberman2012indoor}               & Indoor Scene        & 9.2K                          & 2.4K                          \\
                              & KITTI~\citep{geiger2013vision}                      & Outdoor Scene       &                               &                               \\
\hline
OCR                           & COCO-Text~\citep{veit2016cocotextdatasetbenchmarktext}                  & Open                & 8.8K                          & 2.2K                          \\
                              & IIIT5K ~\citep{MishraBMVC12}                     & Open                &                               &                               \\
                              & TextVQA~\citep{balanced_vqa_v2}                     & Open                &                               &                               \\
\hline
Orientation                   & EgoOrientBench~\citep{jung2024right}             & Open                & 6.9K                          & 1.6K                          \\
                              & CURE-OR~\citep{Temel2018_ICMLA}                    & Indoor              &                               &                               \\
\hline
Object Recognition            & Objects365~\citep{shao2019objects365}                 & Open                & 37.9K                         & 7K                          \\
                              & LVIS~\citep{gupta2019lvis}                       & Open                &                               &                               \\
                              & iNaturalist-2021~\citep{van2021benchmarking}           & Bird, Animal        &                               &                               \\
                              & DIOR~\citep{li2020object}                       & Remote-Sensing      &                               &                               \\
\hline
Action Recognition            & MiT~\citep{monfort2019moments}                        & Open                & 12K                         & 3K                          \\
\hline
Texture                       & DTD~\citep{cimpoi14describing}                        & Open                & 10.6K                         & 2.7K                          \\
                              & Kylberg~\citep{kylberg2011kylberg}                    & Open                &                               &                               \\
                              & KTH-TIPS~\citep{kth-tips_dataset}                   & Open                &                               &                               \\
                              & KTH-TIPS2~\citep{article}                   & Open                &                               &                               \\
\hline
Spatial Reasoning                      & Objects365~\citep{shao2019objects365}                 & Open                & 9.9K                         & 1.6K                          \\
                              & LVIS~\citep{gupta2019lvis}                       & Open                &                               &                               \\
                              & NYU-Depth V2~\citep{silberman2012indoor}               & Indoor Scene        &                               &                               \\
\hline
Scene Recognition             & Places434~\citep{zhou2017places}                  & Open                & 11.1K                         & 2.8K                          \\
                              & AID~\citep{xia2017aid}                         & Remote-Sensing      &                               &                               \\
\hline
Emotion                       & RAF-DB~\citep{li2019reliable}                      & Human               & 11.9K & 5.1K \\
                              & ExpW~\citep{lian2020expression}                       & Human               &                               &                               \\
\hline
Color                         & Objects365~\citep{shao2019objects365}                 & Open                & 11.2K & 2.8K\\
\hline
Total & - & - & 182.2K & 44.5K\\
\hline
\end{tabular}
}
\vskip10pt
\caption{\textbf{Detailed statistics of \ourbench.} }
\label{tab: ava_detail}
\end{table*}

\mypar{Fine-grained~\citep{paul2023simple, gu2025bioclip, zhang2025finer}}
We curate a total of \textbf{9K images} from five fine-grained recognition domains—\textbf{Bird}, \textbf{Animal}, \textbf{Fungi}, \textbf{Plant}, and \textbf{Object}—to assess species-level recognition capabilities. The dataset sources include CUB-200-2011~\citep{wah2011caltech} for birds, iNat21~\citep{van2021benchmarking}  for animals, fungi, and plants, and FGVC Aircraft~\citep{maji2013fine} for objects. Each sample contains an image and a question prompting the model to identify the specific species or object type(\autoref{fig:append_fine_grained_AVA}). Construction details are as follows:

\begin{itemize}
    \item We select 100 bird species from CUB-200-2011, and 50 random classes each from the Animal, Fungi, and Plant categories of iNat21, as well as 50 classes from FGVC Aircraft for the Object category. All random selections use a fixed random seed to ensure reproducibility.

    \item This results in 300 total object ids. For each class, we uniformly sample 30 images.

    \item For iNat21 entries, we format species names by retaining only the last two words of their taxonomic labels for clarity and consistency.
    \item For each object class, an 80\% training and 20\% testing split was established to ensure balanced per-class evaluation.

    \item For each multiple-choice question, the candidate list includes all 50 or 100 (Object only) class names used in that task split, ensuring consistent, closed-set evaluation.
    \item The question for each image is: \textit{``What species of bird is in the image? Choose one from below: 1. Cerulean\_Warbler, 2. American\_Crow, ..., 100. Pine\_Warbler ''}
\end{itemize}

\begin{figure*}
    \centering
    \includegraphics[width=1\linewidth]{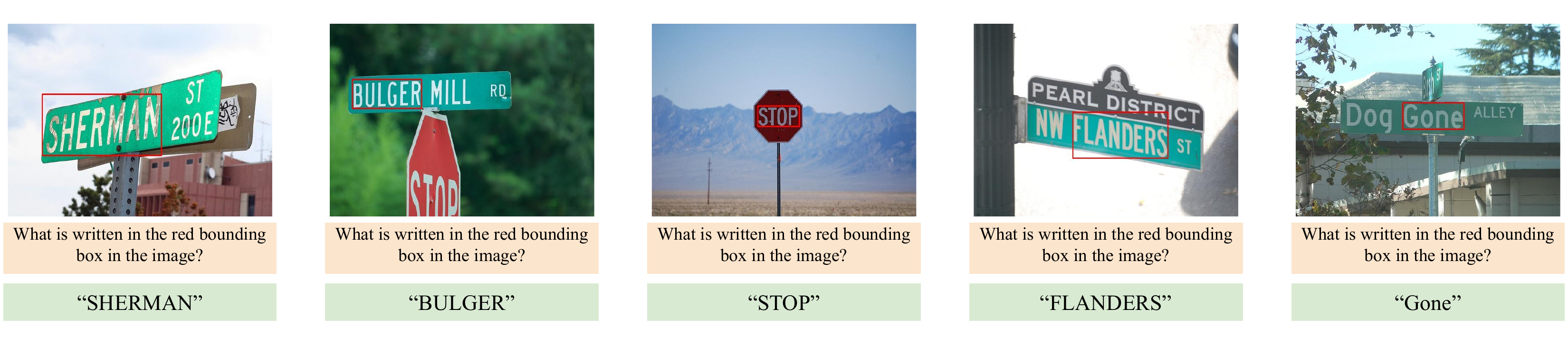}
    \caption{\small Examples of OCR AVA Samples. }
    \label{fig:append_ocr_AVA}
\end{figure*}

\mypar{OCR~\citep{huang2025ocr, liu2024ocrbench}}
We curate \textbf{10.9K} images from three OCR datasets—\textbf{COCO-Text}~\citep{veit2016cocotextdatasetbenchmarktext}, \textbf{IIIT5K}~\citep{MishraBMVC12}, and \textbf{TextVQA}~\citep{balanced_vqa_v2}—each containing word-level bounding box annotations. In each image, a red bounding box highlights the word to be transcribed. The model is prompted to recognize the textual content inside the box based on visual context(\autoref{fig:append_ocr_AVA}). The preprocessing steps for dataset creation are summarized below:

\begin{itemize}
    \item For COCO-Text, we retain only word-level boxes with an area greater than 1500 pixels to ensure sufficient visual resolution.

    \item For TextVQA, we apply a stricter filtering criterion by retaining only word boxes with area larger than 2000 pixels.

    \item For IIIT5K, we randomly sample 2,000 images from the original dataset without applying any area-based filtering.

    \item Each dataset is split into 80\% training and 20\% validation subsets individually, before merging the resulting splits to form the final OCR benchmark.
    \item A red bounding box is rendered on each image to highlight the target word location during inference.
    \item The question for each image: \textit{``What is written in the red bounding box in the image?''}
\end{itemize}

\begin{figure*}
    \centering
    \includegraphics[width=1\linewidth]{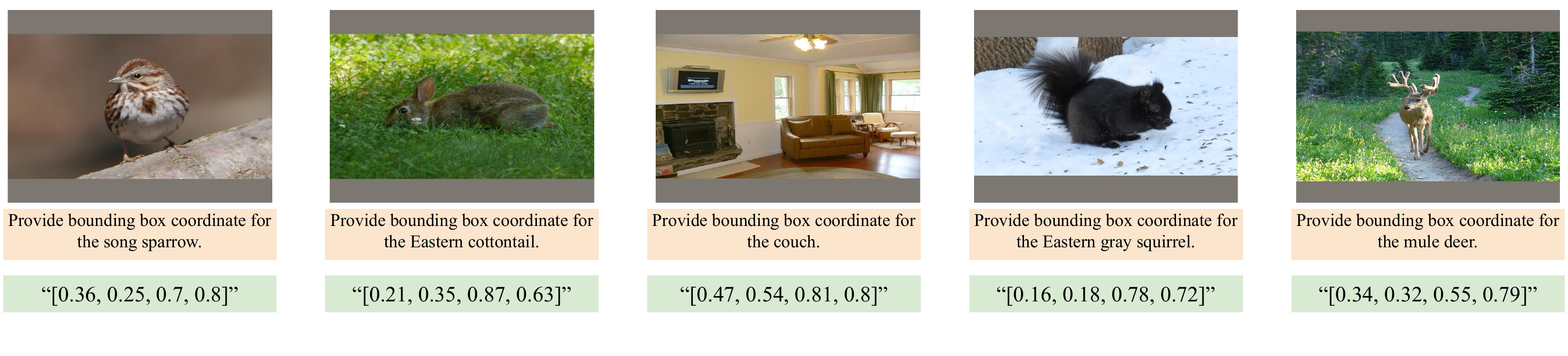}
    \caption{\small Examples of Localization AVA Samples. }
    \label{fig:append_localization_AVA}
\end{figure*}

\mypar{Localization~\citep{wu2024visionllm, sapkota2025object}}
We curate \textbf{34.8K} localization-focused image–question pairs sourced from \textbf{Objects365}~\citep{shao2019objects365} and \textbf{LVIS}~\citep{gupta2019lvis} (open-domain), \textbf{iNaturalist-2021}~\citep{van2021benchmarking} (birds and animals), and \textbf{DIOR}~\citep{li2020object} (remote-sensing). Each image contains a single object instance, and the model is prompted to identify its location by providing the bounding box coordinates(\autoref{fig:append_localization_AVA}). The preprocessing steps for dataset creation are summarized below:

\begin{itemize}
    \item We only retain object instances whose category appears exactly once in an image, to avoid ambiguity in localization supervision.

    \item We filter out objects with extremely small or large bounding boxes, retaining only those whose normalized area falls within the range of $0.002 < \text{area} < 0.5$ relative to the image.

    \item For \textbf{Objects365} and \textbf{LVIS}, we manually select 20 target object categories each. If the number of valid images for a category exceeds 700, we randomly sample 700 using a fixed seed for reproducibility.

    \item For \textbf{iNaturalist-2021}, we select 10 categories from the \texttt{aves} (birds) class and 10 from the \texttt{mammalia} (mammals) class, following the same filtering and sampling strategy. All scientific names are mapped to common names to improve interpretability and model alignment.

    \item For \textbf{DIOR}, we follow the same filtering steps and manually select 10 target categories, sampling up to 700 images per category as needed.

    \item All images are padded to square format using a consistent background color computed as the mean RGB value across multiple image processors:
    \[
    \text{background color} = RGB(124, 120, 111)
    \]
    The padding preserves content aspect ratio and ensures uniform input dimensions across models.

    \item For each object category, an 80\% training and 20\% testing split was performed after filtering and sampling, ensuring balanced and fair evaluation.

    \item The question for each image follows the format: \textit{``Provide bounding box coordinate for \textit{red-tailed hawk}.''} The object name is dynamically replaced depending on the image.
\end{itemize}

\begin{figure*}
    \centering
    \includegraphics[width=1\linewidth]{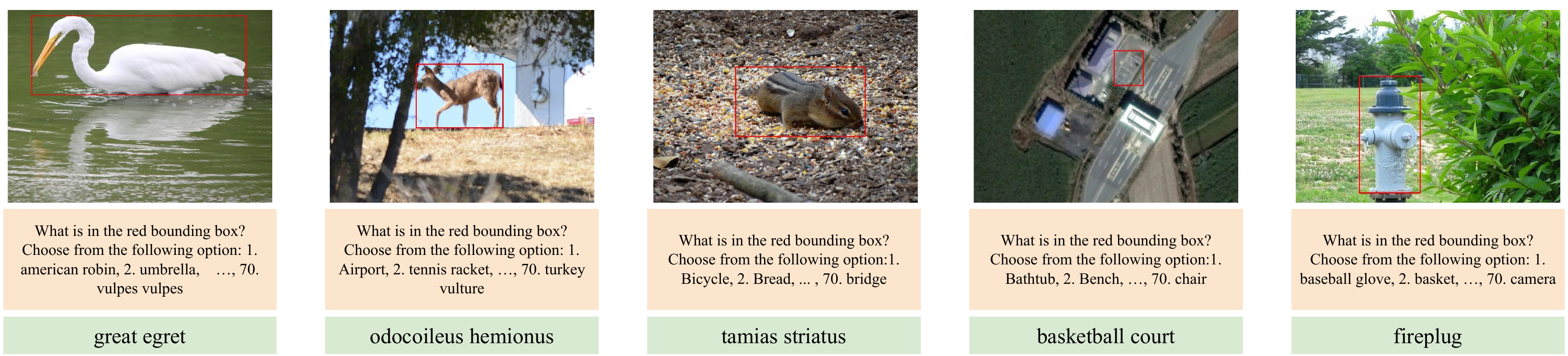}
    \caption{\small Examples of Recognition AVA Samples. }
    \label{fig:append_recognition_AVA}
\end{figure*}

\mypar{Recognition~\citep{fu2024blink, zhang2025revisiting, tu2023holistic}}
We curate \textbf{44.9K} recognition samples from four datasets spanning diverse visual domains—\textbf{Objects365}~\citep{shao2019objects365} and \textbf{LVIS}~\citep{gupta2019lvis} (open-domain objects), \textbf{iNaturalist-2021}~\citep{van2021benchmarking} (birds and animals), and \textbf{DIOR}~\citep{li2020object} (remote sensing). These samples are derived from the same images and object instances used in the localization benchmark. However, instead of asking for bounding box prediction, the recognition task requires the model to identify the object within a visually highlighted region (\autoref{fig:append_recognition_AVA}).

The preprocessing steps for dataset creation are summarized below:

\begin{itemize}
    \item We apply the same curation strategy as in localization: only one valid instance per image, with normalized bounding box area between $0.2\%$ and $50\%$ of the image. For each dataset, 10 or 20 object categories are manually selected.

    \item From \textbf{Objects365} and \textbf{LVIS}, we select 20 object categories each, and randomly sample up to 700 images per category (using a fixed seed for reproducibility).

    \item From \textbf{iNaturalist-2021}, we retain 10 species from the \texttt{Aves} (birds) and 10 from \texttt{Mammalia} (mammals) branches. Scientific names are converted to common English names for accessibility. Each species contributes up to 700 images.

    \item From \textbf{DIOR}, we select 10 object categories and apply the same filtering and sampling strategy (max 700 images per class).

    \item All images are padded to square shape using a consistent background color, computed from the average mean pixel values across nine vision-language processors, to ensure uniform input dimensions.
    \item For each object category, an 80\% training and 20\% testing split was performed after filtering and sampling, ensuring balanced and fair evaluation.

    \item \textbf{Unlike localization}, where bounding boxes are not rendered and must be predicted, in recognition the \textbf{red bounding box is explicitly drawn} onto each image to guide the model's attention.

    \item The question format is: \textit{``What is in the red bounding box? Choose from the following option: 1. airport, 2. american robin, ..., 70. vulpes vulpes''} The 70 object categories are shared across datasets and randomly shuffled for each question instance.
\end{itemize}

\mypar{Color~\citep{wang2025harnessing, chiu2024megacoin}}
We curate \textbf{14K} images from \textbf{Objects365}~\citep{shao2019objects365} and \textbf{LVIS}~\citep{gupta2019lvis}, each contributing 7K samples. This AVA focuses on assessing \textbf{color perception} in natural scenes. For each image, we extract a coherent color region using the following pipeline:

\begin{itemize}
    \item We apply SLIC superpixel segmentation and convert the image to LAB color space.

    \item Superpixels with similar color values are merged to form larger regions of consistent color.

    \item Among all candidate regions, we select the \textbf{top-1 region with the lowest internal color variance} as the final choice.

    \item A red bounding box is drawn on the selected region, and the most frequent RGB color within this region is used as the answer.
    
    \item For each object category, an 80\% training and 20\% testing split was performed after filtering and sampling, ensuring balanced and fair evaluation.

    \item The question format for each sample is: ``What color is shown within the red bounding box?''
\end{itemize}

\mypar{Action~\citep{szot2024grounding, wang2025harnessing, xie2025efficiently}}
We curate a total of \textbf{15K image–question pairs} from the \textbf{Moments in Time}~\citep{monfort2019moments} dataset, covering a wide range of human actions and activities. Each sample is derived from a short video clip annotated with a specific action label (\autoref{fig:append_action_AVA}). Construction details are as follows:

\begin{itemize}
    \item Similar action labels are merged into a unified category for clarity. We have 301 categories in total.  

    \item For each class, we randomly sample up to 40 training videos and 10 testing videos, ensuring broad yet balanced category coverage.

    \item The middle frame of each selected video is extracted and used as the image representing the associated action.

    \item Since listing all classes may exceed the token limits of many vision–language models, we randomly sample 100 action options per question, ensuring the ground-truth answer is always included.

    \item The question format for each sample is: ``Which action or activity is shown in the image? Choose from the following option: 1. buying, 2. catching, ..., 100. boxing''
\end{itemize}

\mypar{Emotion~\citep{yang2024emollm, li2024eald}}
We curate \textbf{17K image-question pairs} from two large-scale facial expression datasets: \textbf{RAF-DB}~\citep{li2019reliable} and \textbf{ExpW}~\citep{lian2020expression}. These datasets consist of human portraits labeled with one of seven basic emotions: \textit{happy}, \textit{sad}, \textit{angry}, \textit{fear}, \textit{surprise}, \textit{neutral}, and \textit{disgust}. Each image is annotated with a bounding box localizing the face of interest.

The preprocessing steps for dataset creation are summarized below:
\begin{itemize}
    \item Emotion labels across datasets were unified by consolidating synonymous terms (e.g., \texttt{happiness} and \texttt{happy}, \texttt{anger} and \texttt{angry}) to ensure consistent categorization across all samples.

    \item Bounding box annotations provided in the datasets were used to highlight the specific individual in multi-person scenes.

    \item An 80/20 train-test split was applied independently per emotion category to maintain class balance during evaluation.

    \item The question for each image is framed as: \textit{``Which of the following best describes the person's emotion in the red box? 1. happy, 2. sad, 3. angry, 4. fear, 5. surprise, 6. neutral, 7. disgust.''}
\end{itemize}

\mypar{Scene~\citep{dai2025humanvlm, fan2024mllm}}
We curate \textbf{13.9K image-question pairs} from two diverse datasets: \textbf{Places434}~\citep{zhou2017places} (open-domain) and \textbf{AID}~\citep{xia2017aid} (remote sensing). Each image is paired with a multiple-choice question, where the model selects the correct scene category from a pool of 30 randomly sampled options. The final set includes \textbf{463 unique scene classes} spanning a wide range of environments(\autoref{fig:append_scene_AVA}). The preprocessing steps for dataset creation are summarized below:

\begin{itemize}

\item GPT-4o was utilized to standardize labels across datasets by converting fine-grained labels into single, unified labels. Humans carefully checked each conversion to merge the newly converted labels conveying the same meaning with existing labels, ensuring semantic clarity and fluency.

\begin{figure*}
    \centering
    \includegraphics[width=1\linewidth]{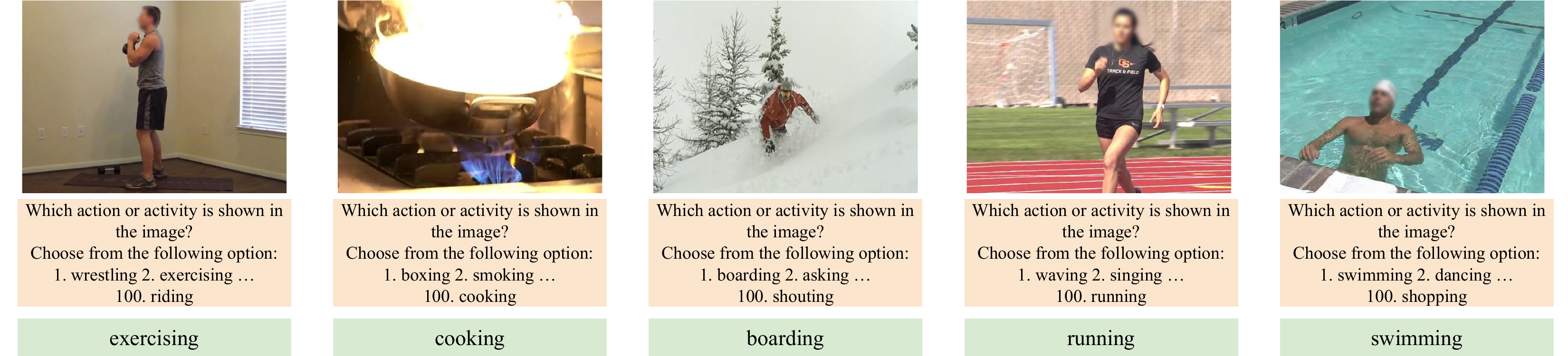}
    \caption{\small Examples of Action AVA Samples. }
    \label{fig:append_action_AVA}
\end{figure*}

\item A uniform distribution was maintained by extracting exactly 30 images per scene category, preventing category imbalance and ensuring consistent representation across classes. Classes with less than 30 images were discarded.

\item For each scene category within both datasets, an 80\% training and 20\% testing split was established, ensuring balanced and fair evaluation conditions.

\item The question for each pair: ``What is the scene class of the image? Choose one from below: 1. Entrance hall, 2. Lawn, ... 30. Snowy Mountain.'' These 30 options were selected by randomly sampling from the complete set of scene classes within each respective dataset, maintaining diversity and preventing predictable patterns.
\end{itemize}

\begin{figure*}
    \centering
    \includegraphics[width=1\linewidth]{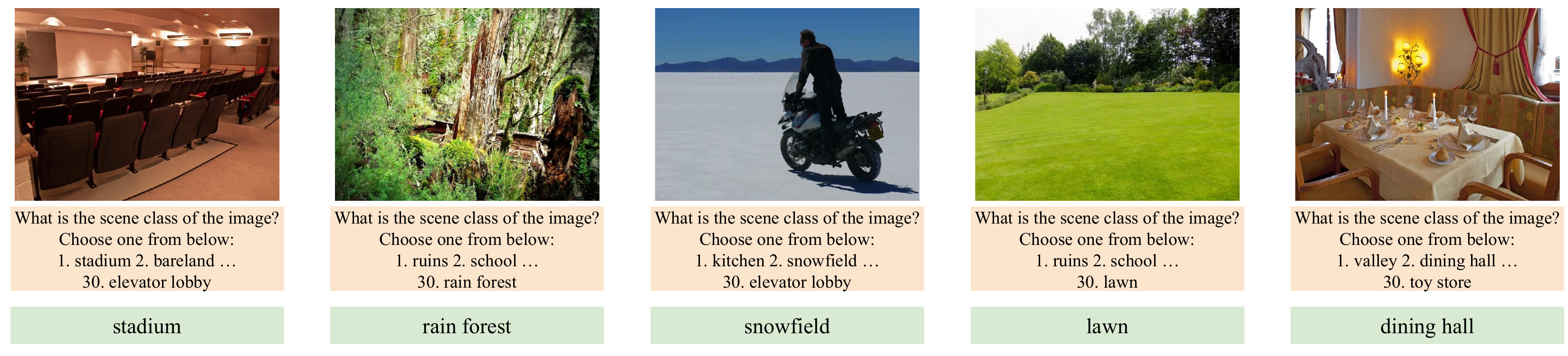}
    \caption{\small Examples of Scene AVA Samples. }
    \label{fig:append_scene_AVA}
\end{figure*}

\begin{figure*}
    \centering
    \includegraphics[width=1\linewidth]{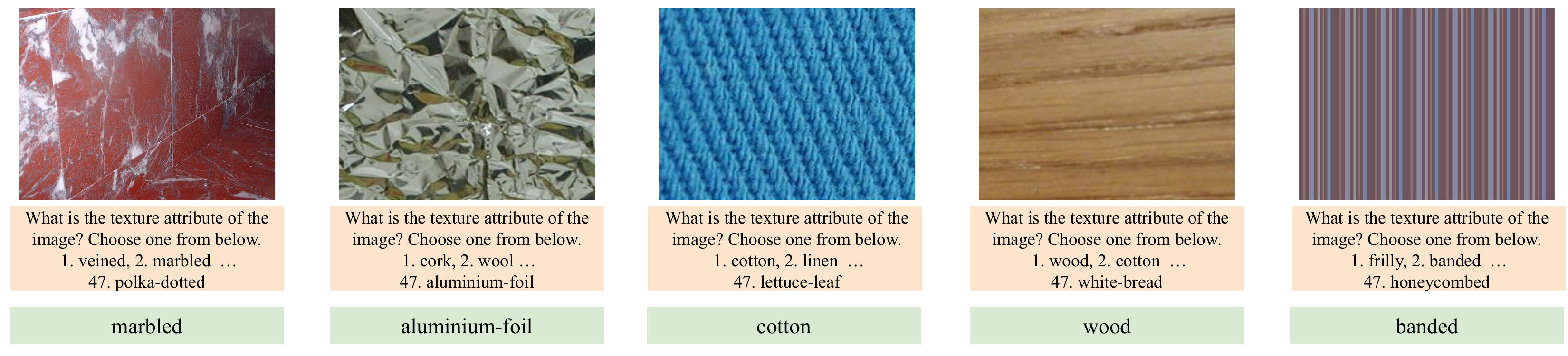}
    \caption{\small Examples of Texture AVA Samples. }
    \label{fig:append_texture_AVA}
\end{figure*}

\mypar{Texture~\citep{eppel2025shape, gavrikov2024vision}}
To assess a VFM’s ability to distinguish fine-grained visual patterns, we curate \textbf{13.2K image-question pairs} from diverse surface textures using close-up images from four open-domain datasets: \textbf{DTD}~\citep{cimpoi14describing}  , \textbf{Kylberg}~\citep{kylberg2011kylberg} , \textbf{KTH-TIPS}~\citep{kth-tips_dataset}   and \textbf{KTH-TIPS2-b}~\citep{article}. These datasets encompass a diverse array of texture types—such as \emph{striped}, \emph{aluminum foil}, and \emph{zigzagged}—capturing subtle visual patterns that are essential for accurate texture recognition. Each image is paired with a multiple-choice question, requiring the model to select the correct texture label from a set of options(\autoref{fig:append_texture_AVA}). The preprocessing steps for dataset creation are summarized below:

\begin{itemize}

\item Images where textures appeared as part of larger objects in cluttered scenes or within complex real-world photographs were discarded, to ensure that textures were clearly localized and recognizable without contextual interference.

\item Each texture attribute was represented by multiple images, with a minimum of 120 and a maximum of 480 samples per attribute, ensuring diversity and preventing memorization of fixed patterns by the models.

\item For each texture attribute, an 80\% train and 20\% test split was ensured, reaching uniform distribution and fair evaluation.

\item The question for each pair:\textit{``What is the texture attribute of the image? Choose one from below: 1. banded, 2. blotchy, …, 47. veined.''} The provided options exactly match the entire option pool from each respective dataset and were shuffled to avoid bias.
\end{itemize}

\begin{figure*}
    \centering
    \includegraphics[width=1\linewidth]{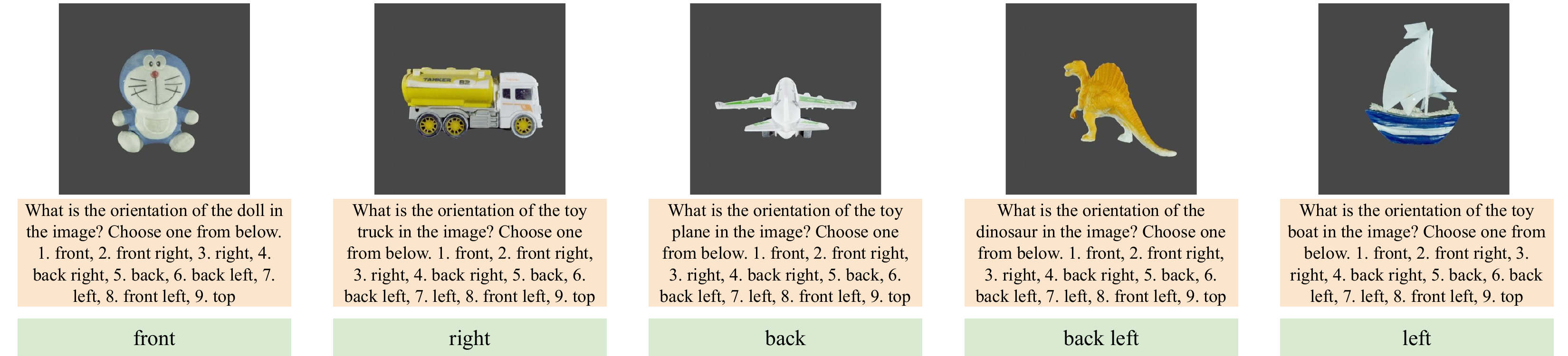}
    \caption{\small Examples of Orientation AVA Samples. }
    \label{fig:append_orientation_AVA}
\end{figure*}

\mypar{Orientation~\citep{yin2025multimodal, jung2024right}}
To evaluate viewpoint understanding, we curate \textbf{8.5K image-question pairs} from two specialized datasets: \textbf{CURE-OR}~\citep{Temel2018_ICMLA} and \textbf{EgoOrientBench}~\citep{jung2024right} . These datasets provide uncluttered images of objects captured from nine distinct orientations—\emph{front}, \emph{back}, \emph{left}, \emph{right}, \emph{top}, \emph{front left}, \emph{front right}, \emph{back left}, and \emph{back right}—allowing models to learn pose-specific cues without requiring bounding boxes(\autoref{fig:append_orientation_AVA}). The preprocessing steps for dataset creation are summarized below:
\begin{itemize}
\item For EgoOrientBench, each object class was ensured to appear in multiple orientations, with at least 10 and at most 40 samples per orientation label. This encourages models to learn generalized representations rather than memorizing specific arrangements.

\item From CURE-OR, we selected object instances photographed against two different background conditions using three distinct capture devices, ensuring variation in imaging style without compromising clarity.

\item  For each object, 80\% of the images from each orientation were assigned to the training set, and the remaining 20\% to the test set, ensuring balanced representation and fair evaluation across orientations.

\item The question for each pair: \textit{"What is the orientation of the toy plane in the image? Choose one from below: 1. front, 2. front right, 3. right, 4. back right, 5. back, 6. back left, 7. left, 8. front left, 9. top"}. These nine options represent the common orientation labels provided by both datasets. 
\end{itemize}

\begin{figure*}
    \centering
    \includegraphics[width=1\linewidth]{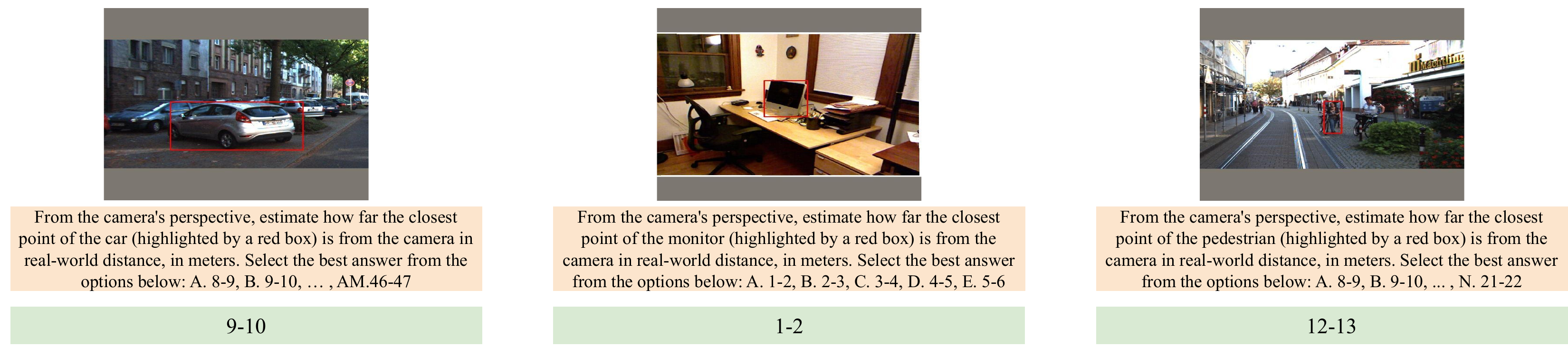}
    \caption{\small Examples of Absolute Depth AVA Samples. }
    \label{fig:append_abs_depth_AVA}
\end{figure*}

\mypar{Absolute Depth~\citep{xia2024large, mi2024hierarchical, zhang2025vision}}
We curate \textbf{9K} image-object pairs from \textbf{NYU-Depth V2}~\citep{silberman2012indoor} for indoor scenes and \textbf{KITTI}~\citep{geiger2013vision} for outdoor scenes. These datasets contain aligned RGB and depth information. Each image includes an object annotated with a bounding box, and the task requires estimating the absolute depth of the object in meters. This value is then matched against discretized ground-truth bins for evaluation(\autoref{fig:append_abs_depth_AVA}). 

The preprocessing steps for \textbf{indoor} dataset creation are summarized below:
\begin{itemize}
    \item Ambiguous object categories were excluded via a manually curated list of label IDs that often lack clear boundaries or meaningful depth interpretations (e.g., wall, floor, ceiling, etc.).
    \item Images were resized to a fixed resolution of $384\times384$, padding vertically as needed to preserve the original aspect ratio.
    \item A minimum bounding box area threshold was enforced after resizing all images. Specifically, we filtered out bounding boxes smaller than 500 pixels to ensure sufficient spatial resolution for the model.
    \item To ensure depth variation and avoid trivial samples, only object classes with at least 3 distinct depth bins (i.e., meaningful distribution over depth) were retained.
    \item For each label, depth bins were required to have a minimum of 10 and a maximum of 30 image samples. We removed bins with insufficient data to meet this requirement and capped those with excess samples by sorting instances based on bounding box area.
    
    \item After filtering, we retained 45 object classes, resulting in 4.4K unique image-object pairs. The dataset was split into 80\% train and 20\% test, preserving label and bin balance.

    \item The question for each pair: \textit{``From the camera's perspective, estimate how far the closest point of the cabinet (highlighted by a red box) is from the camera in real-world distance, in meters. Select the best answer from the options below: A. 1-2, B. 2-3, C. 3-4, D. 4-5, E. 5-6, F. 6-7''}.   
\end{itemize}

The preprocessing steps for \textbf{outdoor} dataset creation are summarized below:
\begin{itemize}
    \item To ensure the depth estimation task remains non-trivial, only objects with a minimum distance of 8 meters from the camera were considered. This avoids bias toward near-field predictions and better evaluates model precision in far-range perception.

    \item Depth values were discretized into bins, and for each class, we selected between 20 and 60 samples per bin to ensure coverage while avoiding overrepresentation. Bins with fewer than 20 samples were discarded. When bins exceeded 60 samples, selection was sorted by bounding box area to prioritize larger, more reliable objects.

    \item Image crops were extracted per object while maintaining the following aspect ratio constraint to preserve visual consistency: the width of the crop must be within the range $[\text{height}, 2\times\text{height}]$. This was enforced using the original image dimensions before padding or resizing.

    \item Objects touching any edge of the image were excluded to reduce the likelihood of partial occlusion or clipping.

    \item Images were padded vertically as needed to preserve the original aspect ratio.

    \item The final outdoor absolute depth set contains approximately 6K samples. For each object class and depth bin, an 80/20 train-test split was applied to maintain consistency in evaluation.
    
    \item The question for each outdoor sample: \textit{``Estimate the distance from the camera to the closest part of the cyclist (highlighted by a red box) in meters. Choose the best option: A. 8-9, B. 10-11,..., H. 30-31.''}
\end{itemize}

\begin{figure*}
    \centering
    \includegraphics[width=1\linewidth]{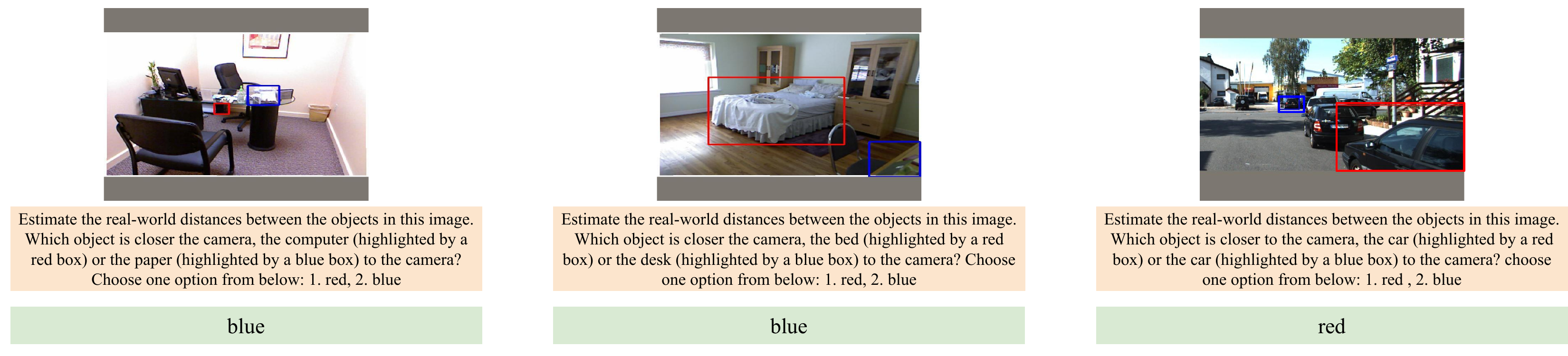}
    \caption{\small Examples of Relative Depth AVA Samples. }
    \label{fig:append_rel_depth_AVA}
\end{figure*}

\mypar{Relative Depth~\citep{xia2024large, mi2024hierarchical, zhang2025vision}}
We curate \textbf{11.6K} image-object pairs from \textbf{NYU-Depth V2}~\citep{silberman2012indoor} for indoor scenes and \textbf{KITTI}~\citep{geiger2013vision} for outdoor scenes, targeting the task of identifying which of two objects in an image is closer to the camera. Each image contains two distinct objects, each annotated with a bounding box. The model is asked to compare their absolute depth and choose the object that appears closer to the camera (\autoref{fig:append_rel_depth_AVA}). 

The preprocessing steps for \textbf{indoor} dataset creation are summarized below:
\begin{itemize}
    \item Ambiguous object categories were excluded via a manually curated list of label IDs that often lack clear boundaries or meaningful depth interpretations (e.g., wall, floor, ceiling, etc.).

    \item Only object pairs with valid bounding boxes (i.e., non-overlapping, fully inside image boundaries) were considered.

    \item To ensure perceptual clarity, candidate object pairs were filtered by requiring an absolute depth difference of at least \textbf{0.5 meters} between them. 

    \item For a given object class, only those with at least \textbf{10} valid pairings were retained, and a maximum of \textbf{30} total pairings per label were allowed.

    \item After filtering and sampling, we retained \textbf{131} object pairs across \textbf{5.7K} total questions. Images were padded vertically as needed to preserve their original aspect ratio. These were split into 80\% train and 20\% test sets while maintaining object label and depth-difference balance.

    \item Each question is posed as: \textit{``Estimate the real-world distances between the objects in this image. Which object is closer the camera, the sink (highlighted by a red box) or the towel (highlighted by a blue box) to the camera? Choose one option from below: 1. red, 2. blue''}.
\end{itemize}

The preprocessing steps for \textbf{outdoor} dataset creation are summarized below:
\begin{itemize}
    \item Object annotations were sourced for the following categories: \texttt{Car}, \texttt{Van}, \texttt{Pedestrian} (merged with \texttt{Person sitting}), and \texttt{Cyclist}.

    \item Pairs were formed using both intra-class (e.g., Car vs. Car) and inter-class (e.g., Car vs. Pedestrian) combinations.

    \item Pairs were retained only if the depth difference between the two objects was at least 0.5 meters, ensuring a meaningful perceptual gap.

    \item To avoid ambiguity and incomplete visual evidence, the following filters were applied:
    \begin{itemize}
        \item Pairs with occluded objects were excluded.
        \item Pairs where either object was touching the image edge were discarded.
        \item Only crops satisfying the aspect ratio constraint $\text{height} \leq \text{width} \leq 2 \times \text{height}$ were included, ensuring visual consistency.
        \item Images were padded vertically as needed to preserve their original aspect ratio.
    \end{itemize}

    \item After filtering, approximately 6K valid image-object pairs were retained. For each pair type, an \textbf{80/20 train-test split} was applied while maintaining distributional balance over object categories and depth separations.

    \item Each question is posed as: \textit{``Which object is closer to the camera, the van (highlighted in red) or the cyclist (highlighted in blue)? Choose one: 1. red, 2. blue.''}
\end{itemize}

\section{Experiment Details}
\label{-ss: experiment_details}

\subsection{Hyperparameter Details}
To ensure reproducibility and fairness, we carefully followed the official TinyLLaVA hyperparameter recommendations for stages 1 and 2, maintaining both the global batch size and learning rate as prescribed (in \autoref{tab:combined_hyperp}.). For stage 3, which incorporates LoRA-based fine-tuning, we selected a learning rate of 1e-4 and explored multiple LoRA dimensions (64, 128, 256). To validate these choices, we conducted preliminary experiments using three representative VFMs: DINOv2, CLIP, and SigLIP-2. We evaluated performance on two representative AVA tasks (OCR and Recognition), as summarized in \autoref{-tab:try_hyperparameter}.

The results consistently show that the recommended learning rate of 1e-4 yields stable and strong performance, whereas alternative learning rates often underperform or lead to instability. Similarly, LoRA dimensions between 64 and 128 produce comparable and reliable results, while extreme values show diminishing returns. Based on these observations, we adopt the recommended configuration (learning rate 1e-4, LoRA dimension 128) throughout our experiments. The overall hyperparameters of Stage-1 vision-language alignment pretraining, Stage-2 visual instruction
tuning and Stage-3 \ourbench evaluation are shown in \autoref{tab:combined_hyperp}.

\begin{table*}[]
\centering
\begin{tabular}{c|cccc|ccc }
\hline
{ \textbf{Task}}        & { \textbf{Model}} & { \textbf{lr 1e-5}} & { \textbf{lr 1e-4}} & { \textbf{lr 5e-4}} & { \textbf{LoRA 64}} & { \textbf{LoRA 128}} & { \textbf{LoRA 256}} \\
\hline
{\textbf{OCR}}         & { DINOv2}         & { 7.26}             & { 9.97}             & { 10.6}             & { 10.85}            & { 9.97}              & { 10.98}             \\
{ }                     & { SigLIP-2}       & { 79.68}            & { 81.18}            & { 77.63}            & { 81.51}            & { 81.18}             & { 81.25}             \\
{ }                     & { CLIP}           & { 54.23}            & { 60.44}            & { 60.9}             & { 60.79}            & { 60.44}             & { 61.73}             \\
{\textbf{Recognition}} & { DINOv2}         & { 83.92}            & { 86.31}            & { unstable}         & { 86.39}            & { 86.31}             & { 86.46}             \\
{ }                     & { SigLIP-2}       & { 87.04}            & { 88.19}            & { unstable}         & { 88.42}            & { 88.19}             & { 88.36}             \\
{ }                     & { CLIP}           & { 83.08}            & { 85.02}            & { unstable}         & { 84.79}            & { 85.02}             & { 84.18}  \\\hline          
\end{tabular}
\caption{Hyperparameter exploration for OCR and Recognition tasks using three representative VFMs. Results are reported across different learning rates and LoRA dimensions.}
\label{-tab:try_hyperparameter}
\end{table*}

\begin{table*}[h!]
\centering
\begin{tabular}{l|cc|c}
\hline
& \multicolumn{2}{c|}{\textbf{TinyLLaVa}} & \textbf{AVA-BENCH} \\
\cline{2-4}
Hyperparameter & Stage 1 & Stage 2 & Stage 3 \\
\hline
batch size & 16 & 4 & 4 \\
grad accum steps & 4 & 8 & 1 \\
LR & 1e-3 & 2e-5 & 1e-4 \\
LR schedule & \multicolumn{2}{c|}{cosine decay} & cosine decay \\
LR warmup ratio & \multicolumn{2}{c|}{0.03} & 0.03 \\
weight decay & \multicolumn{2}{c|}{0} & 0 \\
epoch & \multicolumn{2}{c|}{1} & 10 (20 for localization AVA) \\
optimizer & \multicolumn{2}{c|}{AdamW} & AdamW \\
DeepSpeed stage & \multicolumn{2}{c|}{3} & 3 \\
components finetuned & Connector & Connector + LLM & Connector + LoRA on LLM \\
sample size & 558K & 665K & -- \\
\hline
\end{tabular}
\vskip10pt
\caption{{Hyperparameters of TinyLLaVa and \ourbench Evaluation Stage}}
\label{tab:combined_hyperp}
\end{table*}

\subsection{Metric Details}
\paragraph{Color Recognition.} We use CIEDE2000~\citep{luo2001development} to calculate the color differences using the \texttt{colour} Python library. Specifically, we convert the predictions and ground-truths from CIE XYZ tristimulus format to CIE L*a*b* colour space with \texttt{colour.XYZ\_to\_Lab}, followed by the \texttt{colour.delta\_E(pred, gt, method="CIE 2000")} for color differences. 

\paragraph{Absolute Depth \& Counting. } We use the mean absolute error relative to the ground-truth (see \autoref{eq: ae_gt}). This normalization ensures that errors involving greater distances or counts, which are inherently more challenging, are proportionally penalized less severely.

\begin{equation}
\mathrm{MAE / GT}
\;=\;
\frac{1}{N}
\sum_{i=1}^{N}
\frac{\bigl|\,y_i - \hat y_i\,\bigr|}{y_i}.
\label{eq: ae_gt}
\end{equation}

where N is the number of test samples, $y_i$ is the ground‐truth (depth or count) for sample $i$ and $\hat y_i$ the model prediction.

\paragraph{Localization. } We use the Generalized Intersection-over-Union (GIoU~\citep{rezatofighi2019generalized}, \autoref{eq: giou}):
\begin{equation}
\mathrm{GIoU}(A,B)
=
\frac{|A \cap B|}{|A \cup B|}
\;-\;
\frac{\bigl|\,C \setminus (A \cup B)\bigr|}{|C|}\,.
\label{eq: giou}
\end{equation}

where $A$ and $B$ are the prediction and ground‐truth bounding boxes and $C$ is the smallest (axis–aligned) enclosing box of $A \cup B$.

\paragraph{OCR. } We evaluate OCR performance with Average Normalized Levenshtein Similarity (ANLS)~\citep{biten2019scene}:

\begin{equation}
\begin{aligned}
\mathrm{NLS}(p,g)
&= 1 - \frac{\mathrm{Lev}(p,g)}{\max(|p|,|g|)}, \\
\mathrm{ANLS}
&= \frac{1}{N}\sum_{i=1}^{N} \mathrm{NLS}(p_i,g_i) \\
&= \frac{1}{N}\sum_{i=1}^{N}
   \left(
     1 - \frac{\mathrm{Lev}(p_i,g_i)}{\max(|p_i|,|g_i|)}
   \right).
\end{aligned}
\end{equation}

where $\mathrm{Lev}(p,g)$ is the (Levenshtein) edit distance and $|.|$ denotes string length and N is the number of testing samples. 

\paragraph{Others. } All other AVAs employ standard accuracy metrics.

\section{More Results and Analysis}
\label{-ss: results}

\subsection{Detailed Overall Results}
The results used for plotting Figure 7 and Figure 8 are presented in \autoref{-tab: all} where the best performance for each is AVA is \textbf{bold} and the second best is in \textit{italics}. For each VFM, the first row is the performance, and the second row is the rank. 

\begin{table*}[]
\centering
\large
\resizebox{\textwidth}{!}{%
\begin{tabular}
{|c|c|c|c|c|c|c|c|c|c|c|c|c|c|c|c|}

\hline
AVA                           & Abs. Depth                                            & Rel. Depth                                       & OCR                                               & Counting                                              & Localization                                      & Object                                      & Fine-grained                                     & Scene                                            & Action                                           & Spatial                                          & Emotion                                          & Orientation                                      & Texture                                          & Color                                                    & \multirow{2}{*}{\makecell{Average \\ Ranking}} \\ \cline{1-15}
Metric & MAE/GT$\bm{\downarrow}$ & ACC$\bm{\uparrow}$ & ANLS$\bm{\uparrow}$ & MAE/GT$\bm{\downarrow}$ & GIOU$\bm{\uparrow}$ & ACC$\bm{\uparrow}$ & ACC$\bm{\uparrow}$ & ACC$\bm{\uparrow}$ & ACC$\bm{\uparrow}$ & ACC$\bm{\uparrow}$ & ACC$\bm{\uparrow}$ & ACC$\bm{\uparrow}$ & ACC$\bm{\uparrow}$ & CIEDE2000$\bm{\downarrow}$ & \\ \hline

\multirow{2}{*}{SigLIP-2}     & 0.0843                                                & \textit{97.38}                                   & \textbf{81.18}                                    & \textbf{0.225}                                        & \textbf{0.6738}                                   & \textbf{88.19}                                   & 90.17                                            & 72.60                                            & \textit{45.33}                                   & \textit{99.50}                                            & 58.39                                            & 79.2                                             & \textit{94.08}                                   & \textit{12.25}                                           & \multirow{2}{*}{2.4}             \\ \cline{2-15}
                              & 3                                                     & 2                                                & 1                                                 & 1                                                     & 1                                                 & 1                                                & 3                                                & 3                                                & 2                                                & 2                                                & 5                                                & 4                                                & 2                                                & 2                                                        &                                  \\ \hline
\multirow{2}{*}{AIMv2}        & 0.1008                                                & 95.71                                            & 62.44                                             & 0.254                                                 & 0.5896                                            & 85.04                                            & \textbf{91.78}                                   & \textit{72.86}                                   & 43.61                                            & 99.13                                            & \textbf{60.68}                                   & 79.71                                            & \textbf{94.11}                                   & 19.61                                                    & \multirow{2}{*}{3.9}             \\ \cline{2-15}
                              & 9                                                     & 8                                                & 3                                                 & 3                                                     & 5                                                 & 4                                                & 1                                                & 2                                                & 3                                                & 6                                                & 1                                                & 3                                                & 1                                                & 6                                                        &                                  \\ \hline
\multirow{2}{*}{SigLIP-1}     & 0.0953                                                & 96.58                                            & \textit{80.3}                                     & \textit{0.229}                                        & 0.6103                                            & \textit{87.84}                                   & \textit{90.94}                                   & \textbf{73.4}                                    & \textbf{45.76}                                   & 99.07                                            & 59.6                                             & 78.91                                            & 93.66                                            & 12.64                                                    & \multirow{2}{*}{3.6}             \\ \cline{2-15}
                              & 6                                                     & 7                                                & 2                                                 & 2                                                     & 4                                                 & 2                                                & 2                                                & 1                                                & 1                                                & 7                                                & 4                                                & 5                                                & 3                                                & 3                                                        &                                  \\ \hline
\multirow{2}{*}{CLIP}         & 0.08461                                               & 97.04                                            & 60.44                                             & 0.290                                                 & 0.5787                                            & 85.02                                            & 86.83                                            & 72.32                                            & 41.59                                            & 99.25                                            & \textit{60.42}                                   & 77.77                                            & 93.25                                            & 19.85                                                    & \multirow{2}{*}{4.9}             \\ \cline{2-15}
                              & 4                                                     & 4                                                & 6                                                 & 7                                                     & 7                                                 & 5                                                & 4                                                & 5                                                & 4                                                & 5                                                & 2                                                & 6                                                & 5                                                & 7                                                        &                                  \\ \hline
\multirow{2}{*}{InternVL-2.5} & \textit{0.08212}                                      & 97.00                                            & 60.88                                             & 0.269                                                 & 0.5850                                            & 83.19                                            & 72.83                                            & 71.63                                            & 36.13                                            & 99.5                                  & 59.99                                            & 75.71                                            & 91.21                                            & 20.09                                                    & \multirow{2}{*}{5.4}             \\ \cline{2-15}
                              & 2                                                     & 5                                                & 5                                                 & 5                                                     & 6                                                 & 7                                                & 7                                                & 6                                                & 7                                                & 4                                                & 3                                                & 7                                                & 7                                                & 8                                                        &                                  \\ \hline
\multirow{2}{*}{RADIOv2.1}    & \textbf{0.07645}                                      & \textbf{97.92}                                   & 62.44                                             & 0.257                                                 & \textit{0.6617}                                   & 84.90                                            & 85.44                                            & 72.50                                            & 38.77                                            & \textbf{99.69}                                   & 56.59                                            & \textit{83.71}                                   & 93.32                                            & 17.16                                                    & \multirow{2}{*}{3.6}             \\ \cline{2-15}
                              & 1                                                     & 1                                                & 4                                                 & 4                                                     & 2                                                 & 6                                                & 6                                                & 4                                                & 5                                                & 1                                                & 6                                                & 2                                                & 4                                                & 5                                                        &                                  \\ \hline
\multirow{2}{*}{DINOv2}       & 0.08469                                               & 97.25                                            & 9.97                                              & 0.272                                                 & 0.6598                                            & 86.31                                            & 85.5                                             & 70.99                                            & 37.45                                            & 99.50                                            & 54.44                                            & \textbf{85.54}                                   & 93.06                                            & 21.54                                                    & \multirow{2}{*}{5.3}             \\ \cline{2-15}
                              & 5                                                     & 3                                                & 7                                                 & 6                                                     & 3                                                 & 3                                                & 5                                                & 7                                                & 6                                                & 3                                                & 7                                                & 1                                                & 6                                                & 9                                                        &                                  \\ \hline
\multirow{2}{*}{SAM}          & 0.09792                                               & 94.13                                            & 9.79                                              & 0.313                                                 & 0.5216                                            & 76.68                                            & 40.06                                            & 58.28                                            & 17.25                                            & 90.22                                            & 36.88                                            & 69.37                                            & 81.02                                            & \textbf{9.87}                                            & \multirow{2}{*}{7.8}             \\ \cline{2-15}
                              & 8                                                     & 9                                                & 8                                                 & 8                                                     & 8                                                 & 8                                                & 8                                                & 9                                                & 9                                                & 8                                                & 9                                                & 8                                                & 9                                                & 1                                                        &                                  \\ \hline
\multirow{2}{*}{MiDas-3.0}    & 0.09563                                               & 96.63                                            & 7.72                                              & 0.336                                                 & 0.4490                                            & 75.05                                            & 32.83                                            & 60.19                                            & 18.25                                            & 53.58                                            & 40.40                                            & 67.37                                            & 86.38                                            & 13.28                                                    & \multirow{2}{*}{8}               \\ \cline{2-15}
                              & 7                                                     & 6                                                & 9                                                 & 9                                                     & 9                                                 & 9                                                & 9                                                & 8                                                & 8                                                & 9                                                & 8                                                & 9                                                & 8                                                & 4                                                        &                                  \\ \hline
\end{tabular}%
}
\vspace{10pt}
\caption{ The detailed overall results where the best performance for each is AVA is \textbf{bold} and the second best is in \textit{italics}. For each VFM, the first row is the performance, and the second row is the rank. Arrows indicate whether lower ($\downarrow$) or higher ($\uparrow$) is better.}
\label{-tab: all}
\end{table*}

\subsection{Detailed Analyses for Each AVA}
In the main results, we reported the overall performance of VFMs across various AVAs. However, aggregate metrics can sometimes obscure important nuanced insights. To gain deeper understanding, we conduct detailed analyses by partitioning test samples based on specific criteria (e.g., object size in localization tasks) and examining whether these subgroup trends align with overall performance. Generally, the detailed analyses affirm overall trends, but notable exceptions exist, particularly in localization.

\paragraph{Localization. } We split localization testing samples based on normalized bounding box sizes (relative to image size), where 0.1 indicates an object occupies 10\% of the image area. As illustrated in \autoref{fig:spatial_vs_localization} (b), VFMs surprisingly exhibit minimal performance differences when localizing large objects (0.3–0.5). Conversely, performance disparities amplify as object size decreases, revealing significant weaknesses in MiDas and SAM for smaller objects. Consequently, the lower overall performance of MiDas and SAM is predominantly due to poor handling of small targets. Practitioners should thus consider object size distributions when selecting VFMs; SAM and MiDas remain viable if target objects are predominantly large.

\begin{figure*}
    \centering
    \includegraphics[width=1\linewidth]{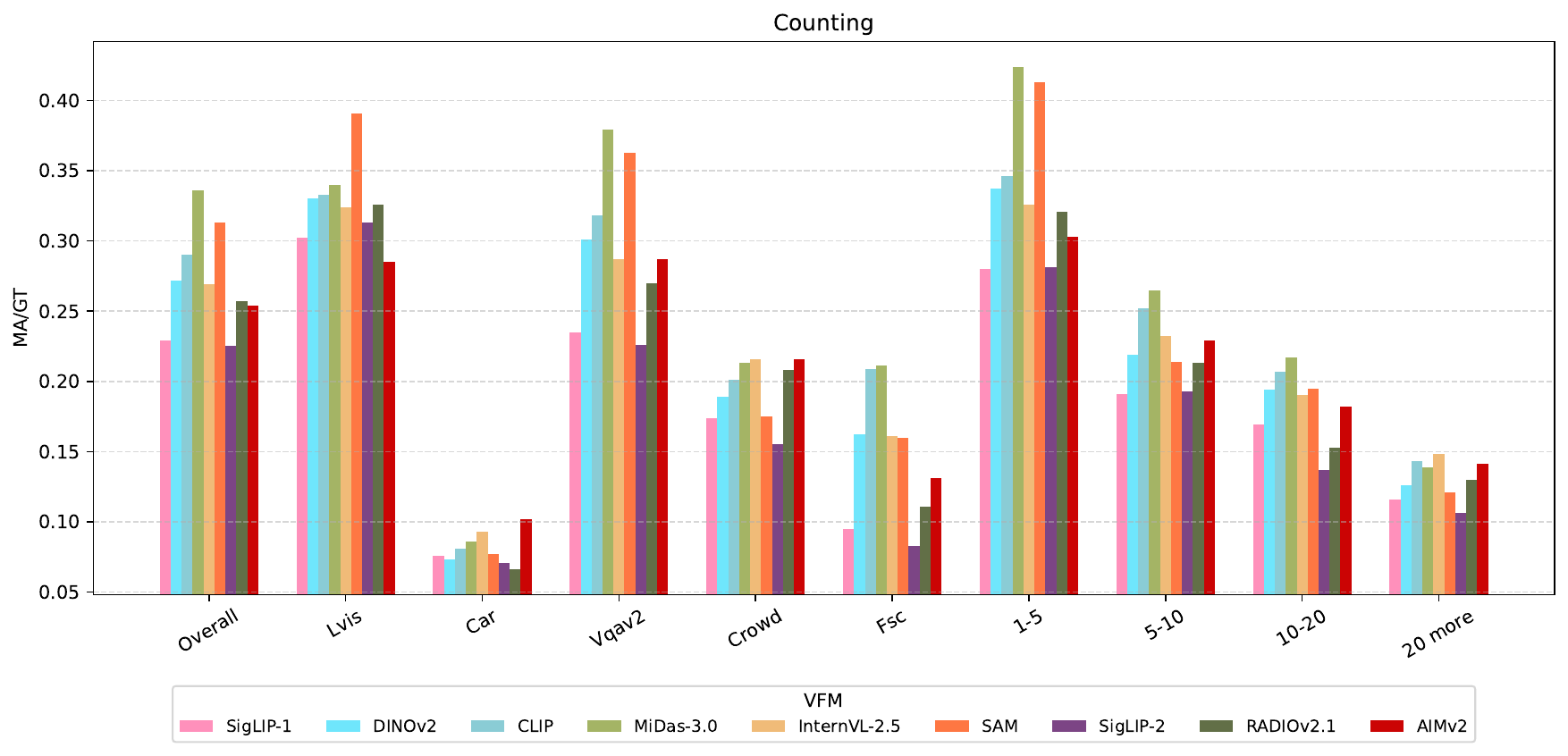}
    \caption{Detail results for counting for overall and different splits based on datasets and ground-truth count range. Lower MAE/GT is better. }
    \label{-fig: counting}
\end{figure*}

\paragraph{Counting. } Counting performance is generally consistent across different datasets and count ranges (\autoref{-fig: counting}). A notable exception is SAM, whose accuracy notably improves in denser scenarios.

\begin{figure*}
    \centering
    \includegraphics[width=1\linewidth]{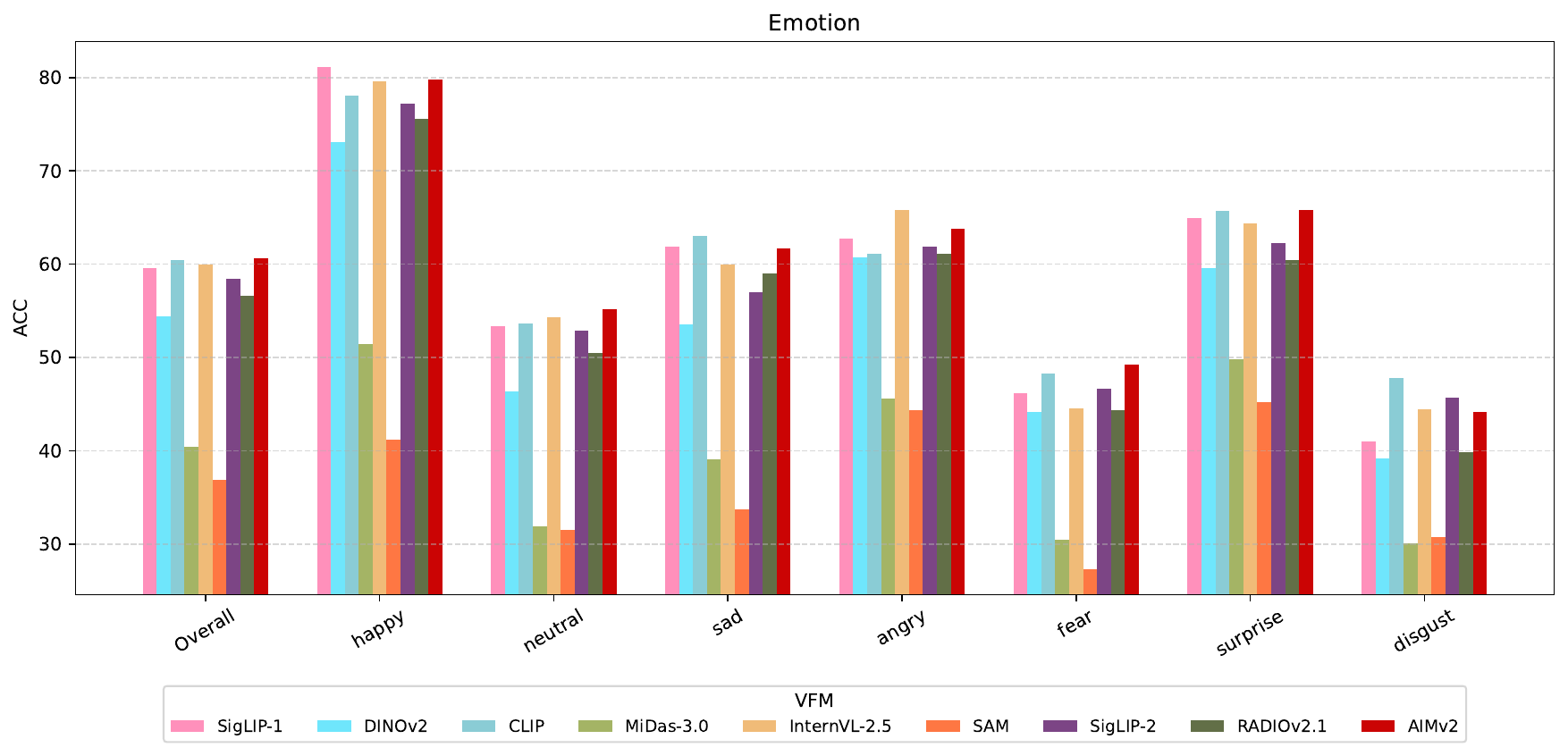}
    \caption{Detail results for emotion for overall and different splits based on emotion types.}
    \label{-fig: emotion}
\end{figure*}

\paragraph{Emotion.} Emotion recognition results exhibit remarkable consistency, with rankings and relative performances highly stable across emotion categories (see \autoref{-fig: emotion}).

\begin{figure*}
    \centering
    \includegraphics[width=1\linewidth]{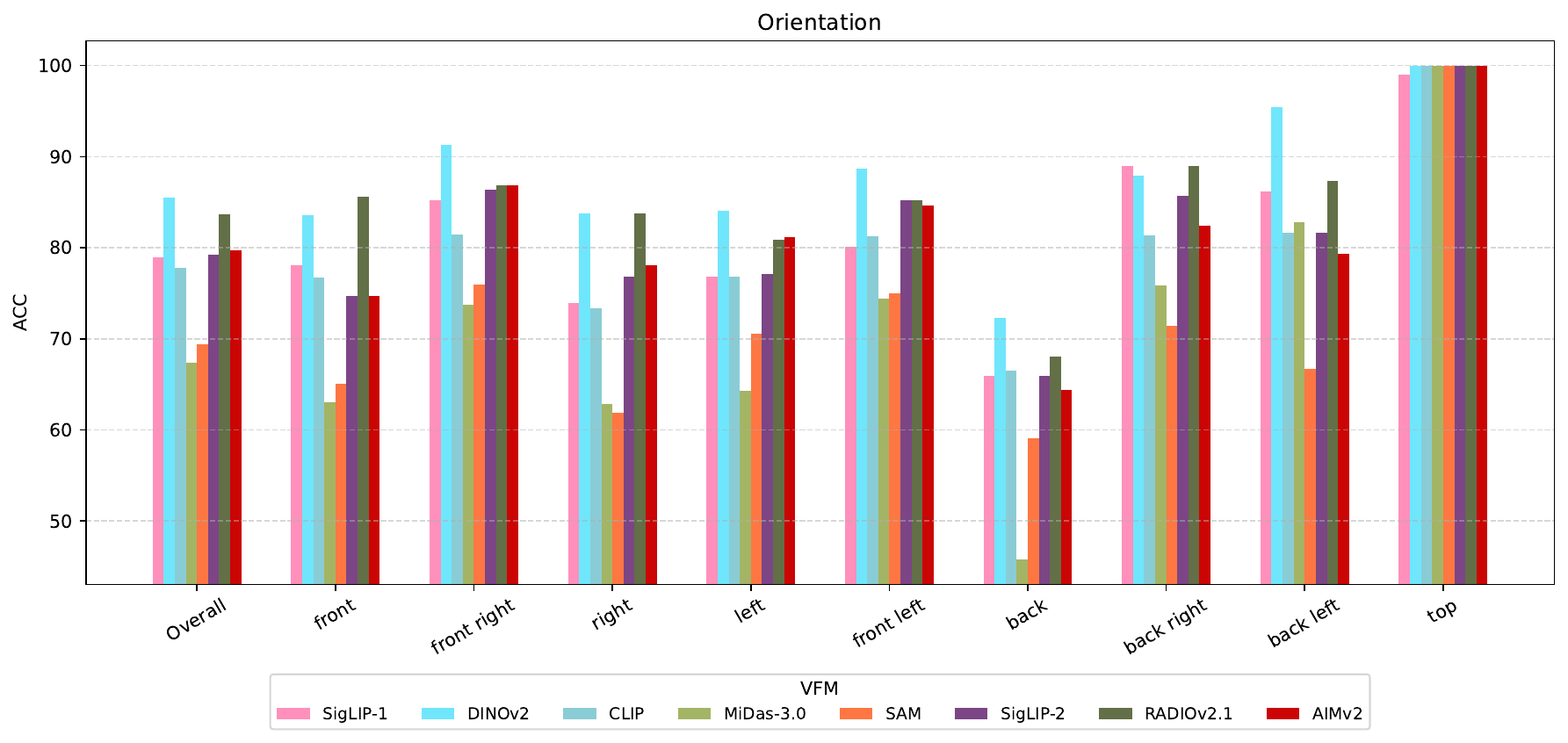}
    \caption{Detail results for orientation for overall and different splits based on viewpoint directions.}
    \label{-fig: orientation}
\end{figure*}

\paragraph{Orientation.} Orientation performance remains consistent overall, with some intriguing exceptions. Specifically, VFMs universally achieve near-perfect accuracy for top-view images, presumably due to the distinctive nature of this viewpoint compared to side or frontal views (see \autoref{-fig: orientation}).

\begin{figure*}
    \centering
    \includegraphics[width=1\linewidth]{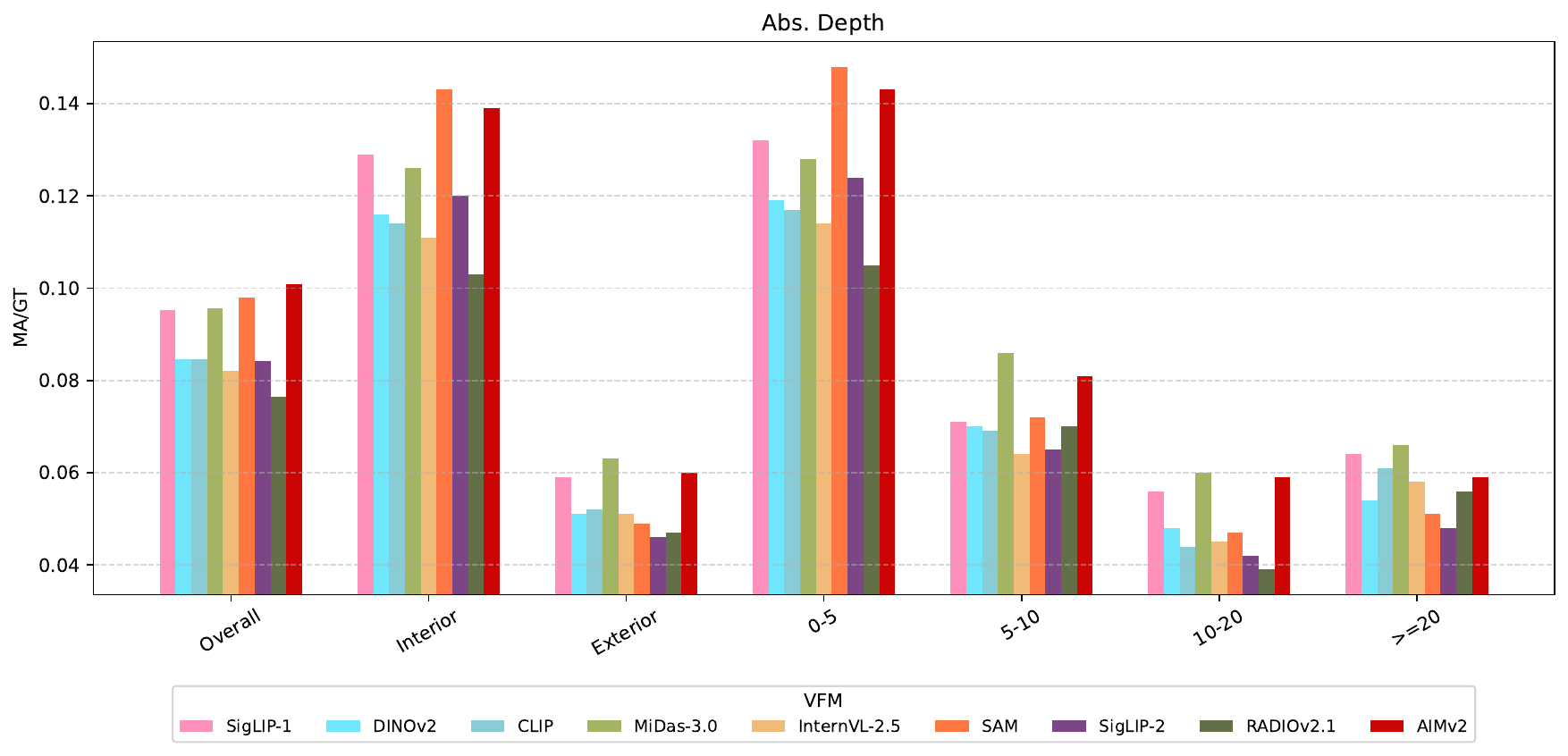}
    \caption{Detail results for absolute depth for overall and different splits based on scene type and object count range. Lower MAE/GT is better.}
    \label{-fig: absolute depth}
\end{figure*}

\paragraph{Absolute Depth.} Overall, absolute depth performance is stable, though specific VFMs exhibit distinctive trends. SAM notably struggles with near objects but improves significantly with increased distance. Conversely, RADIO demonstrates an opposite pattern, excelling with nearer objects but deteriorating with greater distances (see \autoref{-fig: absolute depth}).

\begin{figure*}
    \centering
    \includegraphics[width=1\linewidth]{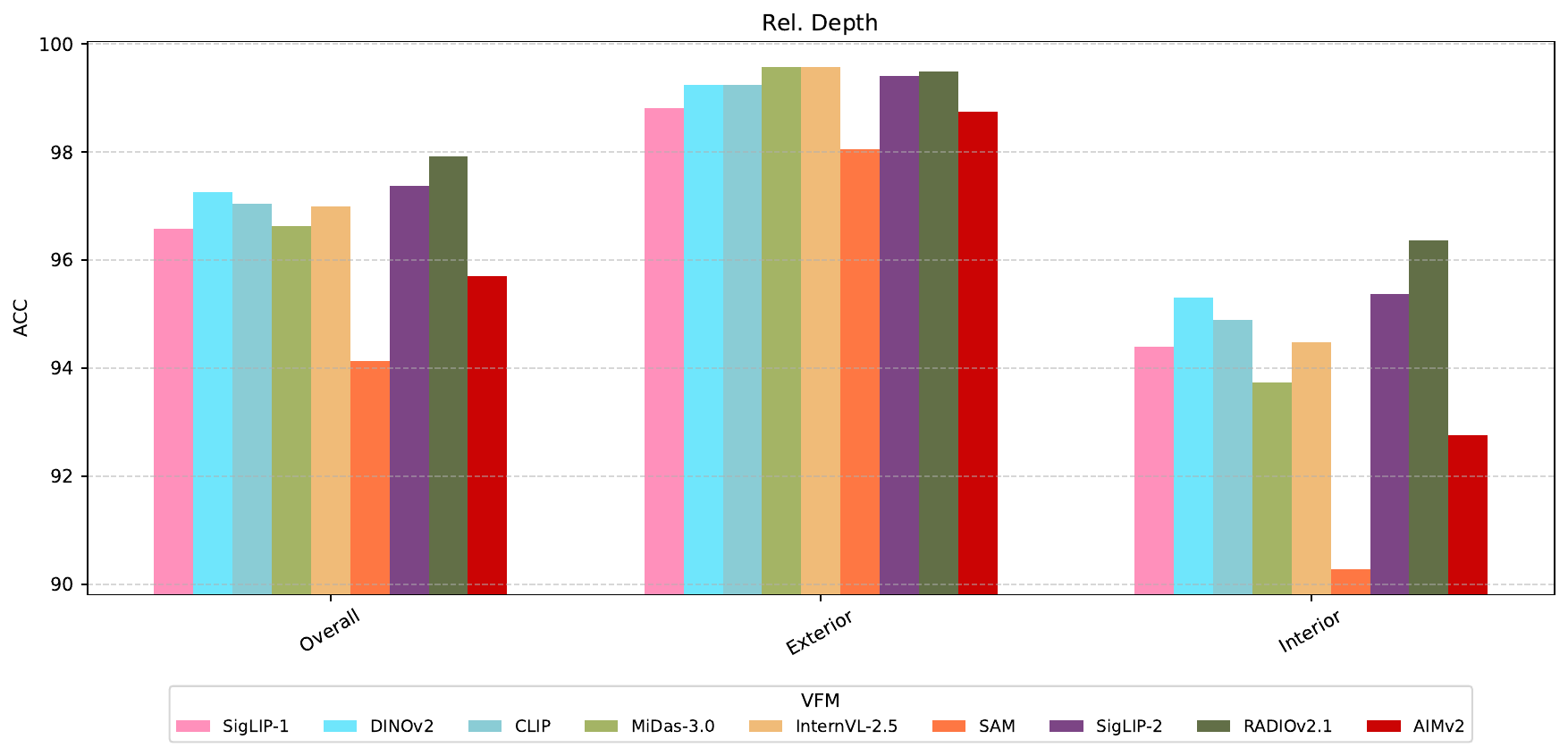}
    \caption{Detail results for relative depth for overall and different splits based on scene type.}
    \label{-fig: relative depth}
\end{figure*}

\paragraph{Relative Depth.} Relative depth estimation shows uniformly high performance across VFMs, consistently surpassing 90\% accuracy. SAM, however, underperforms notably in interior scenes, consistent with the earlier observation in absolute depth and counting that SAM handles smaller, exterior objects better (see \autoref{-fig: relative depth}).

\begin{figure*}
    \centering
    \includegraphics[width=1\linewidth]{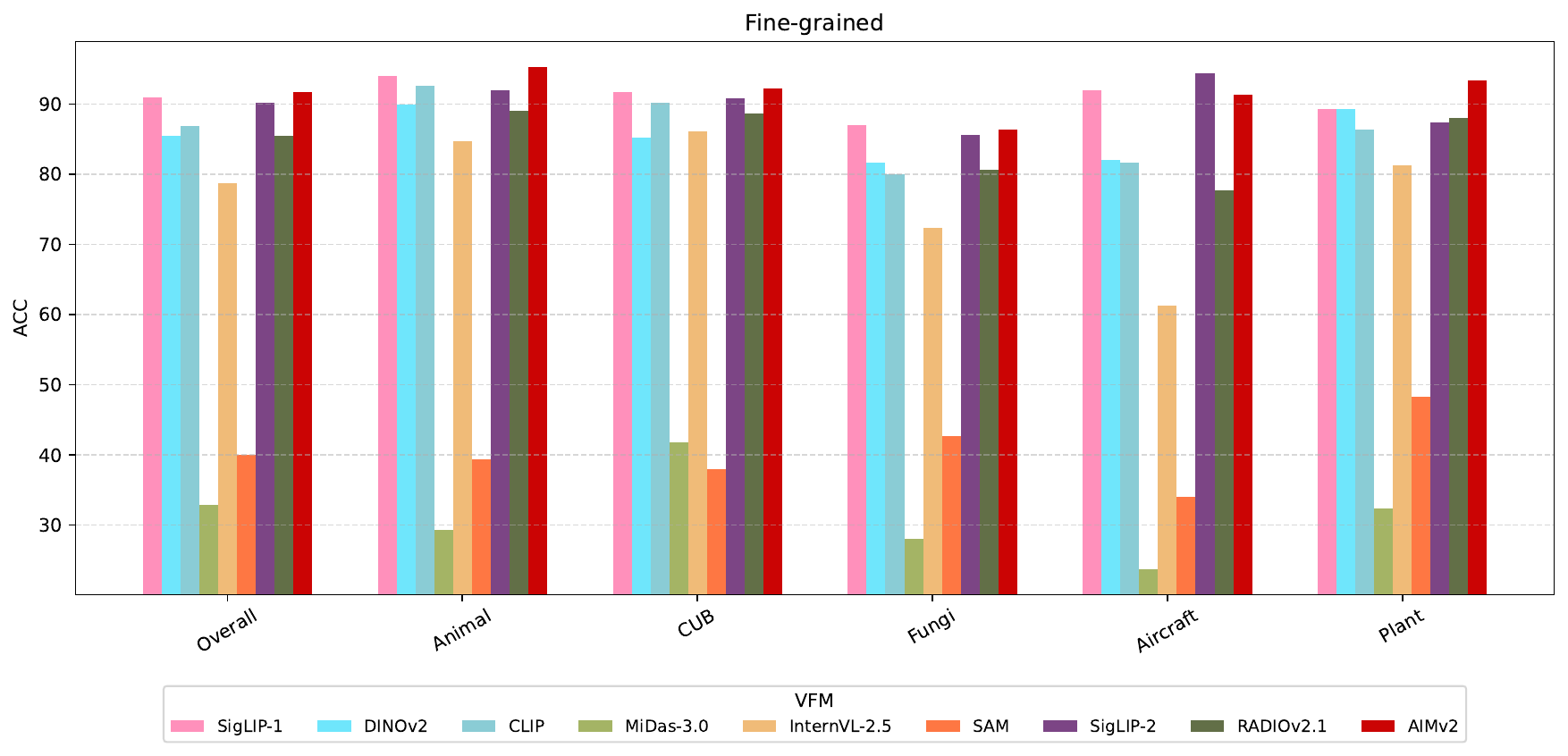}
    \caption{Detail results for fine-grained for overall and different splits based on dataset type.}
    \label{-fig: fine-grained}
\end{figure*}

\paragraph{Fine-grained Classification.} Fine-grained classification results are consistently robust across datasets, with the exception of SAM and MiDas, both of which are known to lack semantically rich features~\cite{espinosa2024there, chen2023segment}, resulting in poorer performance (see \autoref{-fig: fine-grained}).

\begin{figure*}
    \centering
    \includegraphics[width=1\linewidth]{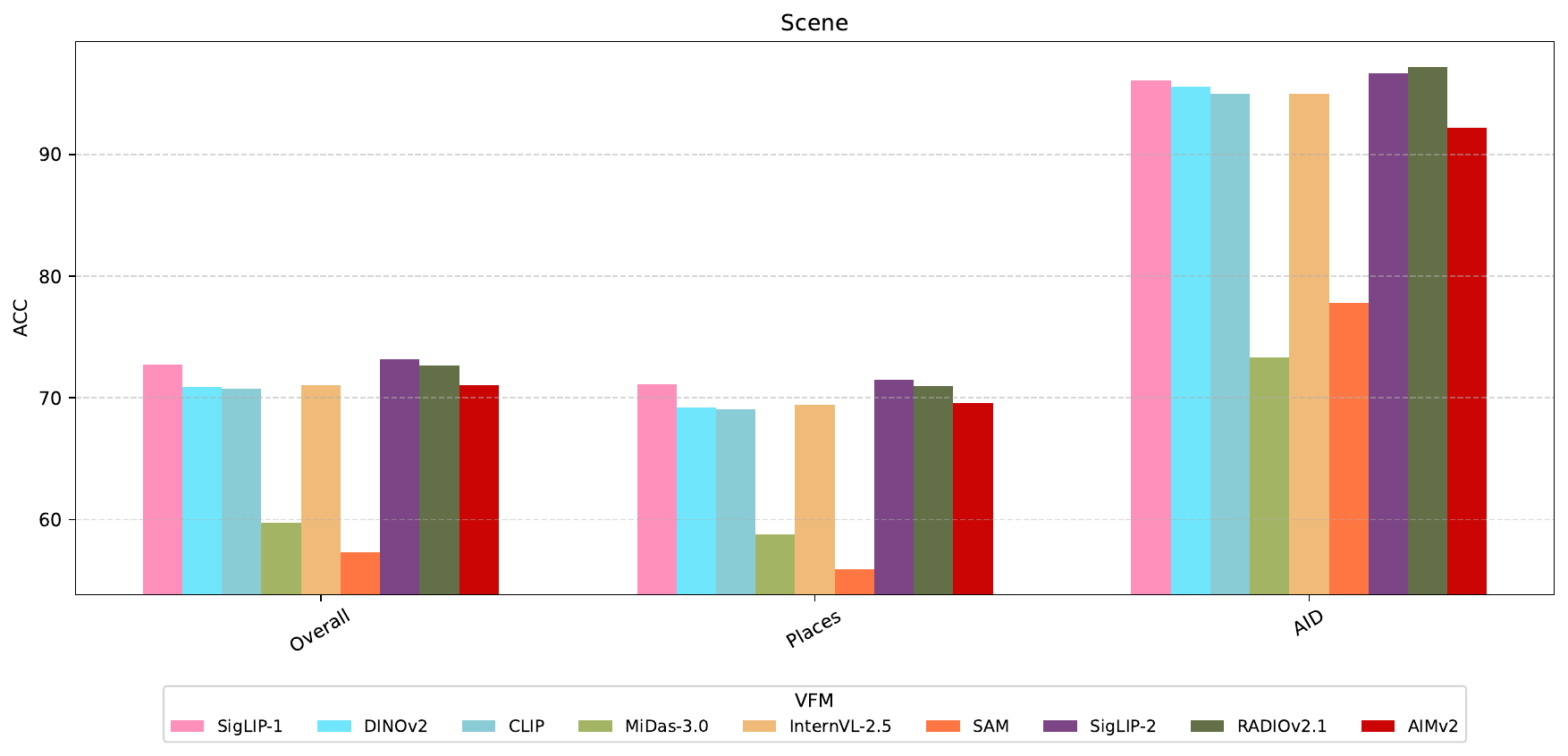}
    \caption{Detail results for scene for overall and different splits based on datasets.}
    \label{-fig: scene}
\end{figure*}

\paragraph{Scene Recognition.} Scene recognition performance is uniformly consistent across all evaluated datasets, echoing the patterns observed in fine-grained classification, where SAM and MiDas again lag behind other VFMs (see \autoref{-fig: scene}).

\begin{figure*}
    \centering
    \includegraphics[width=1\linewidth]{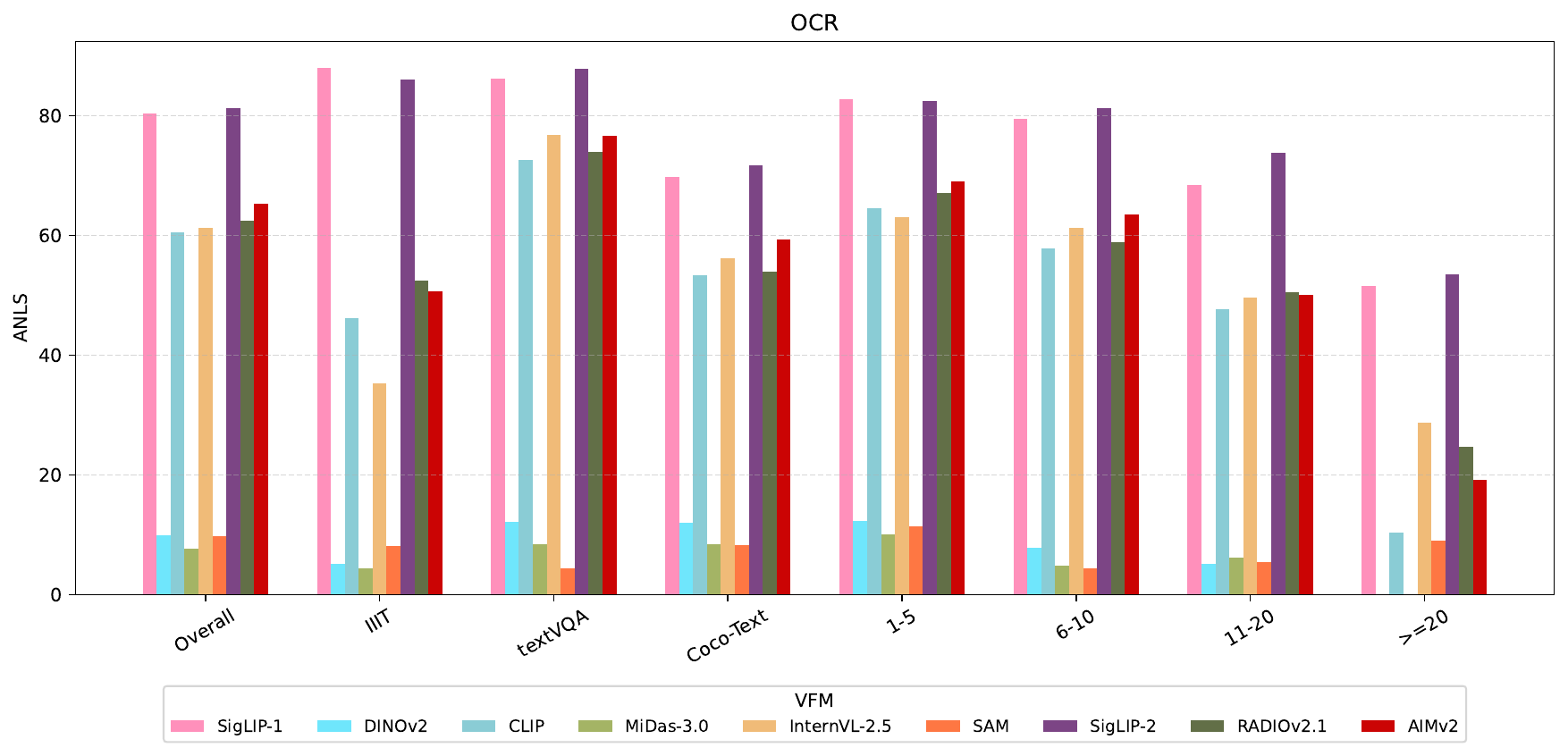}
    \caption{Detail results for counting for overall and different splits based on dataset domain and character length. Higher ANLS is better.}
    \label{-fig: ocr}
\end{figure*}

\paragraph{OCR.} OCR results show consistent patterns with those reported in Section 5.2, highlighting significant underperformance by non-language-aligned VFMs, such as DINOv2 and SAM. Notably, we observe that relative performances across VFMs are stable on short texts (length $<$ 20). However, performance for CLIP and AIM sharply declines with longer text sequences (length $>$ 20), indicating potential limitations in handling extensive textual information (see \autoref{-fig: ocr}).

\section{VFM Details}
\label{-ss: vfm_details}

\autoref{-tab: vfm_details} provides a detailed overview of the vision foundation models (VFMs) evaluated in our study. For each model, we list its architecture, parameter count, and training data, and we further summarize the training methodology in terms of supervision type, process, and loss functions.

\begin{table*}[]
\large
\resizebox{\textwidth}{!}{%
\begin{tabular}{c|p{6cm}|p{3cm}|p{3cm}|p{9cm}}
\hline
{ \textbf{VFM}} & { \textbf{Architecture}}                                                                                           & { \textbf{Parameters}}   & { \textbf{Training Data}}                         & { \textbf{Training Details}}                                                                                                                                                                                       \\\hline
{ SigLIP-2}     & { Dual-tower ViT encoders for image and text embeddings with MAP pooling layers}                                   & { So400m (400M)}         & { WebLI dataset}                                  & { \textbullet Supervision: Supervised (image-text pairs) \textbullet Process: Pretrained on 40B samples, large-batch training (32k) \textbullet Loss: Pairwise sigmoid ITC + captioning/grounding}                                               \\
{ AIMv2}        & { ViT-based vision encoder with an autoregressive multimodal decoder for patch and token reconstruction}           & { AIMv2-Huge (600M)}     & { DFN, COYO, HQITP}                               & { \textbullet Supervision: Supervised (multimodal autoregressive) \textbullet Process: Pretrain (224 px) → finetune (336/448 px), long training \textbullet Loss: Joint reconstruction of image patches and text tokens}                         \\
{ SigLIP-1}     & { Dual-encoder with independent ViT and text transformer projecting to a shared embedding space}                   & { So400m (400M)}         & { WebLI dataset}                                  & { \textbullet Supervision: Supervised (image-text pairs) \textbullet Process: Large-scale pretraining, efficient setup with 32k batch \textbullet Loss: Pairwise sigmoid contrastive loss}                                                       \\
{ CLIP}         & { Dual-tower model using a ViT image encoder and Transformer text encoder for contrastive alignment}               & { ViT-L/14 (428M)}       & { Internet-collected dataset}                     & { \textbullet Supervision: Supervised (image-text pairs) \textbullet Process: Weeks-long pretraining on multi-GPU/TPU with large batches ($\geq$32k) \textbullet Loss: Contrastive InfoNCE}                                                           \\
{ InternVL-2.5} & { Large multimodal architecture combining a high-capacity ViT encoder with an LLM for image-text reasoning}        & { InternVL2.5 (304M)}    & { FaceCaption, GQA, ChartQA, Many other datasets} & { \textbullet Supervision: Supervised (multimodal LLM) \textbullet Process: Pretrained and finetuned on diverse datasets \textbullet Loss: Autoregressive next-token + alignment losses}                                                         \\
{ RADIO v2.1}   & { ViT backbone with conditional positional encoding and multi-teacher feature distillation layers}                 & { RADIO-Huge (653M)}     & { DataComp1B dataset}                             & { \textbullet Supervision: Supervised (teacher-student distillation) \textbullet Process: 600k steps, AdamW (WD=1e-4), batch scaling law (eff. BS 1024) \textbullet Loss: Multi-teacher distillation from CLIP, DINOv2, SAM-H}                   \\
{ DINOv2}       & { ViT backbone with patch embedding and projection heads for self-supervised feature representation learning}      & { ViT-Large (300M)}      & { LVD-142M dataset}                               & { \textbullet Supervision: Self-supervised (no labels) \textbullet Process: Teacher-student distillation pipeline with deduplication and retrieval \textbullet Loss: Self-distillation contrastive objective}                                    \\
{ SAM}          & { MAE-pretrained ViT-H image encoder paired with a prompt encoder handling points, boxes, masks, and text queries} & { ViT-H (637M)}          & { SA-1B dataset}                                  & { \textbullet Supervision: Supervised (segmentation masks) \textbullet Process: Pretrained encoder with MAE, promptable training on SA-1B \textbullet Loss: Segmentation mask prediction loss}                                                   \\
{ MiDaS-3.0}    & { Multi-scale ResNet-based encoder-decoder network designed for monocular depth prediction from single images}     & { ResNet-Encoder (123M)} & { DIML Indoor, MegaDepth, ReDWeb, WSVD}            & { \textbullet Supervision: Supervised (depth ground truth) \textbullet Process: Multi-dataset pretraining, 60 epochs, Adam optimizer with different LR for new vs pretrained layers \textbullet Loss: Trimmed MAE (20\%) + gradient regularizer}\\
\hline
\end{tabular}%
}
\vspace{10pt}
\caption{Details of Vision foundation model (VFM) used, including architecture, parameter scale, training data, and training procedures.}
\label{-tab: vfm_details}
\end{table*}

\section{Related Works}
\label{-ss: related}

\subsection{VFM Evaluation}
Existing evaluation VFM protocols generally fall into two categories. The first focuses on task-specific capabilities, typically attaching tailored heads to VFMs, followed by fine-tuning and evaluation on dedicated datasets such as ImageNet for classification~\citep{han2022survey} and COCO for detection or segmentation~\citep{thisanke2023semantic}. For example, DINOv2~\citep{oquab2023dinov2} is evaluated on image and video classification, instance recognition, image retrieval, semantic segmentation, and depth estimation. For each task, a task-specific head is trained. 

To better capture the diverse and complex perception challenges of the real world, recent studies advocate a more generic approach that leverages large language models (LLMs) as general-purpose heads, evaluating VFMs on broad Visual Question Answering (VQA) benchmarks~\citep{liu2023visual,zhu2023minigpt,chowdhery2023palm}. For example,  in addition to the traditional task-specific evaluation, AIMv2~\citep{fini2024multimodal} and RADIO ~\citep{ranzinger2024radio} follow the LLM-based evaluation and use a Llama-3.0(8B) \citep{grattafiori2024llama} and a Vicuna-1.5(7B) \citep{zheng2023judging} respectively,  demonstrating a shift towards generalized multimodal evaluation.

\subsection{Atomic Visual Abilities}
As discussed in \autoref{-ss: ava}, foundational visual concepts—such as number, color, texture, object identity, and spatial relations—have long been recognized as crucial building blocks in compositional Text-to-Image (T2I) benchmarks. Given their foundational role in generation tasks, these primitives naturally underpin perceptual tasks as well. For example, a concept like 'number' directly translates into the perceptual task of counting.

A recent work \citep{chae24decomposing} introduced AVSBench to evaluate whether MLLMs understand basic \textit{geometric} features, including angle, boundary, orthogonality, and curvature, which they refer to as atomic visual skills. However, AVSBench primarily targets geometric comprehension abilities required for geometric diagrams arising in high-school level mathematics. Moreover, AVSBench provides only test data for MLLMs without addressing potential mismatches between training and test data distributions—an issue highlighted in Section 2.2. Consequently, mispredictions in AVSBench evaluations may result from data distribution mismatches rather than genuine visual deficiencies in VFM. In contrast, \ourbench explicitly emphasizes atomic visual abilities essential for general visual reasoning tasks commonly encountered in real-world scenarios. By aligning training and evaluation data distributions, \ourbench ensures that evaluation outcomes reliably reflect genuine visual perceptual capabilities of VFMs.

Additionally, a concurrent work~\citep{wu2025compact} defines a set of atomic visual capabilities analogous to ours. However, their goal fundamentally differs from ours: \citep{wu2025compact} aims to build a visual compositional tuning data recipe that builds complex capabilities from simple atomic capabilities, which can significantly reduce instruction-tuning data volume while maintaining strong performance. In contrast, \ourbench's objective is to systematically evaluate VFMs against atomic visual abilities, pinpointing their exact strengths and weaknesses, and providing a comprehensive diagnostic tool to advance the continual development~\citep{mai2022online, lomonaco2022cvpr, mai2021supervised, shim2021online, mai2020batch} of robust vision foundation models.

\section{Evaluation efficiency}
\label{-ss: eval_efficiency}

An important advantage of our framework is its efficiency compared to prior LLM-based evaluation protocols. As summarized in \autoref{-tab: efficiency_eval}, existing methods typically rely on large language models such as Vicuna-7B and require $\approx$230 A100 GPU hours per vision foundation model (VFM). By contrast, our approach adopts a lightweight 0.5B LLM and smaller training data (1.2M samples in total for stages 1 and 2), which reduces the cost to $\approx$28 A100 GPU hours while still preserving consistent and reliable VFM rankings. This design choice enables practical scaling to a wide range of models without incurring prohibitive resource demands.

\begin{table*}[]
\centering
\large
\resizebox{\textwidth}{!}{
\begin{tabular}
{c|c|c|c|c}

\hline
{\textbf{Protocol}} & {\textbf{LLM size}}        & {\textbf{Stage 1\&2 data}} & {\textbf{Stage 1\&2 cost}} & {\textbf{Stage 3}}       \\\hline
{Baseline {[}1{]}}  & {Vicuna-7B}                & {1.9M}                     & {$\approx$230 A100 h}              & {n/a}                    \\
\ourbench              & {Qwen2-0.5B$+$LoRa(stage 3)} & {1.2M}                     & {$\approx$28 A100 h}               & {Each AVA: avg 4 A100 h}\\
\hline
\end{tabular}%
}
\vspace{10pt}
\caption{ Evaluation Efficiency Table}
\label{-tab: efficiency_eval}
\end{table*}

\begin{table*}[]
\large
\resizebox{\textwidth}{!}{%
\begin{tabular}{c|c|c}
\hline
{ \textbf{Dataset}}           & { \textbf{Copyright}}                                                    & { \textbf{License}}                             \\
\hline 
{ Object365}                  & { Objects365 Consortium}                                                 & { CC By 4.0}                                    \\

{ LVIS}                       & { LVIS Consortium}                                                       & { CC By 4.0}                                    \\

{ iNaturalist-2021}           & { iNaturalist (Terms of Service)}                                        & { MIT}                                          \\

{ DIOR}                       & { N/A}                                                                   & { N/A}                                          \\

{ VQAv2}                      & { VQA Consortium}                                                        & { CC BY 4.0}                                    \\

{ FSC}                        & { CVLab at StonyBrook}                                                   & { MIT}                                          \\

{ CARPK}                      & { Original image owners (PUCPR/PKLot)}                                   & { N/A}                                          \\

{ Crowd Surveillance Dataset} & { N/A}                                                                   & { N/A}                                          \\

{ CUB-200-2011}               & { Annotations: Catherine Wah et al.; images: original owners}            & { CC0 (Public Domain)}                          \\

{ FGVC-Aircraft}              & { Annotations: S. Maji et al.; images: original owners}                  & { Research only (Non-commercial)}               \\

{ KITTI}                      & { Andreas Geiger, Philip Lenz, Christoph Stiller, Raquel Urtasun}        & { CC BY-NC-SA 3.0}                              \\

{ NYU-DepthV2}                & { N/A}                                                                   & { N/A}                                          \\

{ coco-text}                  & { SE(3) Computer Vision Group, Cornell Tech}                             & { CC BY 4.0}                                    \\

{ IIIT5K}                     & { IIIT Hyderabad (annotations); images: original owners}                 & { N/A}                                          \\

{ TextVQA}                    & { VQA Consortium}                                                        & { CC BY 4.0}                                    \\

{ EgoOrientBench}             & { N/A}                                                                   & { N/A}                                          \\

{ CURE-OR}                    & { OLIVES at Georgia Institute of Technology}                             & { MIT}                                          \\

{ Moment\_int\_time}          & { Moments in Time authors}                                               & { Research/Educational only}                    \\

{ DTD}                        & { N/A}                                                                   & { N/A}                                          \\

{ KTH-TIPS}                   & { N/A}                                                                   & { N/A}                                          \\

{ KTH-TIPS2}                  & { N/A}                                                                   & { N/A}                                          \\

{ Places365}                  & { MIT CSAIL, Bolei Zhou; images: original owners}                        & { MIT (code); images original copyright owners} \\

{ AID}                        & { AID authors (Gui-Song Xia et al.); images from Google Earth providers} & { N/A}                                          \\

{ RAF-DB}                     & { N/A}                                                                   & { N/A}                                          \\

{ ExpW}                       & { N/A}                                                                   & { N/A}  \\                                       
\hline
\end{tabular}%
}
\vspace{10pt}
\caption{ Dataset copyright and licensing information for all datasets used in \ourbench}
\label{-tab: dataset_license}
\end{table*}

For stage 3, our framework further leverages LoRA-based fine-tuning, where each AVA is trained on only 6K--10K samples. This procedure requires $\approx$4 A100 GPU hours per AVA on average, making it highly lightweight compared to full model finetuning. In summary, \ourbench achieves a more diagnostic evaluation with considerably lower overhead than prior work.

\section{Dataset Copyright/License}
\label{-ss: license}

To ensure ethical and legal use of datasets, we summarize the copyright and licensing information of all benchmarks employed in our experiments (\autoref{-tab: dataset_license}). The majority of the datasets we use are publicly available under open licenses such as CC BY 4.0, MIT, or CC0, which permit research and redistribution with proper attribution. Some datasets (e.g., FGVC-Aircraft, Moments in Time) are restricted to research-only or educational use, and we adhered to these conditions. For datasets without explicit licensing details, we used them strictly within the scope of non-commercial academic research.




\end{document}